\documentclass{article}

\usepackage{microtype}
\usepackage{amssymb}
\usepackage{amsmath}
\usepackage{graphicx}
\usepackage{subfigure}
\usepackage{placeins}
\usepackage{multirow}
\usepackage{booktabs} 
\usepackage{graphbox}
\usepackage{tcolorbox}
\usepackage{paralist}
\usepackage{soul}
\usepackage{pifont}

\definecolor{lightblue}{HTML}{ff7f2a}
\definecolor{lighterblue}{HTML}{ffe6d5}
\newtcolorbox{mybox}{colback=lighterblue,colframe=lightblue}

\usepackage{hyperref}

\usepackage[accepted]{arxiv}
\usepackage{xspace}

\icmltitlerunning{What Are Bayesian Neural Network Posteriors Really Like?}

\begin{document}

\twocolumn[
\icmltitle{What Are Bayesian Neural Network Posteriors Really Like?}

\begin{center}
    \begin{tabular}{cccc}
         \textbf{Pavel Izmailov} & \textbf{Sharad  Vikram} & \textbf{Matthew D. Hoffman} & \textbf{Andrew Gordon Wilson} \\
         New York University     & Google Research         & Google Research             & New York University
    \end{tabular}
\end{center}

\vskip 0.3in
]

\begin{abstract}
The posterior over Bayesian neural network (BNN) parameters is extremely high-dimensional and non-convex. For computational reasons, researchers approximate this posterior using inexpensive mini-batch methods such as mean-field variational inference or stochastic-gradient Markov chain Monte Carlo (SGMCMC). 
To investigate foundational questions in Bayesian deep learning, we instead use full-batch Hamiltonian Monte Carlo (HMC) on modern architectures. 
We show that 
(1) BNNs can achieve significant performance gains over standard training and deep ensembles; 
(2) a single long HMC chain can provide a comparable representation of the posterior to multiple shorter chains; 
(3) in contrast to recent studies, we find posterior tempering is not needed for near-optimal performance, with little evidence for a ``cold posterior'' effect, which we show is largely an artifact of data augmentation; 
(4) BMA performance is robust to the choice of prior scale, and relatively similar for diagonal Gaussian, mixture of Gaussian, and logistic priors;
(5) Bayesian neural networks show surprisingly poor generalization under domain shift;
(6) while cheaper alternatives such as deep ensembles and SGMCMC can provide good generalization, they provide distinct predictive 
distributions from HMC. Notably, deep ensemble predictive distributions are similarly close to HMC as standard SGLD, and closer than standard
variational inference.
\end{abstract}

\section{Introduction}

Over the last 25 years, there have been several strong arguments favouring 
a Bayesian approach to deep learning \citep[e.g., ][]{mackay1995probable, neal1996, blundell2015weight, gal2016uncertainty, wilson2020bayesian}.
Bayesian inference for neural networks promises improved predictions, reliable uncertainty estimates, and principled model comparison, naturally supporting active learning, continual learning, and decision-making under uncertainty.
The Bayesian deep learning community has designed multiple successful practical methods inspired by the Bayesian approach \citep{blundell2015weight, gal2016dropout, welling2011bayesian, kirkpatrick2017overcoming, maddoxfast2019, izmailov2019subspace, daxberger2020expressive} with applications ranging from astrophysics \citep{Cranmer2021ABN} to automatic diagnosis of Diabetic Retinopathy \citep{filos2019systematic}, click-through rate prediction in advertising \citep{liu2017pbodl} and modeling of fluid dynamics \citep{geneva2020modeling}.

However, inference with modern neural networks is distinctly challenging. We wish to compute a Bayesian model average corresponding to an integral over a multi-million dimensional multi-modal posterior, with unusual topological properties like mode-connectivity \citep{garipov2018, draxler2018essentially}, under severe computational constraints. 

There are therefore many unresolved questions about Bayesian deep learning practice. Variational procedures typically provide unimodal Gaussian approximations to the multimodal posterior. Practically successful methods such as deep ensembles \citep{lakshminarayanan2017simple, fort2019deep} have a natural Bayesian interpretation \citep{wilson2020bayesian}, but only represent modes of the posterior. While Stochastic MCMC \citep{welling2011bayesian, chen2014stochastic, zhang2019cyclical} is computationally convenient, it could be providing heavily biased
estimates of posterior expectations. Moreover, \citet{wenzel2020good} question the quality of standard Bayes posteriors, citing results where ``cold posteriors'', raised to a power $1/T$ with $T<1$, improve performance.

Additionally, Bayesian deep learning methods are typically evaluated on their ability to generate useful, well-calibrated predictions on held-out or out-of-distribution data. 
However, strong performance on benchmark problems does not imply that the algorithm accurately approximates the true Bayesian model average (BMA).

In this paper, we investigate fundamental open questions in Bayesian deep learning, using multi-chain full-batch Hamiltonian Monte Carlo 
\citep[HMC,][]{neal2011mcmc}.
HMC is a highly-efficient and well-studied Markov Chain Monte Carlo (MCMC) method that is guaranteed to asymptotically produce samples from the true posterior. 
However it is enormously challenging to apply HMC to modern neural networks due to its extreme computational requirements: HMC can take \textit{tens of thousands of training epochs} to produce a single sample from the posterior.
To address this computational challenge, we parallelize the computation over hundreds of Tensor processing unit (TPU) devices.

We argue that full-batch HMC provides the most precise tool for studying the BNN posterior to date. We are not proposing HMC
as a computationally efficient method for practical applications. Rather, using our implementation of HMC 
we are able to explore fundamental questions about posterior geometry, the performance of BNNs, approximate inference, effect of priors and posterior temperature.

In particular, we show:
(1) BNNs can achieve significant performance gains over standard training and deep ensembles; 
(2) a single long HMC chain can provide a comparable representation of the posterior to multiple shorter chains; 
(3) in contrast to recent studies, we find posterior tempering is not needed for near-optimal performance, with little evidence for a ``cold posterior'' effect, which we show is largely an artifact of data augmentation; 
(4) BMA performance is robust to the choice of prior scale, and relatively similar for diagonal Gaussian, mixture of Gaussian, and logistic priors over weights. This result highlights the importance of architecture relative to parameter priors in specifying the prior over functions.
(5) While Bayesian neural networks have good performance for OOD detection, they show surprisingly poor generalization under domain shift;
(6) while cheaper alternatives such as deep ensembles and SGMCMC can provide good generalization, they provide distinct predictive 
distributions from HMC. Notably, deep ensemble predictive distributions are similarly close to HMC as standard SGLD, and closer than standard
variational inference.

We additionally show how to effectively deploy full batch HMC on modern neural networks, including insights about how to tune crucial hyperparameters for good performance, and parallelize sampling over hundreds of TPUs. 
Our HMC samples and implementation will be a \href{https://github.com/google-research/google-research/tree/master/bnn_hmc}{\underline{public resource}}.
We hope this resource will both serve as a reference in evaluating and calibrating more practical alternatives to HMC, and aid in exploring questions for a better understanding of approximate inference in Bayesian deep learning.

\section{Background}
\label{sec:background}

\textbf{Bayesian neural networks}\quad
The goal of classical learning is to find a single best setting of the parameters for the model, typically 
through maximum-likelihood optimization.
In the Bayesian framework, the learner instead infers a \textit{posterior} distribution $p(w \vert \mathcal{D})$ over the
parameters $w$ of the model after observing the data $\mathcal{D}$.
The posterior distribution is given by Bayes' rule:
$p(w \vert \mathcal{D}) \propto p(\mathcal{D} \vert w) p(w)$, where $p(\mathcal{D} \vert w)$ is the likelihood of $\mathcal{D}$ given by the model with parameters $w$,
and $p(w)$ is the prior distribution over the parameters.
The predictions of the model on a new test example $x$ are then given by the \textit{Bayesian model average} (BMA)
\begin{equation}
    \label{eq:bma}
    \textstyle
    p(y \vert x, \mathcal{D}) = \int_w p(y \vert x, w) p(w \vert \mathcal{D}) dw,
\end{equation}
where $p(y \vert x, w)$ is the predictive distribution for a given value of the parameters $w$.
This BMA is particularly compelling in Bayesian deep learning, because the posterior over
parameters for a modern neural network can represent many complementary solutions to a 
given problem, corresponding to different settings of parameters \citep{wilson2020bayesian}.
Unfortunately, the integral in Eq.~\eqref{eq:bma} cannot be evaluated in closed form for 
neural networks, so one must resort to approximate inference. Moreover, approximating 
Eq.~\eqref{eq:bma} is challenging due to a high dimensional and sophisticated posterior 
$p(w \vert \mathcal{D})$. For a detailed discussion of Bayesian deep learning, see e.g. \citet{wilson2020bayesian}.

\textbf{Markov Chain Monte Carlo.}\quad
The integral in Eq. \eqref{eq:bma} can be approximated by sampling:
$p(y \vert x, \mathcal{D}) \approx \frac 1 M \sum_{i=1}^M p(y \vert x, w_i)$,
where $w_i \sim p(w \vert \mathcal{D})$ are samples drawn from the posterior.
MCMC methods construct a Markov chain that, if simulated for long enough, generates approximate samples from the posterior.
In this work, we focus on Hamiltonian Monte Carlo \citep{neal2011mcmc}, a method that produces
asymptotically exact samples  assuming access to the unnormalized
posterior density $p(\mathcal{D} \vert w) p(w)$ and its gradient.

\section{Related work}
\label{sec:rel_work}

The bulk of work on Bayesian deep learning has focused on scalable approximate inference methods.
These methods include stochastic variational inference
\citep{hoffman2013stochastic, graves2011practical, blundell2015weight, kingma2015variational, molchanov2017variational, louizos2017multiplicative, khan2018fast, zhang2018noisy, wu2018deterministic,osawa2019practical,dusenberry2020efficient}, 
dropout \citep{srivastava2014dropout,gal2016dropout,kendall2017uncertainties, gal2017concrete},
the Laplace approximation \citep{mackay1992practical, kirkpatrick2017overcoming, ritter2018scalable, li00, daxberger2020expressive},
expectation propagation \citep{hernandez2015probabilistic},
and leveraging the stochastic gradient descent (SGD) trajectory, either for a deterministic approximation,
or sampling as in SGLD
\citep{mandt2017stochastic,maddoxfast2019,izmailov2018, wilson2020bayesian}. \citet{foong2019expressiveness} and \citet{farquhar2020liberty} 
additionally consider the role of expressive posterior approximations in variational inference.

While these (and many other) methods often provide improved predictions or uncertainty estimates,
to the best of our knowledge \textit{none} of these methods have been directly evaluated on their
ability to match the true posterior distribution using practical architectures and datasets.
Moreover, many of these methods are often designed with train-time constraints in mind, to take roughly the same
amount of compute as regular SGD training.
To evaluate approximate inference procedures, and explore fundamental questions in Bayesian deep learning, 
we attempt to construct a posterior approximation of the highest possible quality,
ignoring the practicality of the method.

The Monte Carlo literature for Bayesian neural networks has mainly focused on stochastic gradient-based
methods \citep{welling2011bayesian,ahn2014distributed,chen2014stochastic,ma2015complete,ahn2012bayesian,ding2014bayesian, zhang2019cyclical, garriga2021exact} 
for computational efficiency reasons.
These methods are fundamentally biased: (1) they omit the Metropolis-Hastings correction step,
and (2) the noise from subsampling the data perturbs their stationary distribution.
In particular, \citet{betancourt2015fundamental} argues that HMC is incompatible with data subsampling.
Notably, \citet{zhang2020amagold} recently proposed a second-order stochastic gradient MCMC method that
is asymptotically exact.

Since the classic work of \citet{neal1996}, there have been a few recent attempts at using full-batch HMC in BNNs \citep[e.g.; ][]{cobb2020scaling,wenzel2020good}. These studies tend to use relatively short trajectory lengths (generally not considering a number of leapfrog steps greater than 100), and tend to focus on relatively small datasets and networks.
We on the other hand experiment with practical architectures and datasets and use up to $10^5$ leapfrog steps per iteration to ensure good mixing. 

Our work is aimed at \textit{understanding} the properties of true Bayesian neural networks.
In a similar direction, \citet{wenzel2020good} have recently explored the effect of the posterior temperature in Bayesian
neural networks. 
We discuss their results in detail in \autoref{sec:cold_posteriors}, and
provide our own exploration of the posterior temperature with a different result:
we find that BNNs achieve strong performance at temperature $1$ and do not require
posterior tempering. Moreover, the scope of our paper extends well beyond temperature scaling,
revealing for instance that while BNNs can provide strong in-domain generalization, they surprisingly 
suffer on the covariate shift problems that approximate inference methods are often applied towards.
We also show that while deep ensembles are often treated as a non-Bayesian alternative, they in fact
are providing higher fidelity approximations of the Bayesian model average than standard approximate inference
procedures, as argued in \citet{wilson2020bayesian}. We also explore several other key questions, including 
prior selection and posterior geometry.

\begin{figure*}[h]
    \centering

    \hspace{-.8cm}
    \label{fig:trajectory-length-ablation}
    \begin{tabular}{cc}
        \begin{tabular}{cccc}
        	\hspace{-.35cm}
            \rotatebox{90}{\scriptsize \quad\quad \textbf{CIFAR-10}} &
            \hspace{-.4cm}
            \includegraphics[height=0.105\textwidth]{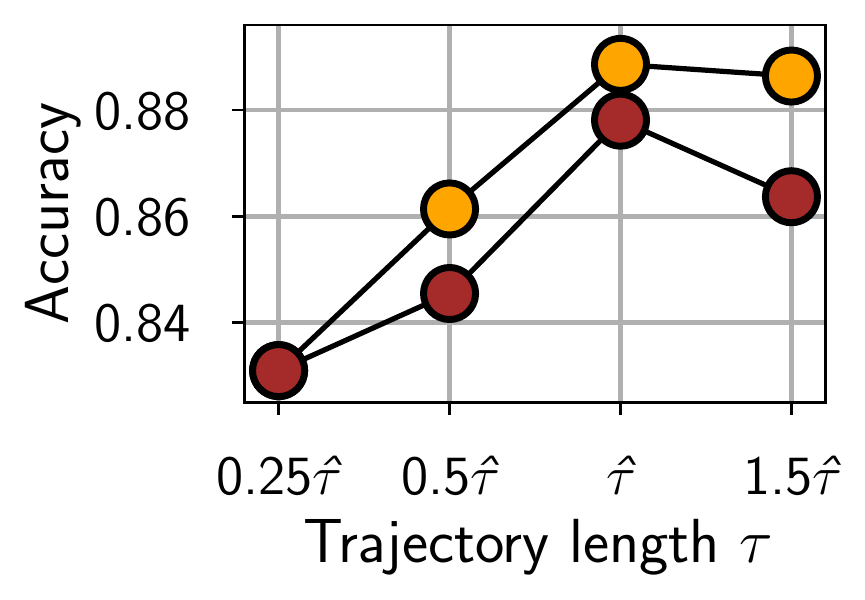} &
            \hspace{-.4cm}
            \includegraphics[height=0.105\textwidth]{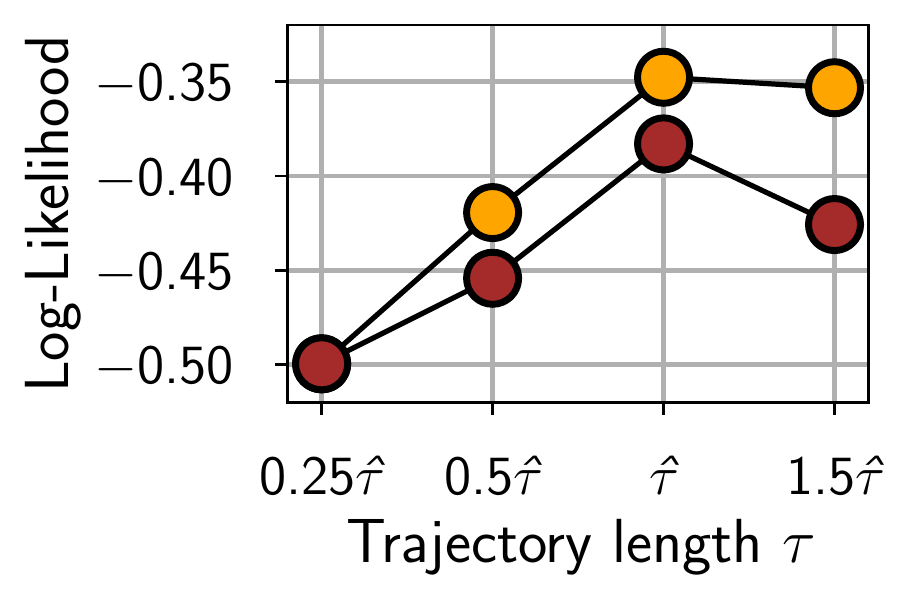} &
            \hspace{-.6cm}
            \includegraphics[height=0.105\textwidth]{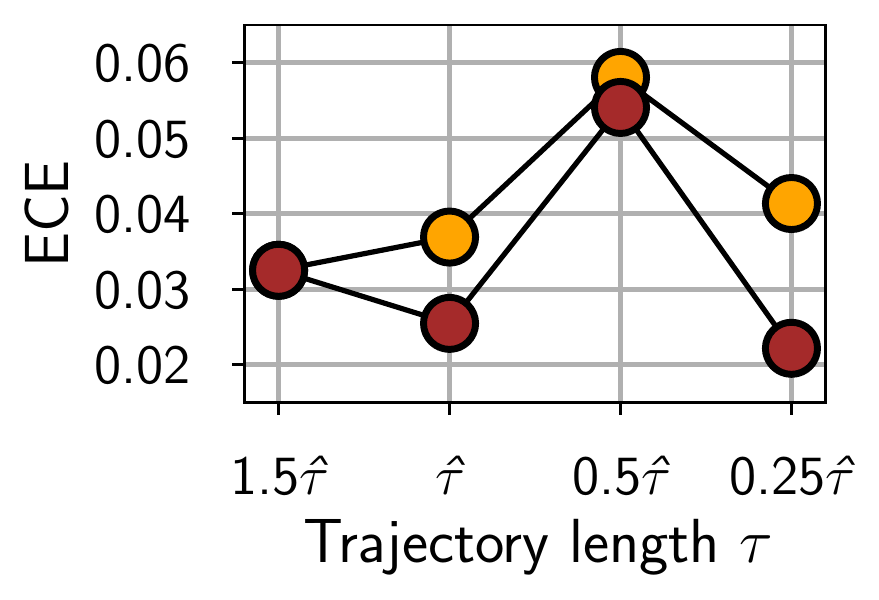}
            \\
            \hspace{-.2cm}
            \rotatebox{90}{\scriptsize \quad\quad\quad \textbf{IMDB}} &
            \hspace{-.4cm}
            \includegraphics[height=0.105\textwidth]{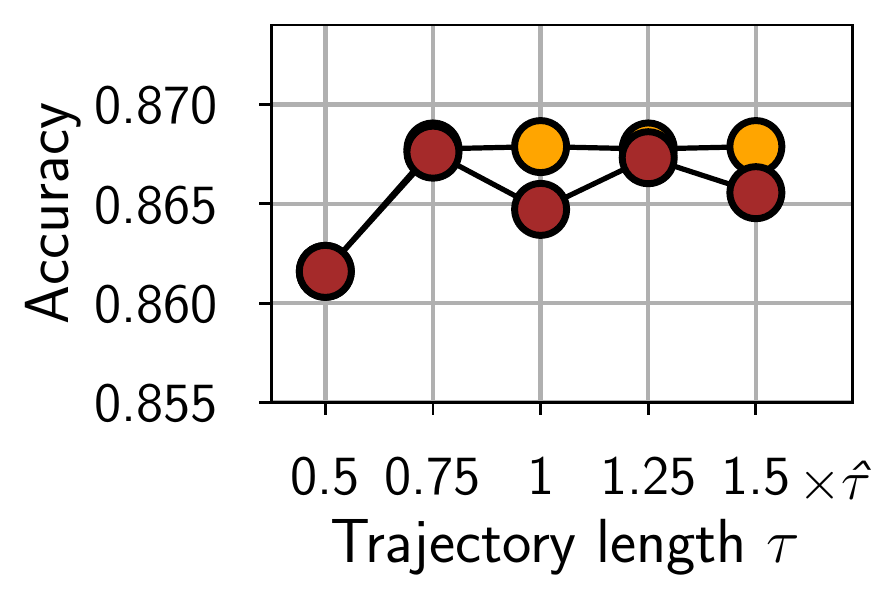} &
            \hspace{-.4cm}
            \includegraphics[height=0.105\textwidth]{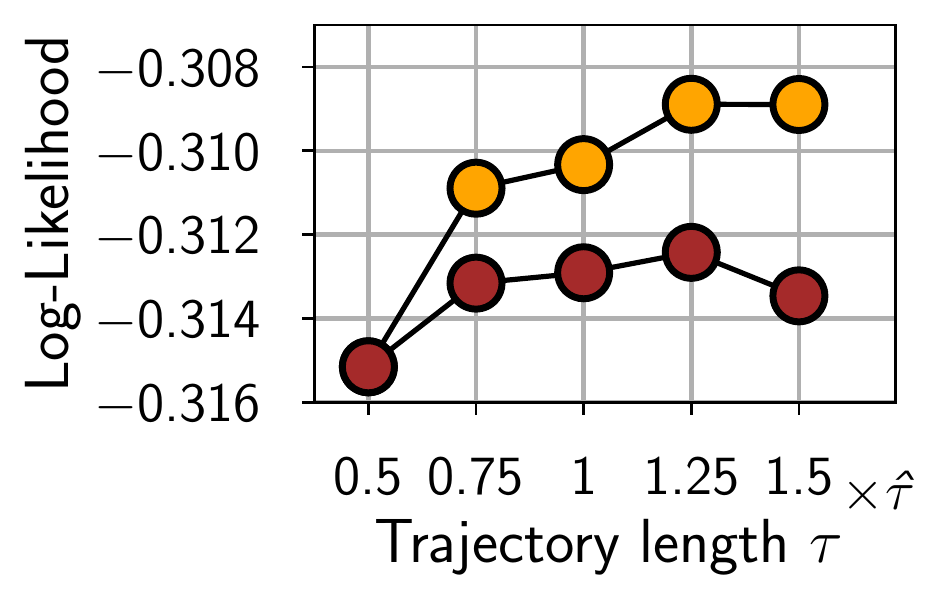} &
            \hspace{-.6cm}
            \includegraphics[height=0.105\textwidth]{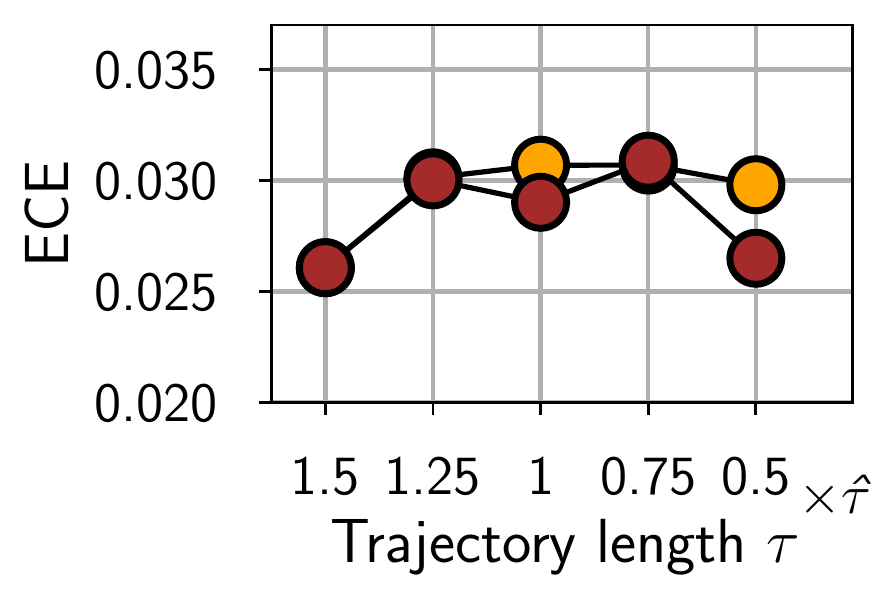} 
        \end{tabular}
        &
        \hspace{-.35cm}
        \label{fig:num-chains-ablation}
        \begin{tabular}{cccc}
            \includegraphics[height=0.105\textwidth]{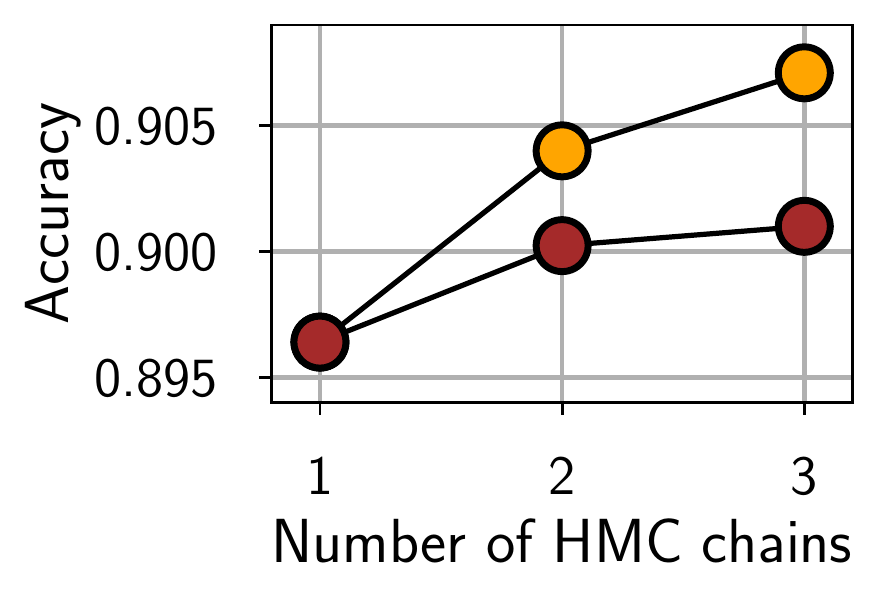} &
            \hspace{-.4cm}
            \includegraphics[height=0.105\textwidth]{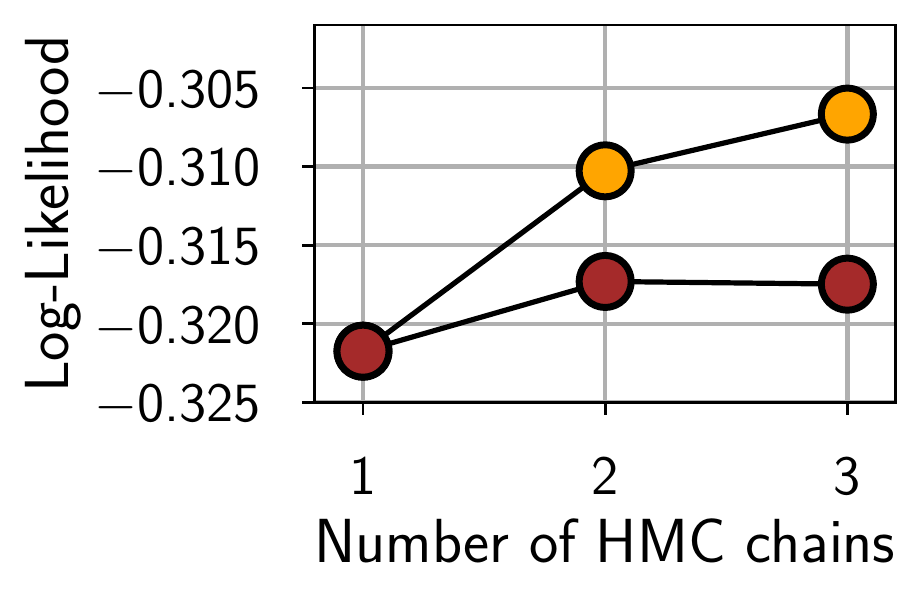} &
            \hspace{-.6cm}
            \includegraphics[height=0.105\textwidth]{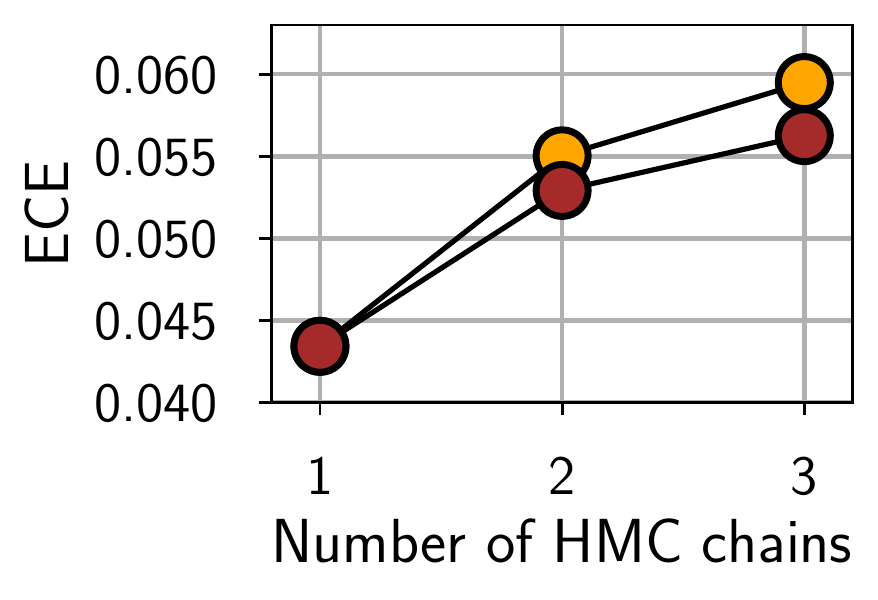}
            \\
            \includegraphics[height=0.105\textwidth]{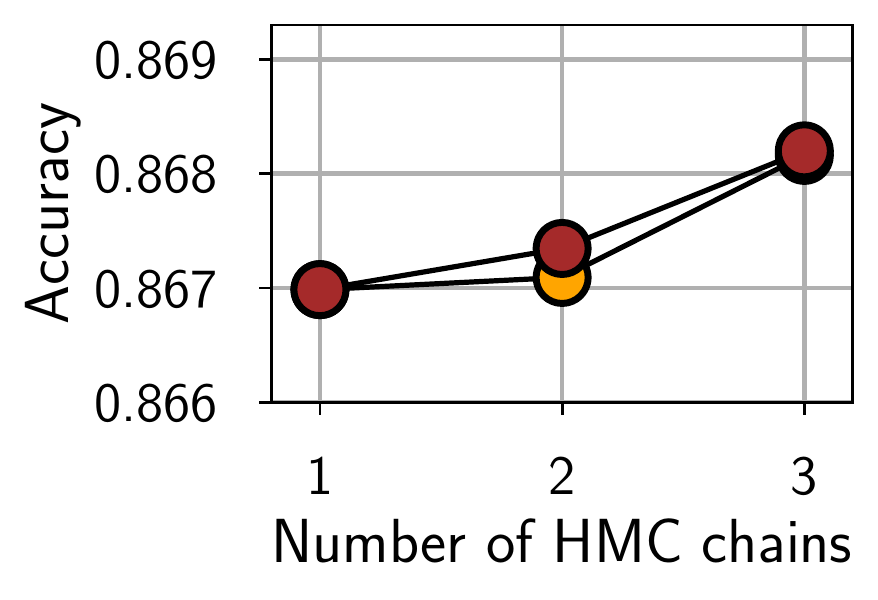} &
            \hspace{-.4cm}
            \includegraphics[height=0.105\textwidth]{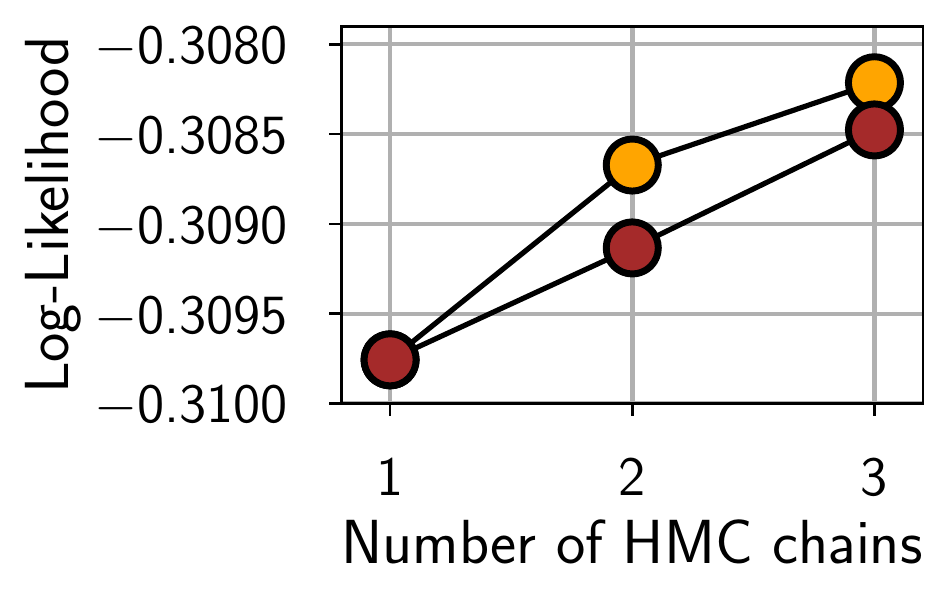} &
            \hspace{-.6cm}
            \includegraphics[height=0.105\textwidth]{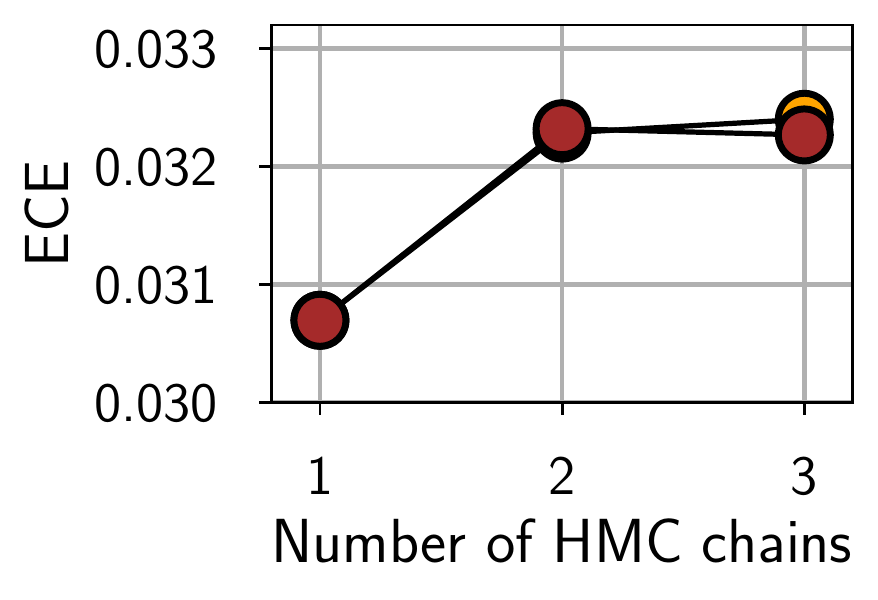}
        \end{tabular}
        \\
        \multicolumn{2}{c}{\includegraphics[height=0.09\textwidth]{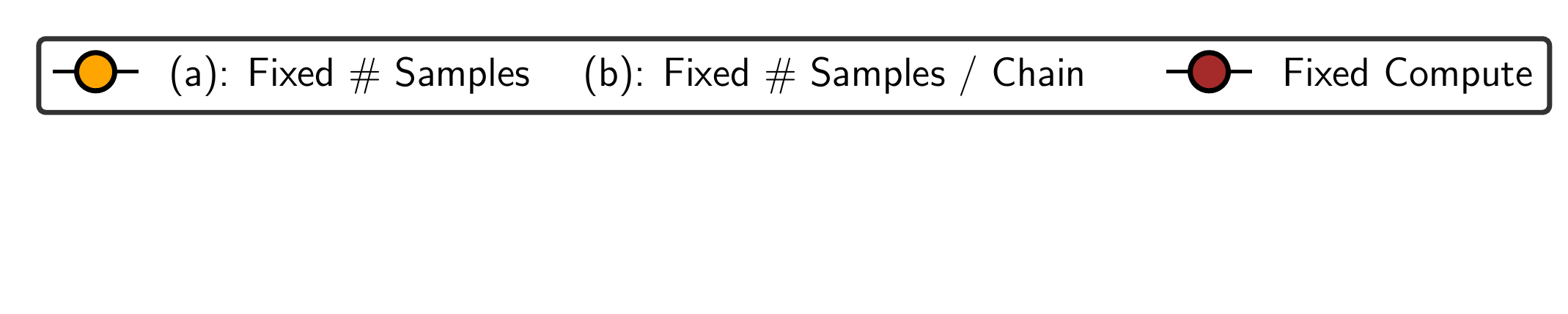}}\\[-0.7cm]
        (a) Trajectory length ablation & (b) Number of chains ablation
    \end{tabular}

	\caption{
	    \textbf{Effect of HMC hyper-parameters.}
	    BMA accuracy, log-likelihood and expected calibration error (ECE)
        as a function of \textbf{(a)}: the trajectory length $\tau$ and \textbf{(b)}: number of HMC chains.
        The orange curve shows the results for a fixed number of samples in (a) and for a fixed number
        of samples per chain in (b); the brown curve shows the results for a fixed amount of compute.
        All experiments are done on CIFAR-10 using the ResNet-20-FRN architecture on IMDB using CNN-LSTM.
	    Longer trajectory lengths decrease correlation between subsequent samples improving accuracy and
	    log-likelihood. 
	    For a given amount of computation, increasing the number of chains from one to two modestly improves the accuracy and log-likelihood.
	}
	\label{fig:tau_ablation}
\end{figure*}

\section{HMC for deep neural networks}
\label{sec:hmc}

We use full-batch Hamiltonian Monte Carlo (HMC) to sample from the parameter posteriors for Bayesian neural networks.
In this section, we show how to make HMC effective for modern Bayesian neural networks, discussing important details 
such as hyper-parameter specification. In the next sections, we use the HMC samples to explore fundamental
questions about approximate inference in modern deep learning.
We summarize HMC in Appendix \autoref{alg:hmc} and \autoref{alg:leapfrog}.
Intuitively, HMC is simulating the dynamics of a particle sliding on the plot of the density function that we are trying to sample from.\footnote{
For a detailed introduction to HMC please see \citet{neal2011mcmc}.
See also interactive visualization here: {\small\url{http://chi-feng.github.io/mcmc-demo/}}.
}

\textbf{Implementation.}\quad 
To scale HMC to modern neural network architectures and
for datasets like CIFAR-10 and IMDB, we parallelize the computation over 512 TPUv3  devices\footnote{
We use other hardware configurations in several experiments. We state the hardware that we used in the corresponding sections.} \citep{jouppi2020domain}. 
We execute HMC in a single-program multiple-data (SPMD) configuration, wherein a dataset is sharded evenly over each of the devices and an identical HMC implementation is run on each device.
Each device maintains a synchronized copy of the Markov chain state, where the full-batch gradients needed for leapfrog integration are computed using cross-device collectives.
We release our JAX \citep{jax2018github} \href{https://github.com/google-research/google-research/tree/master/bnn_hmc}{\underline{implementation}}.

\textbf{Neural network architectures.}\quad
In our evaluation, following \citet{wenzel2020good}, we primarily focus on two architectures: ResNet-20-FRN and CNN-LSTM.
ResNet-20-FRN is a residual architecture \citep{he2016deep} of depth $20$ with batch normalization layers \citep{ioffe2015batch} replaced with filter 
response normalization \citep[FRN; ][]{singh2020filter}.
Batch normalization makes the likelihood harder to interpret by creating dependencies between training examples, whereas
the outputs of FRN layers are independent across inputs.
We use Swish (SiLU) activations \citep{hendrycks2016gaussian, elfwing2018sigmoid, ramachandran2017searching} instead of ReLUs to ensure smoothness of the posterior density surface, which we found improves acceptance rates
of HMC proposals without hurting the overall performance.
The CNN-LSTM is a long short-term memory network \citep{hochreiter1997long} adapted from \citet{wenzel2020good} without modifications.

\textbf{Datasets and Data Augmentation.}\quad
In our main evaluations we use the CIFAR image classification datasets \citep{krizhevsky2014cifar} and the IMDB dataset \citep{imdb} for sentiment analysis.
We do not use any data augmentation, both because the random augmentations introduce stochasticity 
into the evaluation of the posterior log-density and its gradient, and because the expected randomly perturbed log-likelihood does not have a clean interpretation as a valid likelihood function
\citep{wenzel2020good}.
In this section we perform ablations using ResNet-20-FRN on CIFAR-10 and CNN-LSTM on IMDB.

\subsection{Trajectory length $\tau$}
\label{sec:trajectory}

The trajectory length parameter $\tau$ determines the length of the dynamics simulation on each HMC iteration.
Effectively, it determines the correlation of subsequent samples produced by HMC.
To suppress random-walk behavior and speed up mixing, we want the length of the trajectory to be relatively high.
But increasing the length of the trajectory also leads to an increased computational cost: 
the number of evaluations of the gradient of the target density (evaluations of the gradient of the loss on the full dataset)
is equal to the ratio $\tau / \Delta$ of the trajectory length to the step size.

We suggest the following value of the trajectory length $\tau$:
\begin{equation}
    \label{eqn:tau}
    \textstyle \hat \tau = \frac{\pi \alpha_{\text{prior}}}{2},
\end{equation}
where $\alpha_{\text{prior}}$ is the standard deviation of the prior distribution over the parameters.
If applied to a spherical Gaussian distribution, HMC with a small step size and this trajectory length will generate exact samples\footnote{Since the Hamiltonian defines a set of independent harmonic oscillators with period $2\pi\alpha$, $\tau=\pi\alpha/2$ applies a quarter-turn in phase space, swapping the positions and momenta.}.
While we are interested in sampling from the posterior rather than from the spherical Gaussian prior, we argue that in large BNNs the prior tends to determine the scale of the posterior.
We provide more detail and confirm this intuition empirically in the \autoref{sec:app_marginals}.

In order to test the validity of our recommended trajectory length, we perform an ablation 
and report the results in \autoref{fig:tau_ablation}(a).
As expected, longer trajectory lengths provide better performance in terms of accuracy and log-likelihood.
Expected calibration error is generally low across the board.
The trajectory length $\hat \tau$ provides good performance in all three metrics.
This result confirms that, despite the expense, when applying HMC to BNNs it is actually helpful to use tens of thousands of gradient evaluations per iteration.

\subsection{Step size $\Delta$}

The step size parameter $\Delta$ determines the discretization step size of the Hamiltonian dynamics and
consequently the number of leapfrog integrator steps. 
Lower step sizes lead to a better approximation of the dynamics and higher rates of
proposal acceptance at the Metropolis-Hastings correction step.
However, lower step sizes require more gradient evaluations per iteration to hold the trajectory length $\tau$ constant.

Using ResNet-20-FRN on CIFAR-10, we run HMC for 50 iterations with step sizes of $1 \cdot 10^{-5}$, $5 \cdot 10^{-5}$, $1 \cdot 10^{-4}$, and $5 \cdot 10^{-4}$ respectively, ignoring the Metropolis-Hastings correction. We find the chains achieve average accept probabilities of $72.2\%$, $46.3\%$, $22.2\%$, and $12.5\%$, reflecting large drops in accept probability
as step size is increased. We also observe BMA log-likelihoods of $-0.331$, $-0.3406$, $-0.3407$, and $-0.895$, indicating that higher accept rates result in higher likelihoods. 

\subsection{Number of HMC chains}

We can improve the coverage of the posterior distribution by running multiple independent chains of HMC.
Effectively, each chain is an independent run of the procedure using a different random initialization.
Then, we combine the samples from the different chains.
The computational requirements of running multiple chains are hence proportional to the number of chains.

We report the Bayesian model average performance as a function of the number of chains in \autoref{fig:tau_ablation}(b).
Holding compute budget fixed, using two or three chains is only slightly better than using one chain. This result notably 
shows that HMC is relatively unobstructed by energy barriers in the posterior surface 
that would otherwise require multiple chains to overcome. We explore this result further in \autoref{sec:mixing}.

\section{How well does HMC mix?}
\label{sec:mixing}

The primary goal of our paper is to construct accurate samples from the posterior, and use them
to understand the properties of Bayesian neural networks better.
In this section we consider several diagnostics to evaluate whether our HMC sampler has converged, and
discuss their implications to the posterior geometry. We consider mixing in both \emph{weight space} and 
\emph{function space}. A distribution over weights $w$ combined with a neural network architecture 
$f(x,w)$ induces a distribution over functions $f(x)$. Ultimately, since we are using functions to make 
predictions, we care mostly about mixing in function space.

\begin{mybox}
    \textbf{Summary}: HMC is able to mix surprisingly well in function space, and better 
    than in parameter space.
    Geometrically, HMC is able to explore connected basins of the posterior with high
    functional diversity.
\end{mybox}

\subsection{$\hat R$ diagnostics}
\label{sec:r_hat}

We apply the classic \citet{gelman1992inference} ``$\hat R$'' potential-scale-reduction diagnostic to our HMC runs. Given two or more chains, $\hat R$
estimates the ratio between the between-chain variance (i.e., the variance estimated by pooling samples from all chains) and the average within-chain variance (i.e., the variances estimated from each chain independently).
The intuition is that, if the chains are stuck in isolated regions, then combining samples from multiple chains will yield greater diversity than taking samples from a single chain. For the precise mathematical definition of $\hat R$, please see the \autoref{sec:app_rhat}.

\begin{figure}[t]
    \centering
    \begin{tabular}{ccc}
    & \hspace{.5cm}{\scriptsize \textbf{Weight Space}} & {\scriptsize \textbf{Function Space}} 
    \\
    \rotatebox{90}{\scriptsize \quad\quad\quad \textbf{CIFAR-10}} & 
    \hspace{-0.3cm}\includegraphics[height=0.12\textwidth]{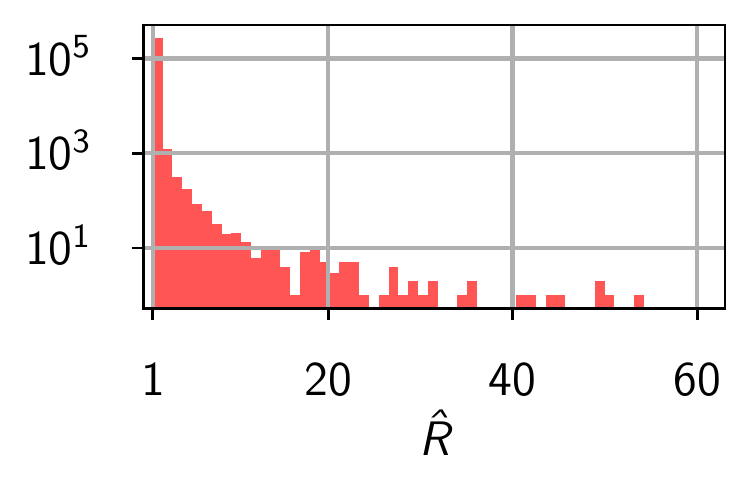} & 
    \hspace{-0.3cm}\includegraphics[height=0.12\textwidth]{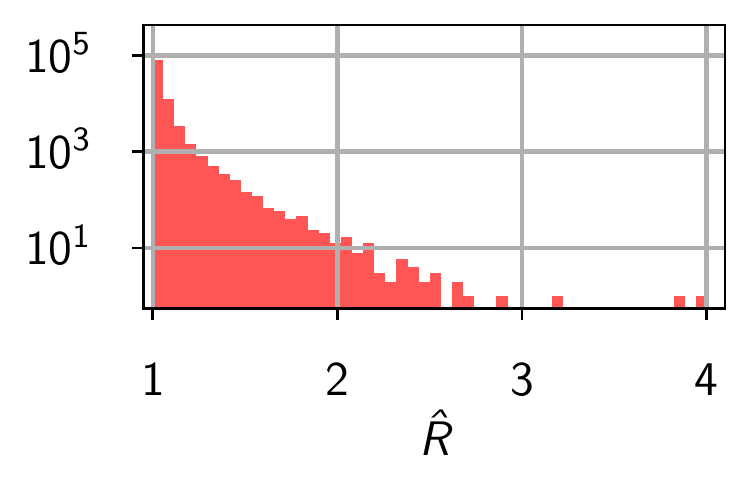} 
    \\[-0.2cm]
    \rotatebox{90}{\scriptsize \quad\quad\quad~~ \textbf{IMDB}} & 
    \hspace{-0.3cm}\includegraphics[height=0.12\textwidth]{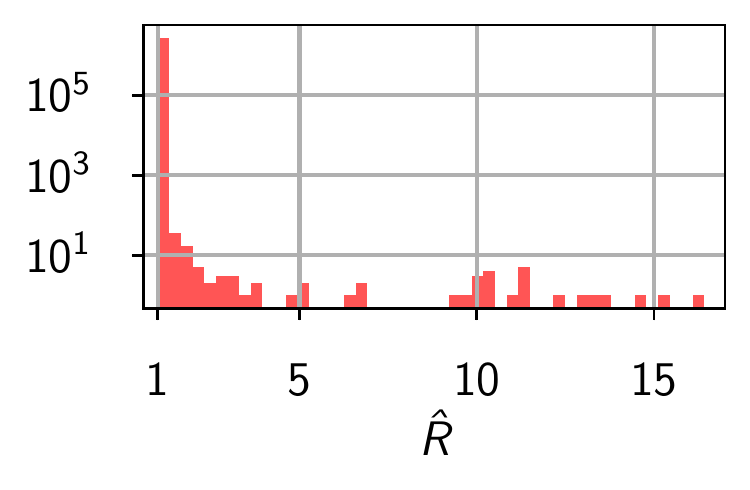} &
    \hspace{-0.3cm}\includegraphics[height=0.12\textwidth]{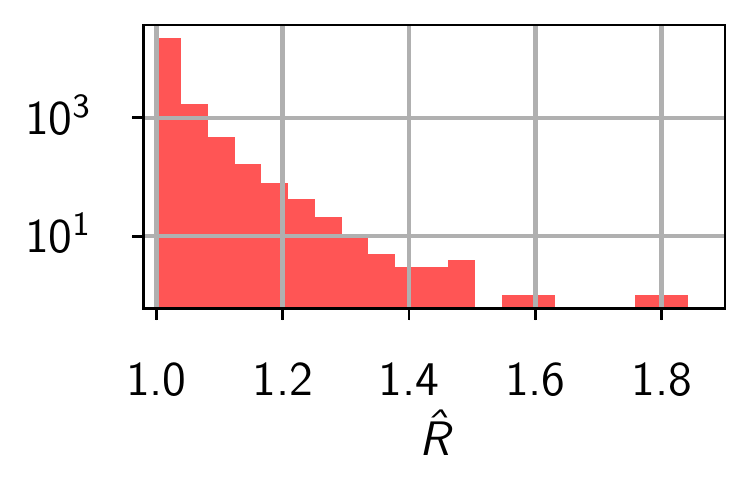}\\[-0.3cm]
    \end{tabular}
	\caption{
	    \textbf{Log-scale histograms of $\hat R$ convergence diagnostics.}
	    Function-space $\hat R$s are computed on the test-set softmax predictions of the classifiers and weight-space $\hat R$s are computed on the raw weights.
	    About $91\%$ of CIFAR-10 and $98\%$ of IMDB posterior-predictive probabilities get an $\hat R$ less than 1.1. Most weight-space $\hat R$ values are quite small, but enough parameters have very large $\hat R$s to make it clear that the chains are sampling from different distributions in weight space.
	}
	\label{fig:r_hats}
\end{figure}

We compute $\hat R$ using TensorFlow Probability's implementation\footnote{\href{https://www.tensorflow.org/probability/api_docs/python/tfp/mcmc/potential_scale_reduction}{tfp.mcmc.potential\_scale\_reduction}} \citep{lao2020tfp} for both the weights and the test-set softmax predictions on CIFAR-10 with ResNet-20-FRN and on IMDB with CNN-LSTM. 
We report the results in \autoref{fig:r_hats}.
We observe that on both IMDB and CIFAR, the bulk of the function-space $\hat R$ values is concentrated near $1$, 
meaning intuitively that a single chain can capture the diversity of predictions on most of the test data points nearly as well as multiple chains.
The mixing is especially good on the IMDB dataset, where only $2\%$ of inputs correspond to $\hat R$ larger than $1.1$.
In \autoref{sec:app_synthreg} we apply HMC to a synthetic regression problem and show that HMC can indeed mix in the prediction space: 
different HMC chains provide very similar predictions.

In weight space, although most parameters show no evidence of poor mixing, some have very large $\hat R$s, indicating that there are directions in which the chains fail to mix. 

\paragraph{Implications for the Posterior Geometry.}\quad
The fact that a single HMC chain is able to mix well in function space (aka prediction space) suggests that the posterior contains
connected regions which correspond to high functional diversity. 
Indeed, a single HMC chain is extremely unlikely to jump between isolated modes, but appears able to produce samples with diverse predictions.
Moreover, HMC is able to navigate these regions effeciently. 
Prior work on \textit{mode connectivity} \citep{garipov2018, draxler2018essentially}  has shown that there exist paths of high density 
connecting different modes of the posterior.
Our observations suggest a stronger version of mode connectivity: not only do mode-connecting paths exist between functionally diverse modes, but
also at least some of these paths can be leveraged by Monte Carlo methods to efficiently explore the posterior.

\subsection{Posterior density visualizations}
\label{sec:surface_visualizations}

\begin{figure}[t]
    \centering
    \begin{tabular}{ccc}
    \hspace{-.5cm}
    \includegraphics[height=0.15\textwidth]{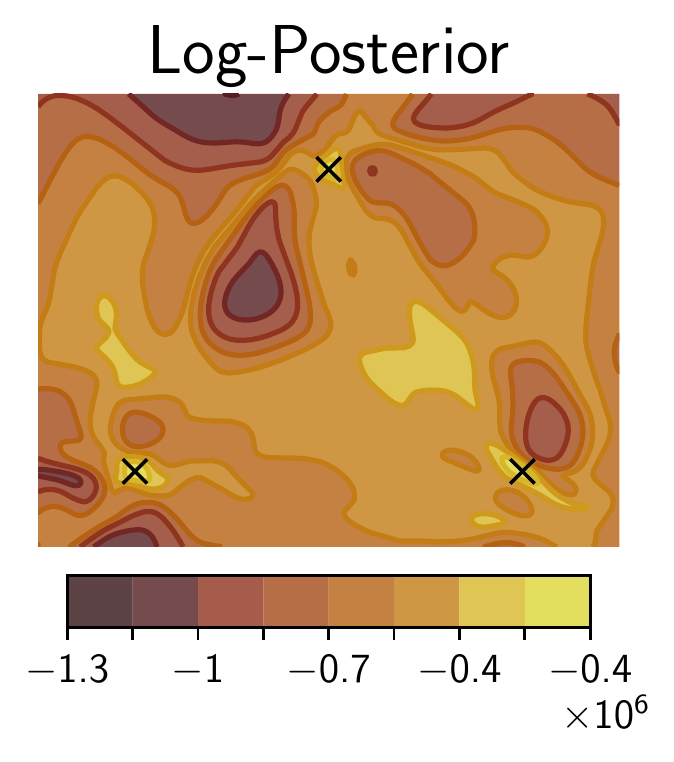}&
    \hspace{-.2cm}
    \includegraphics[height=0.15\textwidth]{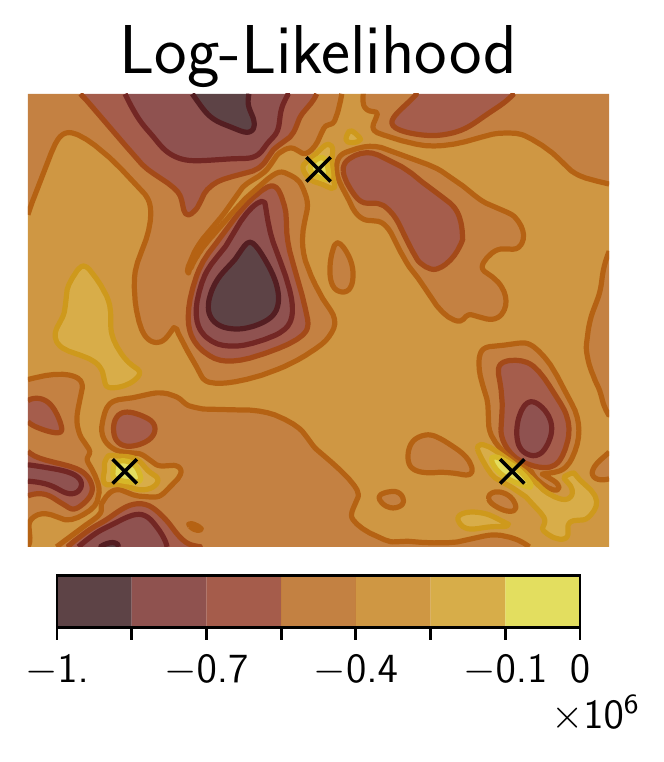}&
    \hspace{-.2cm}
    \includegraphics[height=0.15\textwidth]{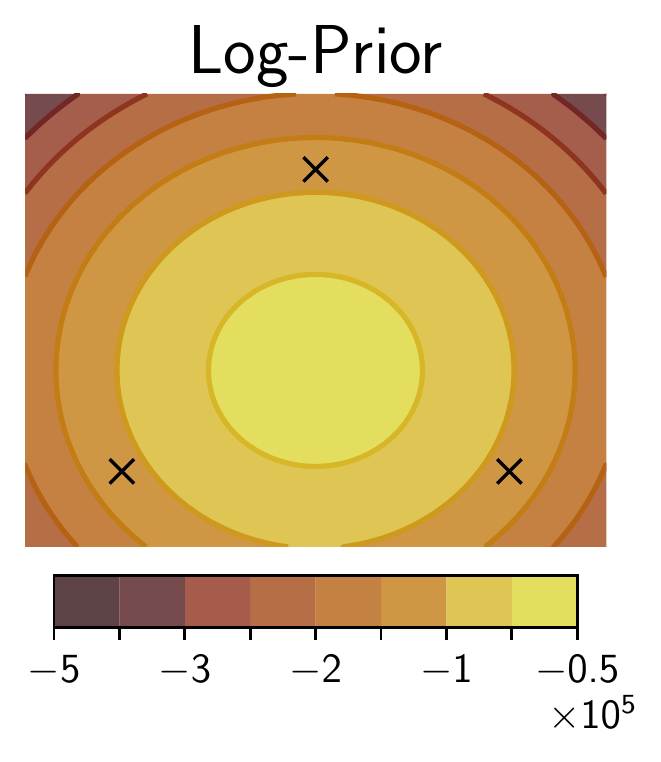}\\[-0.2cm]
    \multicolumn{3}{c}{\small (a) Same chain}\\
    \hspace{-.5cm}
    \includegraphics[height=0.15\textwidth]{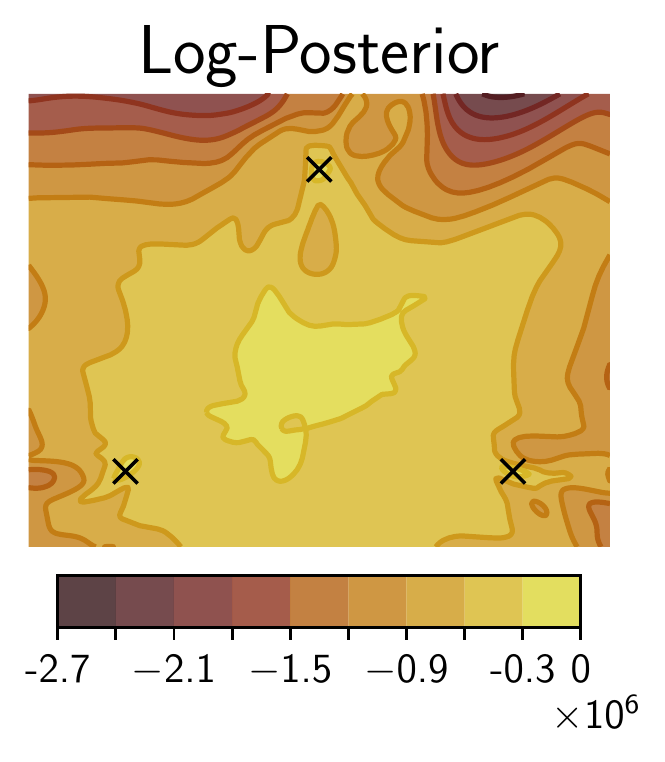}&
    \hspace{-.2cm}
    \includegraphics[height=0.15\textwidth]{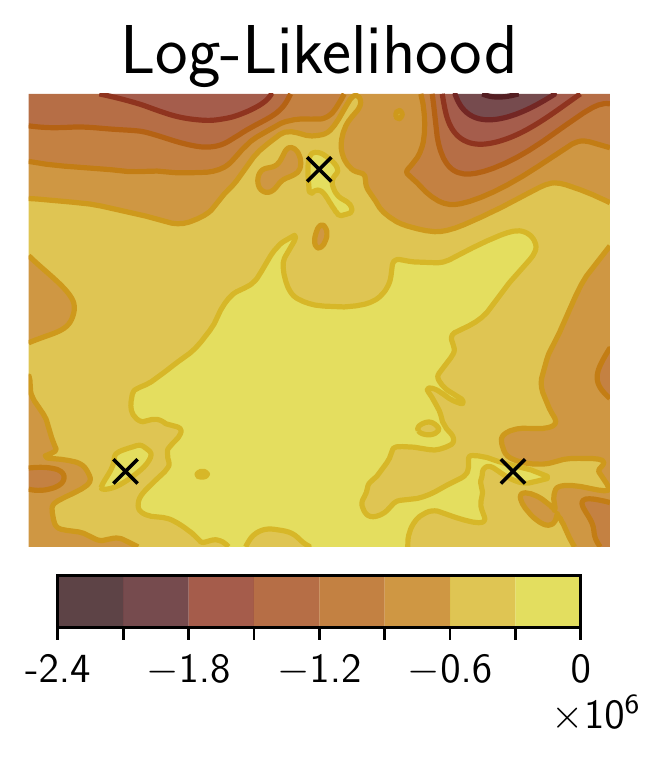}&
    \hspace{-.2cm}
    \includegraphics[height=0.15\textwidth]{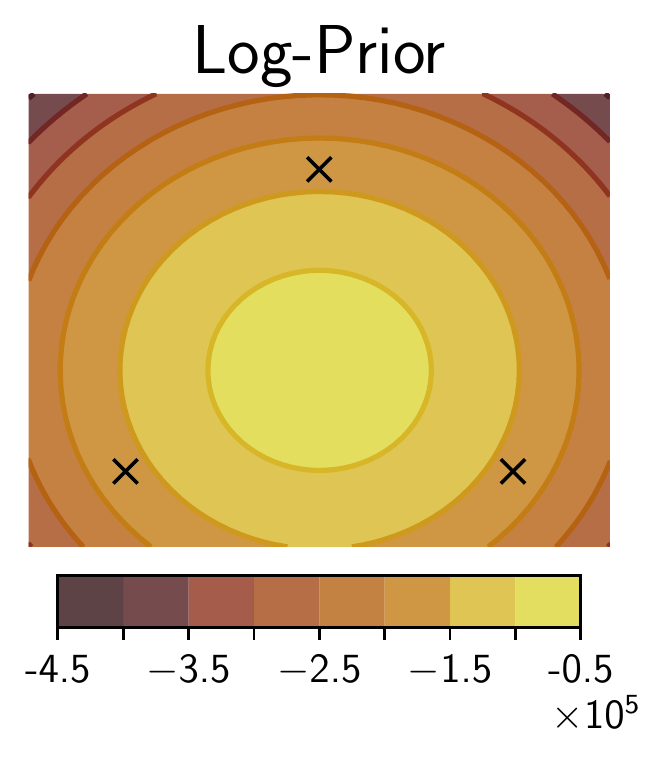}\\[-0.2cm]
    \multicolumn{3}{c}{\small (b) Independent chains}\\
    \end{tabular}
	\caption{
	    \textbf{Posterior density visualization.}
        Visualizations of posterior log-density, log-likelihood and log-prior in the two-dimensional subspace
        of the parameter space spanned by three HMC samples from \textbf{(a)} the same chain and \textbf{(b)} three independent chains.
        Each HMC chain explores a region of high posterior density of a complex non-convex shape, that appears multi-modal
        in the presented cross-sections.
	}
	\label{fig:posterior_density}
\end{figure}

To further investigate how HMC is able to explore the posterior over the weights, we visualize a cross-section of the posterior density in subspaces of the parameter space containing the samples.
Following \citet{garipov2018}, we study two-dimensional subspaces of the parameter space of the form
\begin{equation}
    \label{eq:subspace}
    \mathcal{S} = \{w \vert w = w_1 \cdot a + w_2 \cdot b + w_3 \cdot (1 - a - b)\}.
\end{equation}
$\mathcal{S}$ is the unique two-dimensional affine subspace (plane) of the parameter space that includes parameter vectors $w_1$, $w_2$ and $w_3$.

In \autoref{fig:posterior_density}(a) we visualize the posterior log-density, log-likelihood and log-prior density of a ResNet-20-FRN on CIFAR-10.
For the visualization, we use the subspace $\mathcal S$ defined by the parameter vectors $w_{1}, w_{51}$ and $w_{101}$,
the samples produced by HMC at iterations $1, 51$ and $101$ after burn-in respectively.
We observe that HMC is able to navigate complex geometry: the samples fall in three seemingly isolated modes in our two-dimensional cross-section of the posterior.
In other words, HMC samples from a single chain are not restricted to any specific convex Gaussian-like mode, and instead explore a region of high posterior density of a complex shape in the parameter space. We note that popular approximate inference procedures, such as variational methods, and Laplace approximations, are typically constrained to unimodal Gaussian approximations to the posterior, which we indeed expect to miss a large space of compelling solutions in the posterior.

In \autoref{fig:posterior_density}(b) we provide a visualization for the samples produced by $3$ different HMC chains at iteration $51$ after burn-in.
Comparing the visualizations for samples from the same chain and samples from independent chains in \autoref{fig:posterior_density}, we see that the shapes of
the posterior surfaces are different, with the latter appearing more regular and symmetric.
The qualitative differences between (a) and (b) suggest that while each HMC chain is able to navigate the posterior geometry
the chains do not mix perfectly in the weight space, confirming our results in \autoref{sec:r_hat}.

In \autoref{sec:app_surface_visualizations} we provide additional details on our visualization procedure and visualizations using the CNN-LSTM on IMDB.

\subsection{Convegence of the HMC chains}

\begin{figure}[t]
    \centering
    \hspace{-0.3cm}
    \begin{tabular}{cccc}
    \hspace{-.5cm}
    \rotatebox{90}{\scriptsize \quad\quad~~~\textbf{CIFAR-10}} &
    \hspace{-.5cm}
    \includegraphics[height=0.11\textwidth]{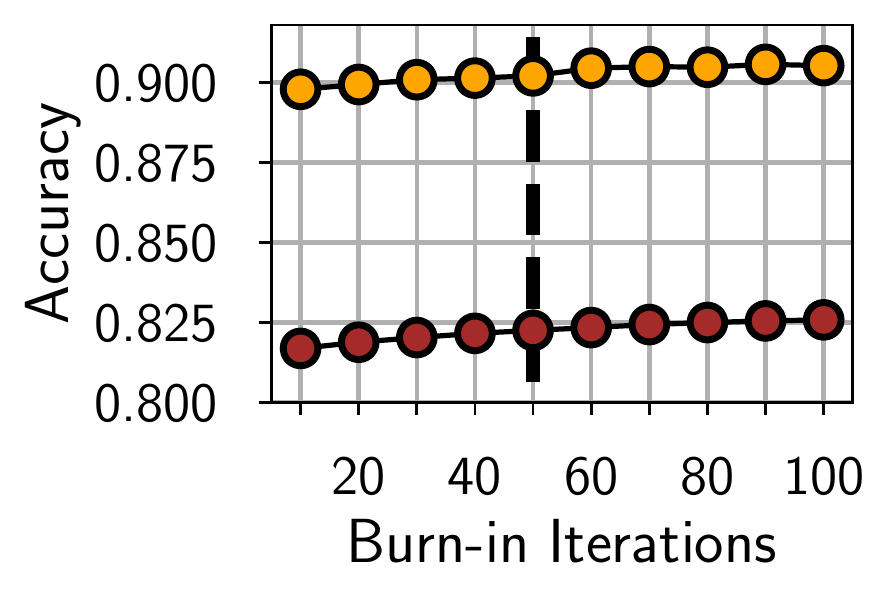} &
    \hspace{-.5cm}
    \includegraphics[height=0.11\textwidth]{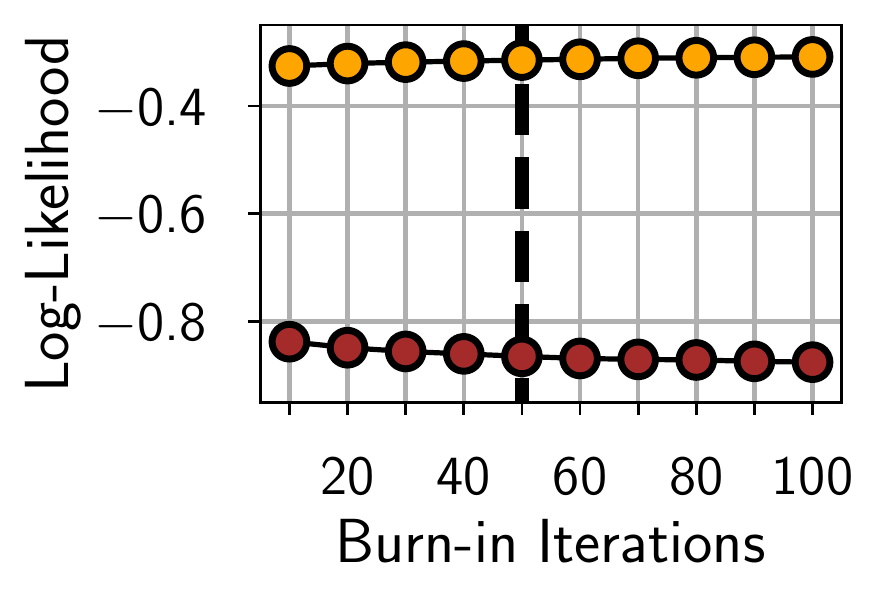} &
    \hspace{-.5cm}
    \includegraphics[height=0.11\textwidth]{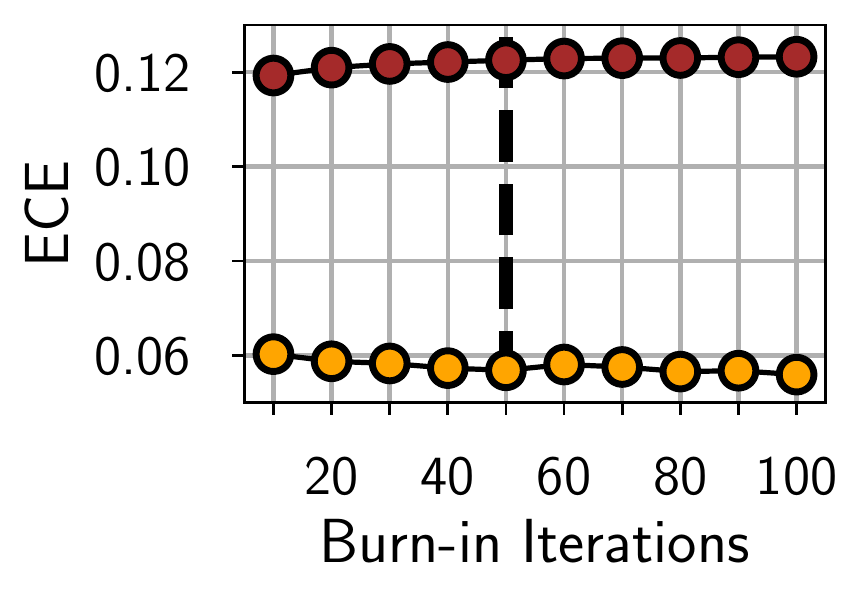}
    \\[-0.cm]
    \hspace{-.5cm}
    \rotatebox{90}{\scriptsize \quad\quad\quad~~~\textbf{IMDB}} &
    \hspace{-.5cm}
    \includegraphics[height=0.11\textwidth]{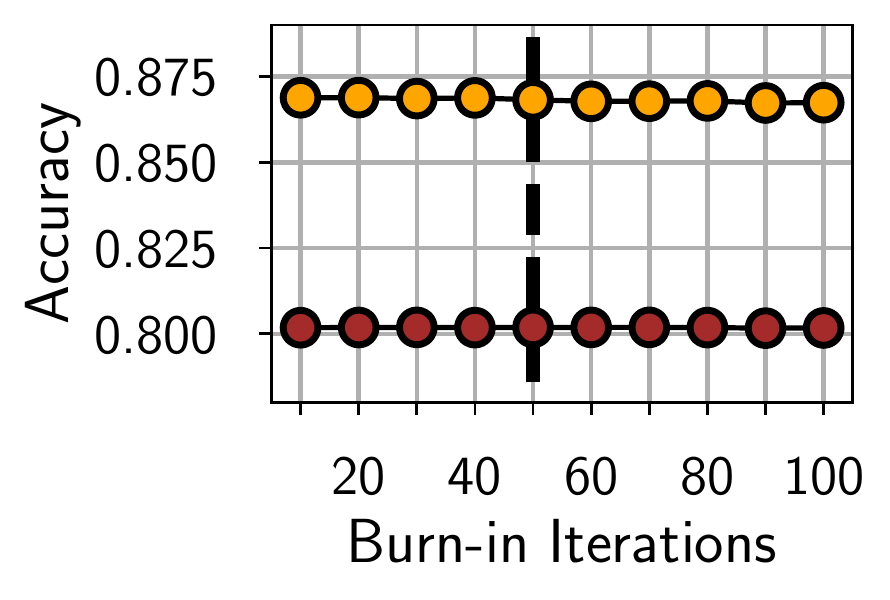} &
    \hspace{-.5cm}
    \includegraphics[height=0.11\textwidth]{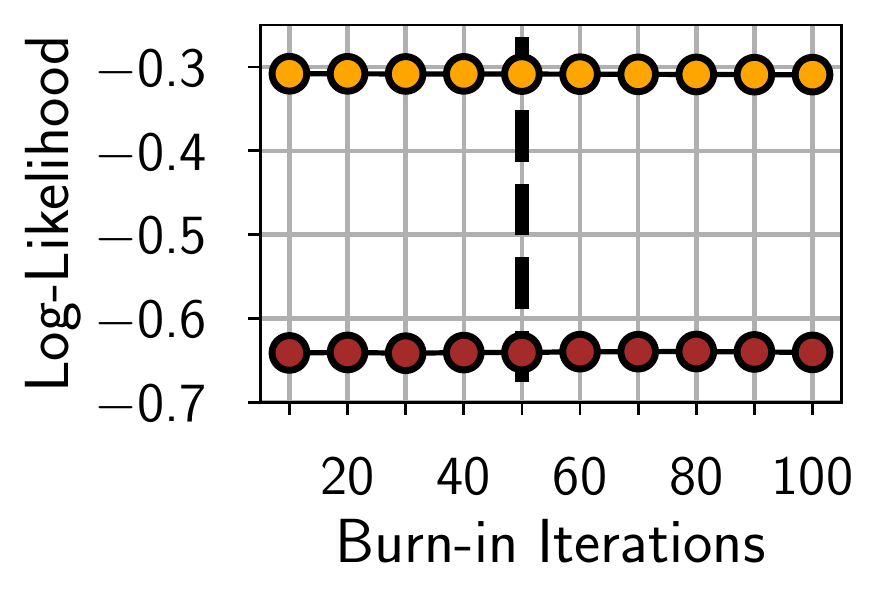} &
    \hspace{-.5cm}
    \includegraphics[height=0.11\textwidth]{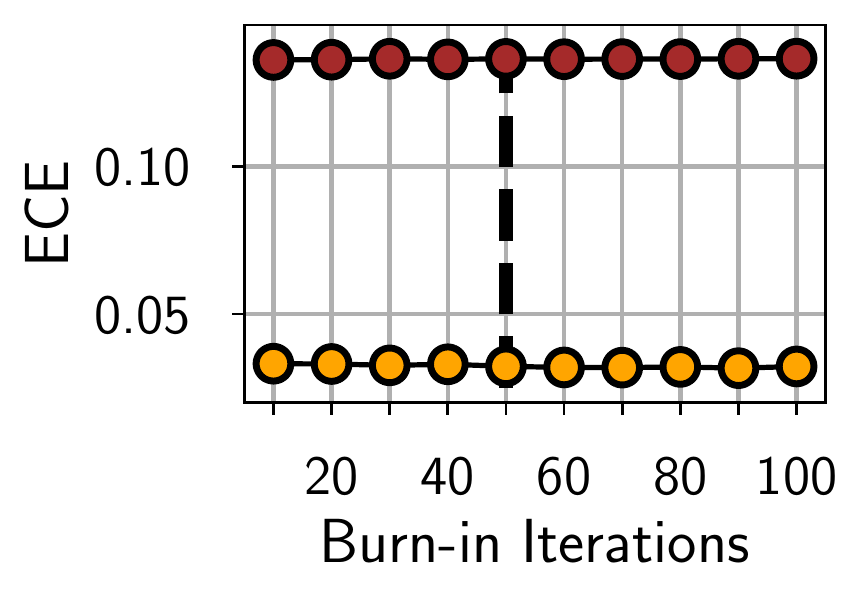}
    \\[-0.3cm]
    &\multicolumn{3}{c}{\includegraphics[height=0.09\textwidth]{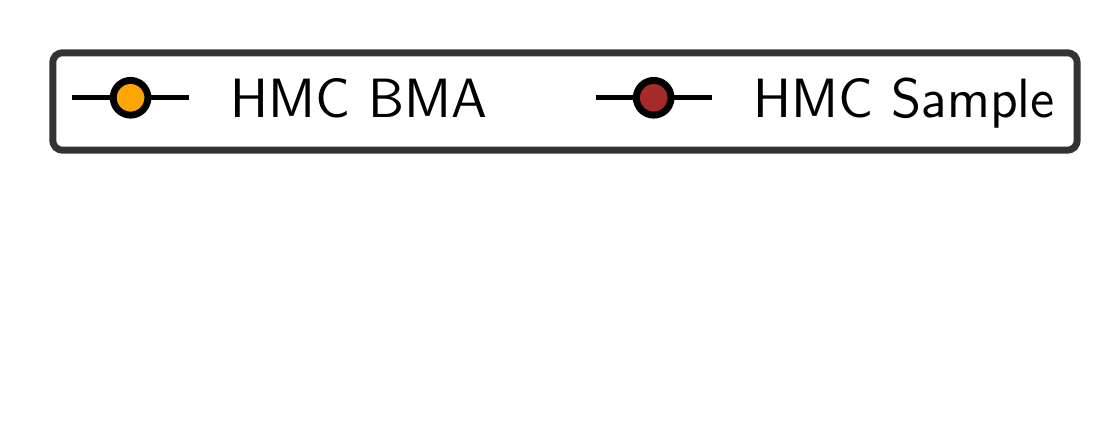}}\\[-0.8cm]
    \vspace{-1.cm}
    \end{tabular}
	\caption{
	    \textbf{HMC convergence.}
	    The performance of an individual HMC sample and a BMA ensemble of $100$ samples from each one of $3$ HMC chains after the burn-in as a function of burn-in length.
	    The dashed line indicates the burn-in length of $50$ that we used in the main experiments in this paper.
	    We use ResNet-20-FRN on CIFAR-10 and CNN-LSTM on IMDB.
	    On IMDB, there is no visible dependence of the results on the burn-in length;
	    on CIFAR-10, there is a weak trend that slows down over time.
	}
	\label{fig:burnin_ablation}
\end{figure}

As another diagnostic, we look at the convergence of the performance of HMC BMA estimates and individual samples as 
a function of the length of the burn-in period. 
For a converged chain, the performance of the BMA and individual samples should be stationary not show any visible trends after a sufficiently long burn-in.
We use the samples from $3$ HMC chains, and evaluate performance of the ensemble of the first $100$ HMC samples in each chain after
discarding the first $n_{bi}$ samples, where $n_{bi}$ is the length of the burn-in.
Additionally, we evaluate the performance of the individual HMC samples after $n_{bi}$ iterations in each of the chains.

We report the results for ResNet-20-FRN on CIFAR-10 and CNN-LSTM on IMDB in \autoref{fig:burnin_ablation}.
On IMDB, there is no visible trend in performance, so a burn-in of just $10$ HMC iterations should be sufficient.
On CIFAR-10, we observe a slowly rising trend that saturates at about $50$ iterations, indicating that a longer burn-in period is needed compared to IMDB.
We therefore use a burn-in period of $50$ HMC iterations on both CIFAR and IMDB for the remainder of the paper.

\begin{figure}[t]
    \centering
    \hspace{-0.1cm}
    \includegraphics[height=0.3\textwidth]{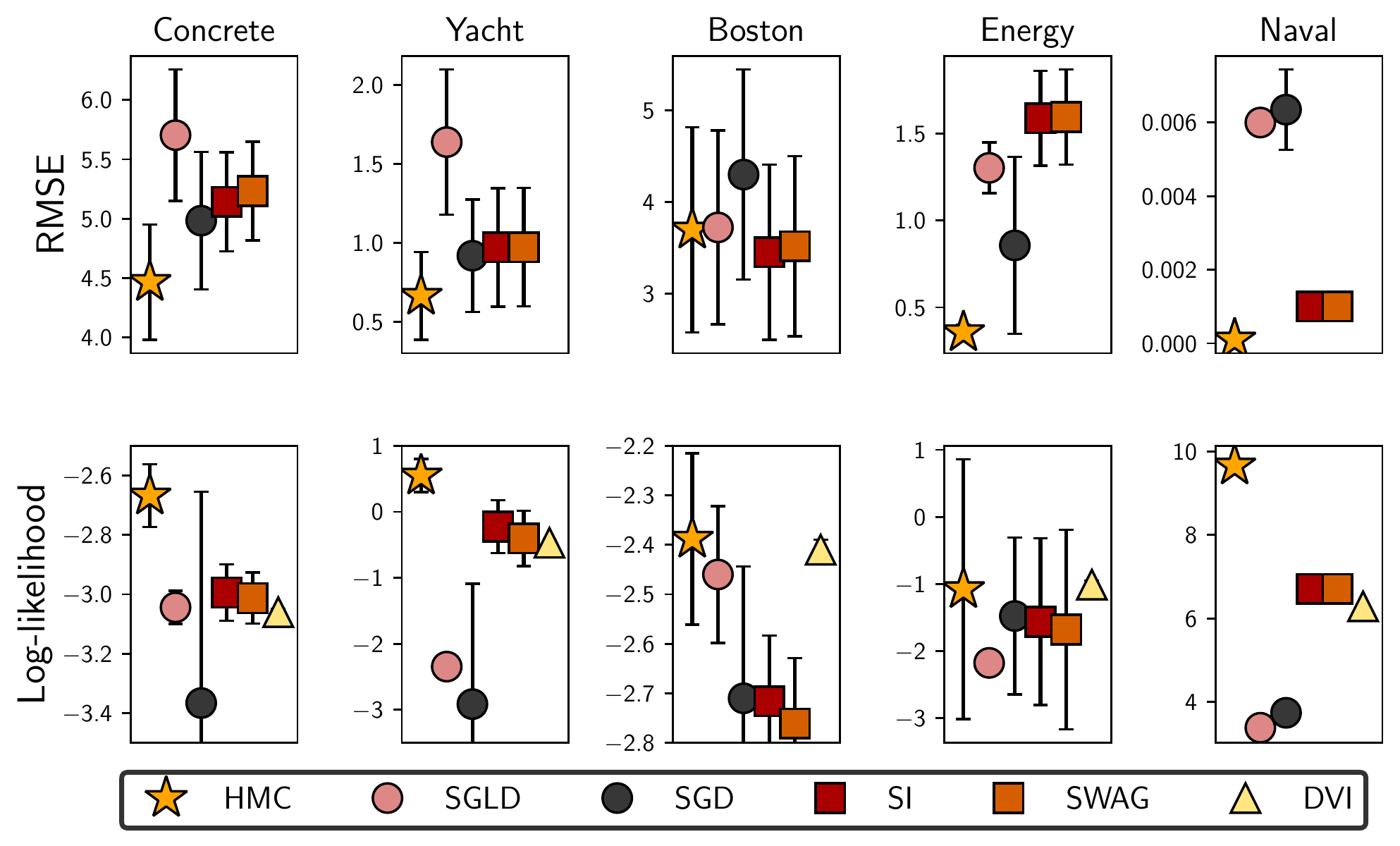}
    \vspace{-0.3cm}
	\caption{
	    \textbf{UCI regression datasets.}
	    Performance of Hamiltonian Monte Carlo (HMC), stochastic gradient Langevin dynamics (SGLD),
	    stochastic gradient descent (SGD), subspace inference (SI) \citep{izmailov2019subspace},
	    SWAG \citep{maddoxfast2019} and deterministic variational inference \citep[DVI; ][]{wu2018deterministic}.
	    We use a fully-connected architecture with a single hidden layer of $50$ neurons.
	    The results reported for each method are mean and standard deviation computed over $20$ random
	    train-test splits of the dataset.
	    For SI, SWAG and DVI we report the results presented in \citet{izmailov2019subspace}.
	    \textbf{Top:} test root-mean-squared error.
	    \textbf{Bottom:} test log-likelihood.
	    HMC performs on par with or better than all other baselines in each experiment, often providing a significant
	    improvement.
	}
	\label{fig:uci}
\end{figure}
\begin{figure}[t]
    \centering
    \begin{tabular}{cc}
      \includegraphics[height=0.274\textwidth]{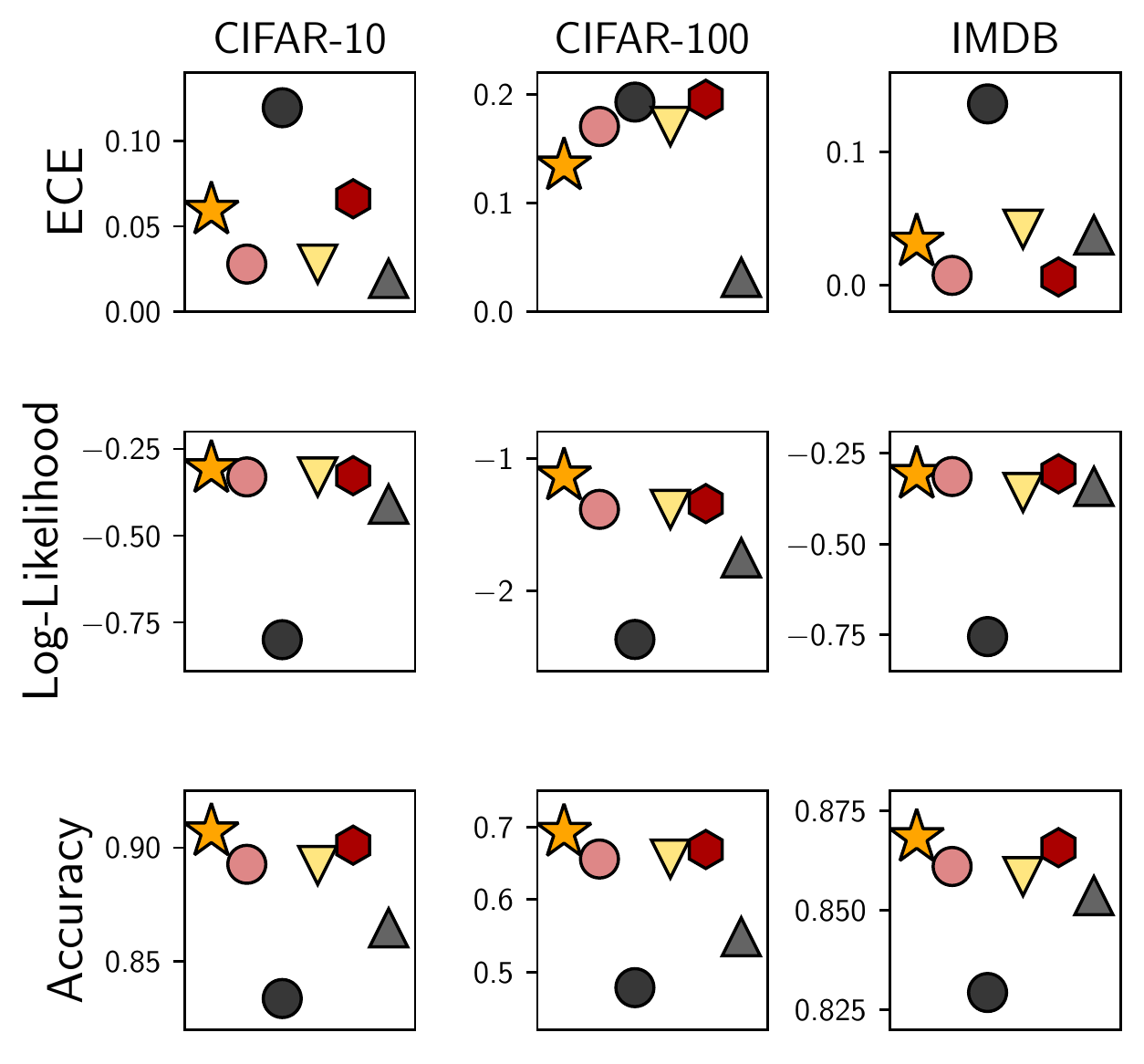} &
      \includegraphics[width=0.145\textwidth]{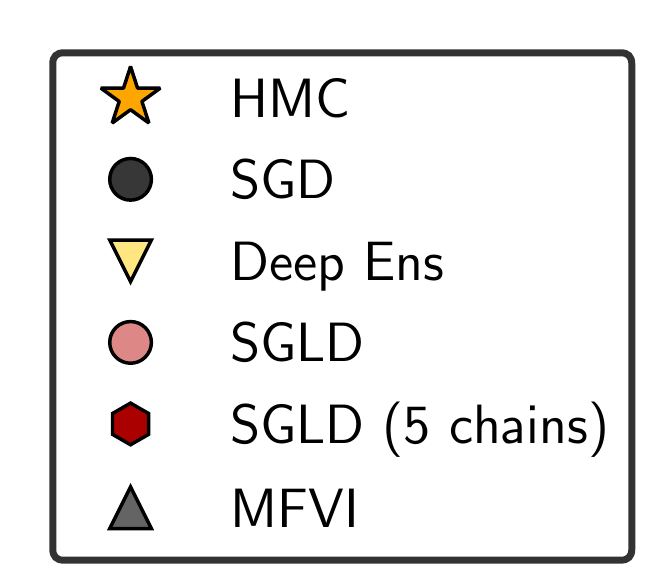}
    \end{tabular}
	\caption{
	    \textbf{Image and text classification.}
	    Performance of Hamiltonian Monte Carlo (HMC), stochastic gradient Langevin dynamics (SGLD) with 1 and 5 chains,
	    mean field variational inference (MFVI),
	    stochastic gradient descent (SGD), and deep ensembles.
	    We use ResNet-20-FRN on CIFAR datasets, and CNN-LSTM on IMDB.
	    Bayesian neural networks via HMC outperform all baselines on all datasets in terms of accuracy and log-likelihood.
	    On ECE, the methods perform comparably.
	}
	\label{fig:cifar_imdb}
\end{figure}

\section{Evaluating Bayesian neural networks}
\label{sec:bnn_evaluation}

Now that we have a procedure for effective HMC sampling, we are primed to explore exciting questions about the 
fundamental behaviour of Bayesian neural networks, such as the role of tempering, the prior over parameters, 
generalization performance, and robustness to covariate shift.
In this section we evaluate Bayesian neural networks in various problems using our implementation of HMC.
Throughout the experiments, we use posterior temperature $T=1$.

We emphasize that the main goal of our paper and this section in particular is to \emph{understand} the behaviour of true
Bayesian neural network posteriors using HMC as a precise tool, and \emph{not} to argue for HMC as a practical method for 
Bayesian deep learning.

\begin{mybox}
    \textbf{Summary:} Bayesian neural networks achieve strong results outperforming even large deep ensembles in a range of evaluations.
    Surprisingly, however, BNNs are \textit{less} robust to distribution shift than conventionally-trained models.
\end{mybox}

\subsection{Regression on UCI datasets}

Bayesian deep learning methods are often evaluated on small-scale regression problems using fully connected networks
\citep[e.g., ][]{wu2018deterministic, izmailov2019subspace, maddoxfast2019}.
Following these works, we evaluate Bayesian neural networks using HMC on five UCI regression datasets:
\textit{Concrete}, \textit{Yacht}, \textit{Boston}, \textit{Energy} and \textit{Naval}.
For each of these datasets, we construct $20$ random $90$-to-$10$ train-test splits and report the mean
and standard deviation of performance over the splits.
We use a fully connected neural network with a single hidden layer of size $50$ and $2$ outputs
representing the predictive mean and standard deviation.
For HMC we used a single chain with $10$ burn-in iterations and $90$ iterations of sampling.
For more details, please see \autoref{sec:app_hypers}.

We report the results in \autoref{fig:uci}.
HMC typically outperforms all the baselines, often by a significant margin, both in test RMSE and log-likelihood.
On the \textit{Boston} dataset, HMC achieves a slightly higher average RMSE compared to the subspace inference and SWAG
\citep{izmailov2019subspace, maddoxfast2019} but outperforms both these methods significantly in terms of
log-likelihood.

\begin{table}[t]
\begin{center}
\begin{small}
\begin{sc}
\begin{tabular}{lllll}
\toprule
& \multicolumn{4}{c}{AUC-ROC} \\
\cline{2-5}\\[-0.2cm]
OOD Dataset & HMC  & DE & ODIN & Mahal. \\
\midrule
CIFAR-100   & 0.857  & 0.853  & 0.858 & \textbf{0.882}        \\
SVHN        & 0.8814 & 0.8529  & 0.967 & \textbf{0.991}       \\
\bottomrule
\end{tabular}
\end{sc}
\end{small}
\end{center}
\caption{\textbf{Out-of-distribution detection.}
We use a ResNet-20-FRN model trained on CIFAR-10 to detect out-of-distribution data coming from SVHN or CIFAR-100.
We report the results for HMC, deep ensembles and specialized ODIN \citep{liang2017enhancing} and Mahalanobis \citep{lee2018simple} methods.
We report the AUC-ROC score (higher is better) evaluating the ability of each method to distinguish between in-distribution and OOD data.
The predictive uncertainty from Bayesian neural networks allows us to detect OOD inputs:
HMC outpefroms deep ensembles on both datasets.
Furthermore, HMC is competitive with ODIN on the harder near-OOD task of detecting CIFAR-100 images,
but underperforms on the easier far-OOD task of detecting SVHN images.
}
\label{tab:ood}
\end{table}

\subsection{Image Classification on CIFAR}
\label{sec:image_classification}

Next, we evaluate Bayesian neural networks using HMC on image classification problems.
We use the ResNet-20-FRN architecture on CIFAR-10 and CIFAR-100.
We picked a random subset of 40960 of the 50000 images for each of the datasets to be able to evenly shard the data across the TPU devices;
we use the same subset for both HMC and the baselines.
We run 3 HMC chains using step size $10^{-5}$ and a prior variance of $1/5$, resulting in 70,248 leapfrog steps per sample.
In each chain we discard the first 50 samples as burn-in, and then draw 240 samples (720 in total for 3 chains)\footnote{
In total, on CIFAR-10 our HMC run requires as many computations as \textit{over 60 million epochs} of standard SGD training}.
For SGLD, we use a single chain with $1000$ burn-in epochs and $9000$ epochs of sampling producing $900$ samples;
we also report the performance of an ensemble of $5$ independent SGLD chains.
Next, we report the performance of a mean field variational inference (MFVI) solution; 
we initialize the mean of MFVI with a solution pre-trained with SGD and use an ensemble of $50$ samples from the VI posterior at evaluation.
Finally, we report the performance of a single SGD solution and a deep ensemble of $50$ models.
For more details, see \autoref{sec:app_hypers}.

We report the results in \autoref{fig:cifar_imdb}. 
Bayesian neural networks outperform all baselines in terms of accuracy and log-likelihood on both datasets.
In terms of ECE, SGD provides the worst results across the board, and the rest of the methods are competitive;
MFVI is particularly well-calibrated on CIFAR-100.

\textbf{Out-of-distribution detection.}\quad
Bayesian deep learning methods are often evaluated on out-of-distribution detection.
In the \autoref{tab:ood} we report the performance of HMC-based Bayesian neural network on out-of-distribution (OOD) detection.
To detect OOD data, we use the level of predicted confidence (value of the softmax class probability for the predicted class)
from the HMC ensemble, measuring the area under the receiving operator characteristic curve (AUC-ROC).
We train the methods on CIFAR-10 and use CIFAR-100 and SVHN as OOD data sources.
We find that BNNs perform competitively with the specialized ODIN method in the challenging near-OOD detection setting 
(i.e. when the OOD data distribution is similar to the training data) of CIFAR-100, while 
underperforming in the easier far-OOD setting on SVHN relative to the baselines \citep{liang2017enhancing, lee2018simple}.

\begin{figure}[t]
    \centering
    \hspace{-0.1cm}
    \begin{tabular}{cc}
    \hspace{-0.5cm}
    \includegraphics[height=0.175\textwidth]{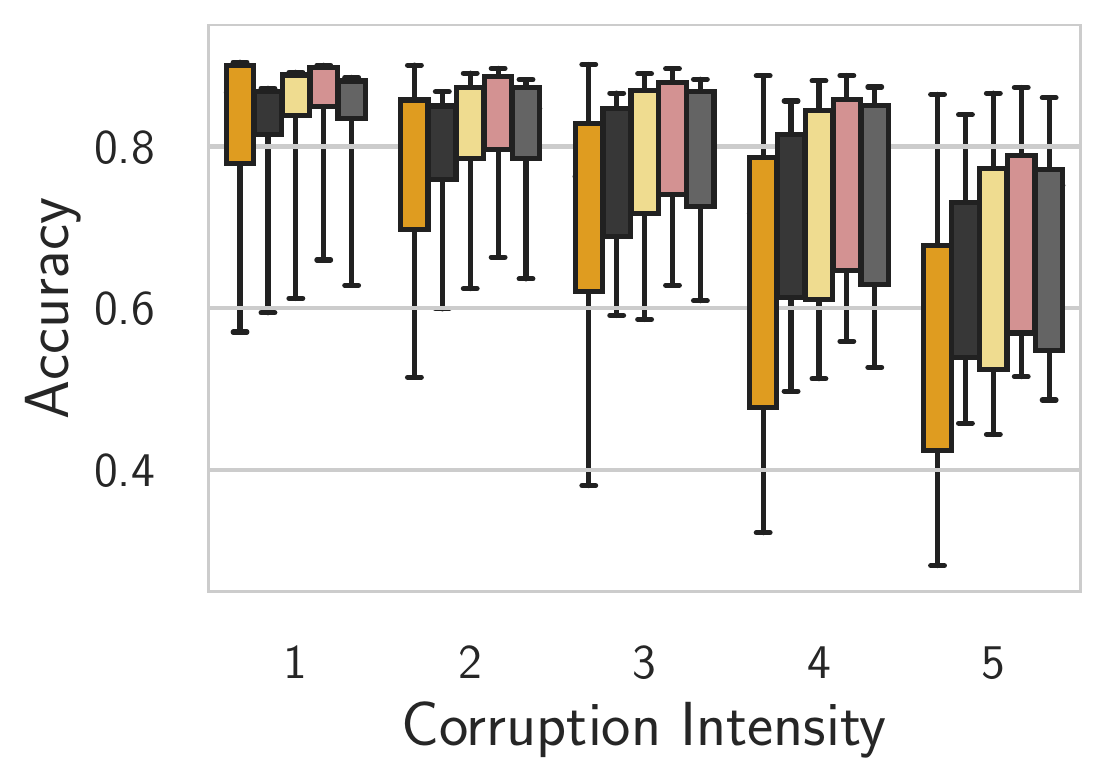} &
    \hspace{-0.5cm}
    \includegraphics[height=0.175\textwidth]{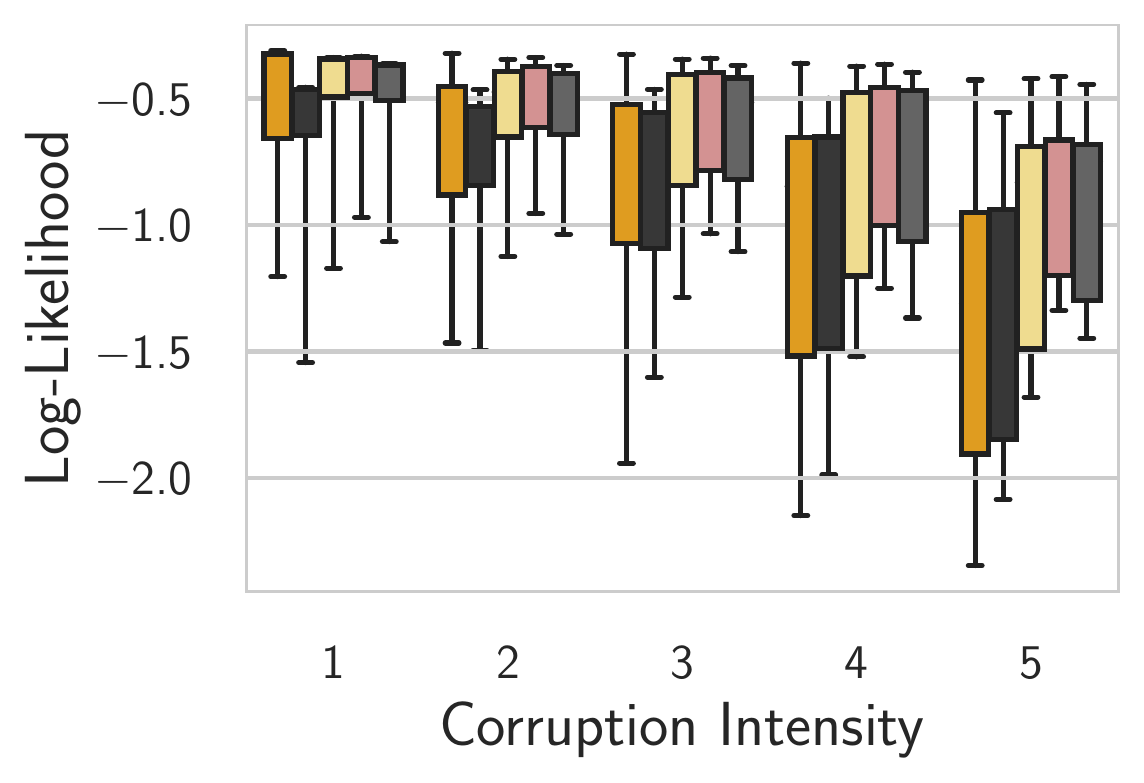} \\
    \multicolumn{2}{c}{\includegraphics[width=0.45\textwidth]{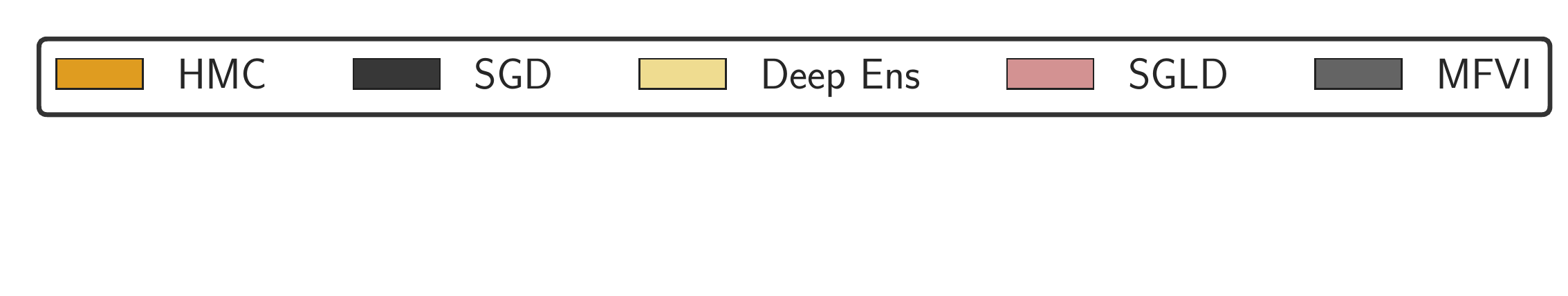}}\\[-1.cm]
    \end{tabular}
	\caption{
	    \textbf{Evaluation on CIFAR-10-C.}
	    Accuracy and log-likelihood of HMC, SGD, deep ensembles, SGLD and MFVI on a distribution shift task, where
	    the CIFAR-10 test set is corrupted in 16 different ways at various intensities on the scale of 1 to 5. 
	    We use the ResNet-20-FRN architecture.
	    Boxes capture the quartiles of performance over each corruption, with
	    the whiskers indicating the minimum and maximum. 
	    HMC is surprisingly the worst of the considered methods: even a single SGD solution provides better OOD robustness.
	}
	\label{fig:cifar10c}
\end{figure}

\textbf{Robustness to distribution shift}
\quad 
Bayesian methods are often specifically applied to covariate shift problems \citep{ovadia2019can, wilson2020bayesian, dusenberry2020efficient}.
We evaluate the the performance of HMC and Deep Ensemble-based Bayesian neural networks on the CIFAR-10-C dataset \citep{hendrycks2019benchmarking}, 
which applies a set of corruptions to CIFAR-10 with varying intensities.
Mimicking the setup in \citet{ovadia2019can}, we use the same 16 corruptions, evaluating the performance at all intensities.
We report the results in \autoref{fig:cifar10c}.
Surprisingly, we find that Deep Ensembles and SGLD are consistently more robust to distribution shift than HMC-based BNNs.
For high corruption intensities, even a single SGD model outperforms the HMC ensemble.

In \autoref{sec:app_robustness} we provide further exploration of this effect, where we see HMC samples are significantly
less robust to many types of noise compared to conventionally-trained SGD models.
Interestingly, the performance of HMC-based BNNs under data corruption can be significantly improved by using posterior tempering.

\subsection{Language Classification on IMDB}

We use a CNN-LSTM architecture on the IMDB binary text classification dataset.
In \autoref{fig:cifar_imdb} we report the results for HMC and the same baselines as in \autoref{sec:image_classification}.
We use HMC with a step size of $10^{-5}$ and a prior variance of $1/40$,
resulting in 24,836 leapfrog steps per sample. 
We run $3$ chains, burning-in for 50
samples, and drawing 400 samples per chain (1,200 total).
For more details on the hyperparameters, please see \autoref{sec:app_hypers}.
Analogously to the image classification experiments, HMC outperforms the baselines on accuracy and log-likelihood and provides competitive performance on ECE.

\section{Do we need cold posteriors?}
\label{sec:cold_posteriors}

Multiple works have considered tempering the posterior in Bayesian neural networks \citep[e.g.][]{wenzel2020good, wilson2020bayesian, zhang2019cyclical, ashukha2020pitfalls, aitchison2020statistical}.
Specifically, we can consider a distribution 
\begin{equation}
    \label{eq:posterior_temp}
    p_T(w \vert \mathcal D) \propto \big ( p(\mathcal D \vert w) \cdot p(w) \big )^{1 / T},
\end{equation}
where $w$ are the parameters of the network, $\mathcal D$ is the training dataset, $p(\mathcal D \vert w)$ is the likelihood of $\mathcal D$ for the network with parameters $w$ and $T$ is the \textit{temperature}.
Note that at temperature $T = 1$, $p_T$ corresponds to the standard Bayesian posterior over the parameters of the network.
Temperatures $T < 1$ correspond to \textit{cold posteriors}, distributions that are sharper than the Bayesian posterior.
Similarly, temperatures $T > 1$ correspond to \textit{warm posteriors} which are softer than the Bayesian posterior.
See Appendix \autoref{fig:app_posterior_density}(d) for a visualization of the log-likelihood density surface at different temperatures.

\citet{wenzel2020good} argue that Bayesian neural networks require a cold posterior, and the performance at temperature $T=1$ is inferior to even a single model trained with SGD.
The authors refer to this phenomenon as \textit{the cold posteriors effect}.
However, our results are different:
\begin{mybox}
    \textbf{Summary}: We show that that cold posteriors are not needed to obtain near-optimal performance with Bayesian neural networks
    and may even hurt performance. We show that the cold posterior effect is largely an artifact of data augmentation.
\end{mybox}

\begin{figure}
    \centering
    \hspace{-.2cm}
    \includegraphics[height=0.125\textwidth]{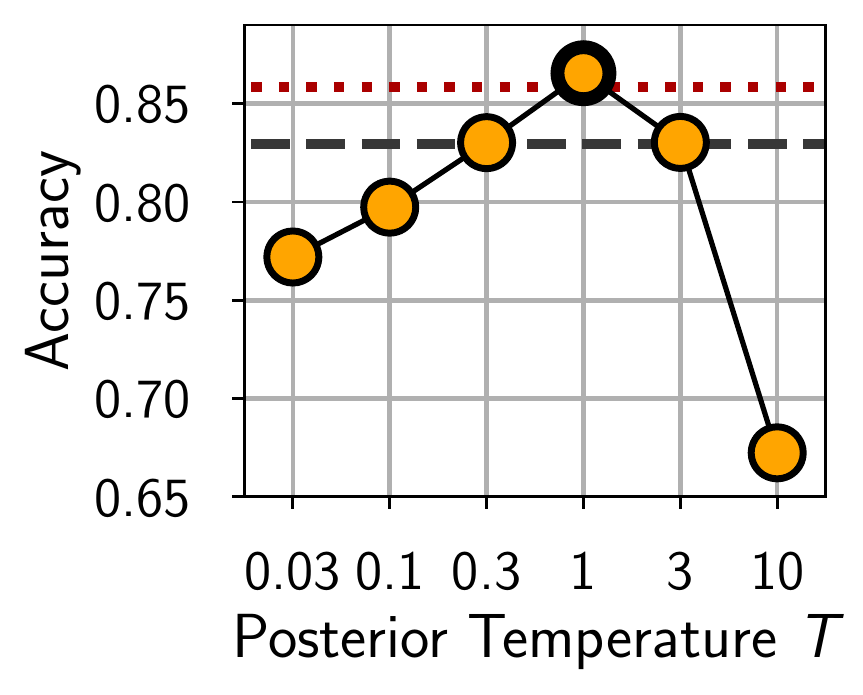}
    \hspace{-.2cm}
    \includegraphics[height=0.125\textwidth]{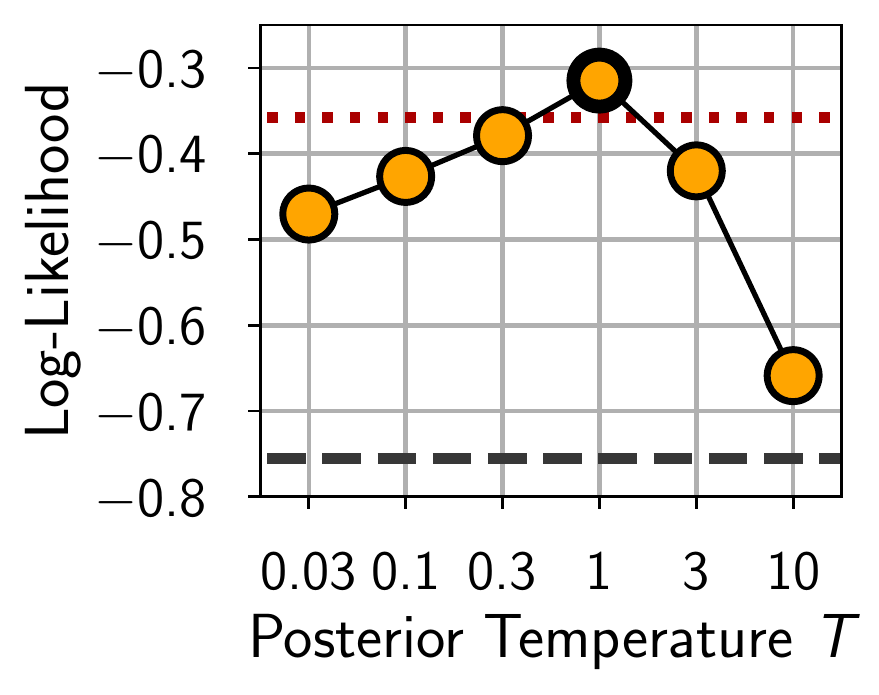}
    \hspace{-.2cm}
    \includegraphics[height=0.125\textwidth]{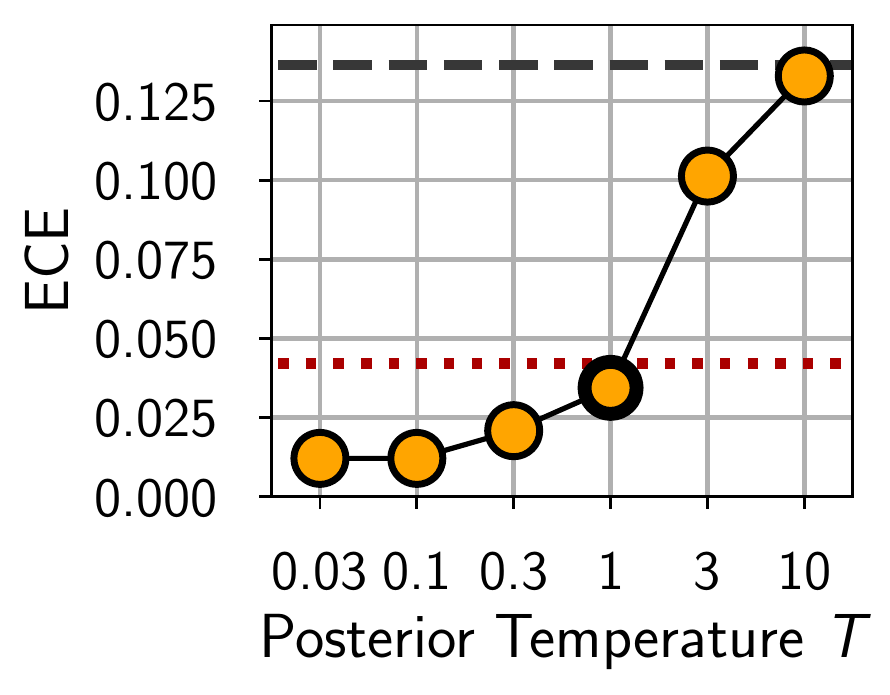}
    \includegraphics[width=0.26\textwidth]{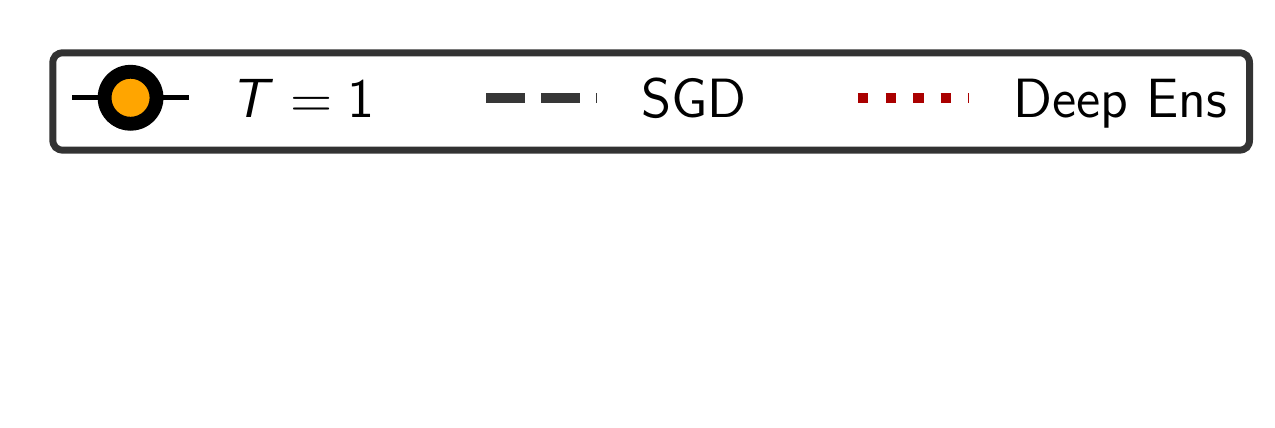}
    \vspace{-0.8cm}
    \caption{\textbf{Effect of posterior temperature.}
    The effect of posterior temperature $T$ on the log-likelihood, accuracy and expected calibration error using the CNN-LSTM model on the IMDB dataset.
    For both the log-likelihood and accuracy $T = 1$ provides optimal performance, while for the ECE the colder posteriors provide a slight improvement. 
    For all three metrics, the posterior at $T = 1$ outperforms the SGD baseline as well as a deep ensemble of $10$ independently trained models. 
    }
    \label{fig:temp_effect}
\end{figure}

\subsection{Testing the cold posteriors effect}

\citet{wenzel2020good} demonstrate the cold posteriors with two main experiments: ResNet-20 on CIFAR-10 and CNN-LSTM on IMDB.
In these experiments the authors show poor performance at temperature $T=1$, with strong benefits from decreasing the temperature.
However, for the CIFAR-10 experiment, it is apparent \citep[Appendix K, Figure 28]{wenzel2020good} that the results at $T = 1$ 
are near-optimal for the ResNet on CIFAR-10 if data augmentation is turned off and batch normalization is replaced with filter response normalization,
which is in fact necessary for a clear Bayesian interpretation of the inference procedure.

Furthermore, in \autoref{sec:bnn_evaluation}, we show that Bayesian neural networks can achieve performance superior to SGD and even deep ensembles at temperature $T=1$,
in particular using the same ResNet-20-FRN model on CIFAR-10 and CNN-LSTM model on IMDB used by \citet{wenzel2020good}.

To further understand the effect of posterior temperature $T$, we compare the performance of the CNN-LSTM model at different $T$ using our Hamiltonian Monte Carlo sampler. 
In all runs we used a fixed prior variance $\alpha^2 = \frac 1 {40}$.
We report the results in \autoref{fig:temp_effect}.
We find that the performance of the BNN at $T=1$ is better than the SGD baseline as well as a deep ensemble of $50$ independent models.
Moreover, the performance at $T=1$ is better compared to all other temperatures we tested in terms of both test accuracy and log-likelihood.

We also note that while posterior tempering does not seem necessary for good predictive performance with BNNs, it may be helpful
under distribution shift.
In \autoref{sec:app_robustness} we show that decreasing the temperature can significantly improve the robustness of BNN predictions
to noise in the test inputs. \citet{wilson2020bayesian} additionally argue that tempering may be a reasonable procedure in general, and is 
not necessarily at odds with Bayesian principles. 

\textbf{Role of data augmentation.}\quad
Our results are in contrast with \citet{wenzel2020good}, who argue that cold posteriors are needed for good performance with BNNs.
In \autoref{sec:app_cold_posteriors} we provide an additional study of what may have caused the poor performance of BNNs in \citet{wenzel2020good},
using the code for inference provided by \citet{wenzel2020good}. 
We identify data augmentation as the key factor responsible for the cold posterior effect, and also show that batch normalizing does not significantly influence this effect:
when the data augmentation is turned off, we do not observe the cold posteriors effect.
Data augmentation cannot be naively incorporated in the Bayesian neural network model (see the discussion in appendix K of \citet{wenzel2020good}),
and arguably it may be reasonable to decrease the temperature when using data augmentation:
intuitively, data augmentation increases the amount of data observed by the model, and should lead
to higher posterior contraction.
We leave incorporating data augmentation in our HMC evaluation framework as an exciting direction of future work.

\section{What is the effect of  priors in Bayesian neural networks?}
\label{sec:app_priors}

\begin{figure}[t]
    \centering
    
    \hspace{-0.3cm}
    \begin{tabular}{cccc}
    \hspace{-.3cm}
    \rotatebox{90}{\scriptsize \quad\quad~~~\textbf{CIFAR-10}} &
    \hspace{-.3cm}
    \includegraphics[height=0.11\textwidth]{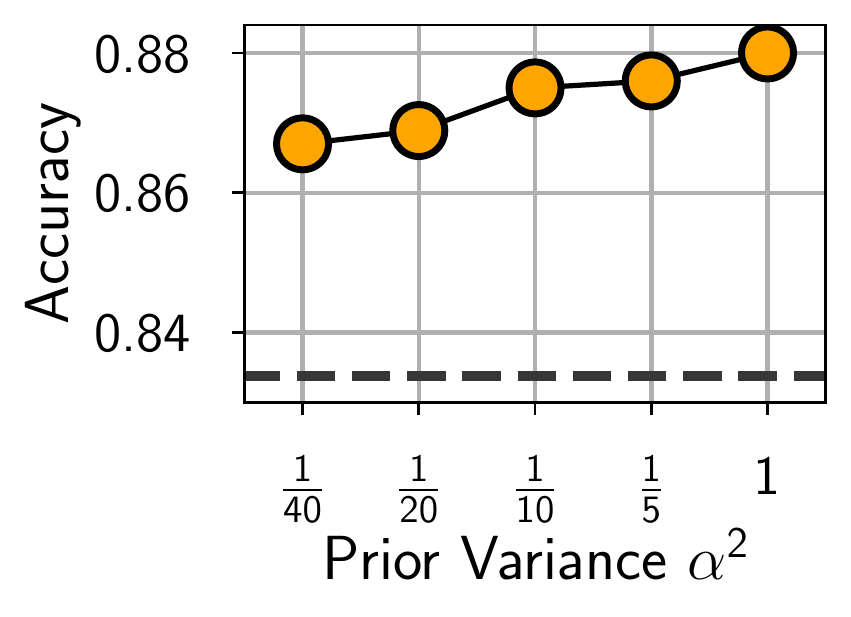} &
    \hspace{-.4cm}
    \includegraphics[height=0.11\textwidth]{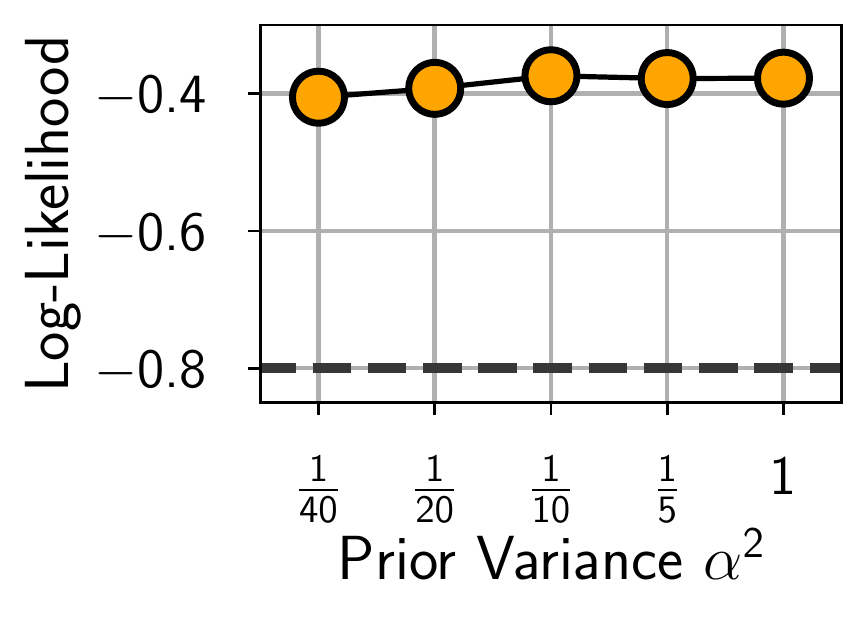} &
    \hspace{-.5cm}
    \includegraphics[height=0.11\textwidth]{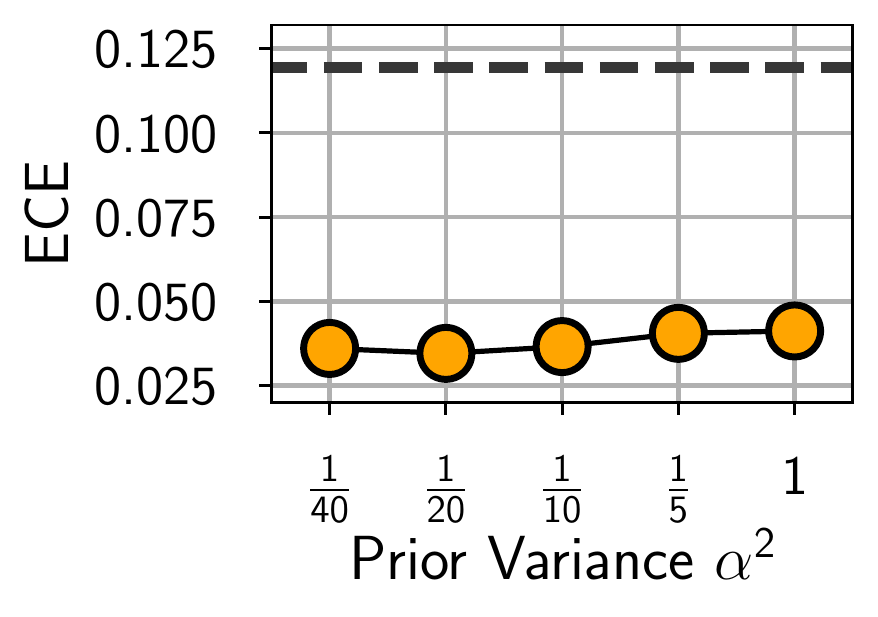}
    \\[-0.cm]
    \hspace{-.3cm}
    \rotatebox{90}{\scriptsize \quad\quad\quad~~~\textbf{IMDB}} &
    \hspace{-.3cm}
    \includegraphics[height=0.11\textwidth]{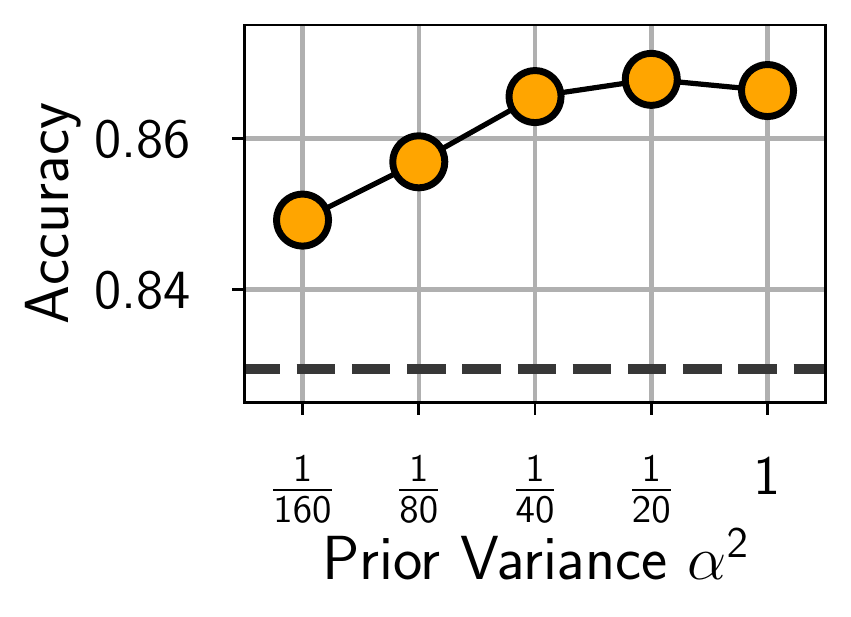} &
    \hspace{-.4cm}
    \includegraphics[height=0.11\textwidth]{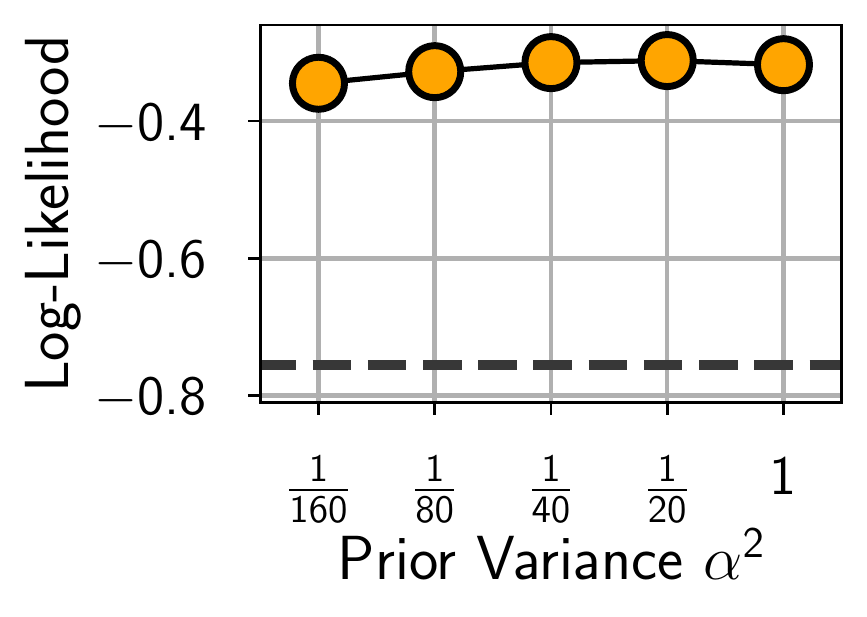} &
    \hspace{-.5cm}
    \includegraphics[height=0.11\textwidth]{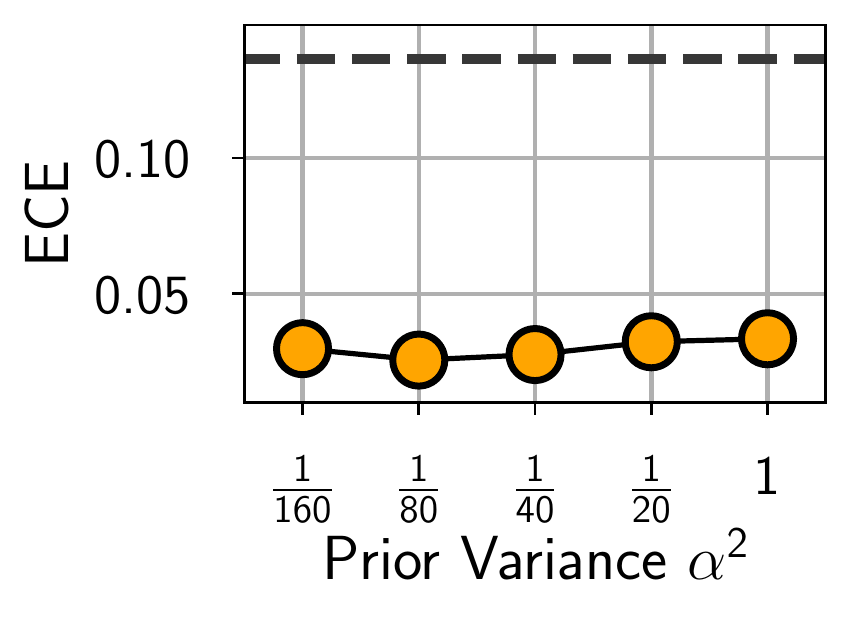}\\[-0.3cm]
    \end{tabular}
	\caption{
	    \textbf{Effect of prior variance.}
        The effect of prior variance on BNN performance. 
        In each panel, the dashed line shows the performance of the SGD model from \autoref{sec:bnn_evaluation}.
        While low prior variance may lead to over-regularization and hurt performance, all the considered 
        prior scales lead to better results than the performance of an SGD-trained neural network of the same architecture.
	}
	\label{fig:prior_ablation}
\end{figure}

Bayesian deep learning is often criticized for the lack of intuitive priors over the parameters.
For example, \citet{wenzel2020good} hypothesize that the popular Gaussian priors of the form $\mathcal{N}(0, \alpha^2 I)$ are inadequate and lead to poor performance.
\citet{tran2020all} propose a new prior for Bayesian neural networks inspired by Gaussian processes \citep{rasmussen10gpml} based on this hypothesis. In concurrent work,
\citet{fortuin2021bayesian} also explore several alternatives to standard Gaussian priors inspired by the cold posteriors effect.
\citet{wilson2020bayesian} on the other hand, argue that vague Gaussian priors in the parameter space induce useful function-space priors.

In \autoref{sec:bnn_evaluation} we have shown that Bayesian neural networks can achieve strong performance with vague Gaussian priors.
In this section, we explore the sensitivity of BNNs to the choice of the prior scale as well as several alternative prior families,
as a step towards a better understanding of the role of the prior in BNNs.

\begin{mybox}
    \textbf{Summary:} High-variance Gaussian priors over parameters of BNNs lead to strong performance.
    The results are robust with respect to the prior scale. Mixture of Gaussian and logistic priors over parameters are not too 
    different in performance to Gaussian priors. These results highlight the relative importance of architecture over parameter priors 
    in specifying a useful prior over functions.
\end{mybox}

\subsection{Effect of Gaussian prior scale}

We use priors of the form $\mathcal{N}(0, \alpha^2 I)$ and vary the prior variance $\alpha^2$.
For all cases, we use a single HMC chain producing $40$ samples.
These are much shorter chains than the ones we used in \autoref{sec:bnn_evaluation}, so the results
are not as good;
the purpose of this section is to explore the \textit{relative} performance of BNNs under different priors.

We report the results for the CIFAR-10 and IMDB datasets in \autoref{fig:prior_ablation}.
When the prior variance is too small, the regularization is too strong, hindering the performance.
Setting the prior variance too large does not seem to hurt the performance as much.
On both problems, the performance is fairly robust: a wide window of prior variances lead 
to strong performance.
In particular, for all considered prior scales, the results are better than those of SGD training.

\textbf{Why are BNNs so robust to the prior scale?}\quad One possible explanation for the relatively flat curves in \autoref{fig:prior_ablation} is that large prior variances imply a strong prior belief that the ``true'' classifier (i.e., the model that would be learned given infinite data) should make high-confidence predictions.
Since the model is powerful enough to achieve any desired training accuracy, the likelihood does not overrule this prior belief, and so the posterior assigns most of its mass to very confident classifiers.
Past a certain point, increasing the prior variance on the weights may have no effect on the classifiers' already saturated probabilities.
Consequently, nearly every member of the BMA may be highly overconfident. 
But the \emph{ensemble} does not have to be overconfident---a mixture of overconfident experts can still make well-calibrated predictions. 
Appendix \autoref{fig:overconfident} provides some qualitative evidence for this explanation; for some CIFAR-10 test set images, the HMC chain oscillates between assigning the true label probabilities near 1 and probabilities near 0.

\begin{table}[t]
\begin{center}
\begin{small}
\begin{sc}
\begin{tabular}{@{}llll@{}}
\toprule
Prior          & Gaussian    & MoG        & Logistic   \\ \midrule
Accuracy       & 0.866       & 0.863        & \textbf{0.869}       \\
ECE            & 0.029       & 0.025        & \textbf{0.024}       \\
Log Likelihood & -0.311      & -0.317       & \textbf{-0.304}      \\ \bottomrule 
\end{tabular}
\end{sc}
\end{small}
\end{center}
\caption{\textbf{Non-Gaussian priors.} BMA accuracy, ECE,
and log-likelihood under different prior families using CNN-LSTM on IMDB.
We produce 80 samples from a single HMC chain for each of the priors.
The heavier-tailed logistic prior provides slightly better performance
compared to the Gaussian and mixture of Gaussians (MoG) priors.}
\label{tab:alternate-priors}
\end{table}

\begin{table*}[t]
    \begin{center}
    \begin{small}
    \begin{sc}
    \begin{tabular}{c c ccc cccc}
    \toprule
    & & & & & \multicolumn{4}{c}{SGMCMC} \\
    \cline{6-9}\\[-0.2cm]
    Metric            & 
    \begin{tabular}{c}HMC\\(reference)\end{tabular}  
    & SGD & Deep Ens     & MFVI    & SGLD   &  SGHMC  & 
    \begin{tabular}{c}SGHMC\\CLR\end{tabular} & 
    \begin{tabular}{c}SGHMC\\CLR-Prec\end{tabular} 
    \\
    \midrule
    \multicolumn{8}{c}{CIFAR-10}\\
    \midrule
    \multirow{2}{*}{Accuracy}    
    & $89.64$      & $83.44$      & $88.49$      & $86.45$      & $89.32$      & $89.38$      & $\mathbf{89.63}$      & $87.46$ \\[-0.1cm]
    & {\scriptsize ${\pm 0.25}$} & {\scriptsize ${\pm 1.14}$} & {\scriptsize ${\pm 0.10}$} & {\scriptsize ${\pm 0.27}$} & {\scriptsize ${\pm 0.23}$} & {\scriptsize ${\pm 0.32}$} & {\scriptsize $\mathbf{{\pm 0.37}}$} & {\scriptsize ${\pm 0.21}$} \\[0.1cm]
    \multirow{2}{*}{Agreement}  
    & $94.01$ &  $85.48$ &  $91.52$ &  $88.75$ &  $91.54$ &  $91.98$ &  $\mathbf{92.67}$ &  $90.96$\\[-0.1cm]
    & {\scriptsize ${\pm 0.25}$  } & {\scriptsize ${\pm 1.00}$  } & {\scriptsize ${\pm 0.06}$  } & {\scriptsize ${\pm 0.24}$  } & {\scriptsize ${\pm 0.15}$  } & {\scriptsize ${\pm 0.35}$  } & {\scriptsize $\mathbf{\pm 0.52}$  } & {\scriptsize ${\pm 0.24}$}\\[0.1cm]
    \multirow{2}{*}{Total Var}
    & $0.074$ &  $0.190$ &  $0.115$ &  $0.136$ &  $0.110$ &  $0.109$ &  $\mathbf{0.099}$ &  $0.111$ \\[-0.1cm]
    & {\scriptsize ${\pm 0.003}$ } & {\scriptsize ${\pm 0.005}$ } & {\scriptsize ${\pm 0.000}$ } & {\scriptsize ${\pm 0.000}$ } & {\scriptsize ${\pm 0.001}$ } & {\scriptsize ${\pm 0.001}$ } & {\scriptsize $\mathbf{\pm 0.006}$ } & {\scriptsize ${\pm 0.002}$} \\
    \midrule
    \multicolumn{8}{c}{CIFAR-10-C}\\
    \midrule
    \multirow{2}{*}{Accuracy}
    & $70.91$ &  $71.04$ &  $76.99$ &  $75.40$ &  $\mathbf{78.80}$ &  $78.20$ &  $76.43$ &  $73.42$ \\[-0.1cm]
    & {\scriptsize ${\pm 0.93}$  } & {\scriptsize ${\pm 1.80}$  } & {\scriptsize ${\pm 0.39}$  } & {\scriptsize ${\pm 0.34}$} & {\scriptsize $\mathbf{\pm 0.17}$  } & {\scriptsize ${\pm 0.25}$  } & {\scriptsize ${\pm 0.39}$} & {\scriptsize ${\pm 0.39}$}\\[0.1cm]
    \multirow{2}{*}{Agreement}
    &$86.00$ &  $72.01$ &  $79.29$ &  $75.47$ &  $77.99$ &  $78.98$ &  $\mathbf{80.93}$ &  $79.65$ \\[-0.1cm]
    &{\scriptsize ${\pm 0.44}$  } & {\scriptsize ${\pm 0.82}$  } & {\scriptsize ${\pm 0.18}$  } & {\scriptsize ${\pm 0.27}$  } & {\scriptsize ${\pm 0.22}$} & {\scriptsize ${\pm 0.22}$} & {\scriptsize $\mathbf{\pm 0.73}$} & {\scriptsize ${\pm 0.35}$} \\[0.1cm]
    \multirow{2}{*}{Total Var}
    & $0.133$ &  $0.334$ &  $0.220$ &  $0.245$ &  $0.214$ &  $0.203$ &  $\mathbf{0.194}$ &  $0.205$ \\[-0.1cm]
    &{\scriptsize ${\pm 0.004}$ } & {\scriptsize ${\pm 0.007}$ } & {\scriptsize ${\pm 0.003}$ } & {\scriptsize ${\pm 0.002}$ } & {\scriptsize ${\pm 0.002}$ } & {\scriptsize ${\pm 0.002}$ } & {\scriptsize $\mathbf{\pm 0.010}$ } & {\scriptsize ${\pm 0.005}$}\\
    \bottomrule
    \end{tabular}
    \end{sc}
    \end{small}
    \end{center}
    \caption{\textbf{Evaluation of cheaper alternatives to HMC.}
    Agreement and total variation between predictive distributions of HMC and approximate inference methods: deep ensembles, mean field variational inference (MFVI), and stochastic gradient Monte Carlo (SGMCMC) variations.
    For all methods we use ResNet-20-FRN trained on CIFAR-10 and evaluate predictions on the CIFAR-10 and CIFAR-10-C test sets.
    For CIFAR-10-C we report the average results across all corruptions and corruption intensities.
    We additionally report the results for HMC for reference: we compute the agreement and total variation between one of the chains and the ensemble of the other two chains.
    For each method we report the mean and standard deviation of the results over three independent runs.
    MFVI provides the worst approximation of the predictive distribution.
    Deep ensembles despite often being considered non-Bayesian, significantly outperform MFVI.
    SG-MCMC methods provide the best results with SGHMC-CLR showing the best overall performance.
    }
    \label{tab:pred_comparison}
\end{table*}

\subsection{Non-Gaussian priors}

In \autoref{tab:alternate-priors}, we report BMA accuracy, ECE, and log-likelihood 
for two non-Gaussian priors on the IMDB dataset: \textit{logistic} and \textit{mixture of Gaussians} (MoG). 
For the MoG prior we use a mixture of two Gaussians centered at $0$, one with variance $\frac 1 {40}$ and the other with variance $\frac 1 {160}$.
We pick prior scale of the logistic prior to have a variance of $\frac{1}{40}$.
We additionally provide the results for a Gaussian prior with variance $\frac 1 {40}$.
We approximate the BMA using 80 samples from a single HMC chain for each of the priors. 
We find that the heavier-tailed logistic prior performs slightly better than the Gaussian and MoG.

\subsection{Importance of Architecture in Prior Specification}
We often think of the prior narrowly in terms of a distribution over parameters $p(w)$. But the prior that matters 
is the prior over functions $p(f(x))$ that is induced when a prior over parameters $p(w)$ is combined with the 
functional form of a neural network $f(x,w)$. All of the results in this section point to the relative importance of
the architecture in defining the prior over functions, compared to the prior over parameters. A vague prior over 
parameters is not necessarily vague in function-space. 
Moreover, while the details of the prior distribution over
parameters $p(w)$ have only a minor effect on performance, the choice of architecture certainly has a major effect
on performance.

\section{Do scalable BDL methods and HMC make similar predictions?}
\label{sec:alternatives}

While HMC shows strong performance in our evaluation in \autoref{sec:bnn_evaluation}, 
in most realistic BNN settings it is an impractical method.
However, HMC can be used as a \textit{reference} to evaluate and calibrate more scalable and practical
alternatives.
In this section, we evaluate the \emph{fidelity} of SGMCMC, variational methods, and deep ensembles in representing the 
predictive distribution (Bayesian model average) given by our HMC reference.

\begin{mybox}
    \textbf{Summary}: While SGMCMC and Deep Ensembles can provide good generalization accuracy and calibration, their predictive distributions differ from HMC. Deep ensembles are similarly close to the HMC predictive distribution as SGLD, and closer than standard variational inference.
\end{mybox}

\subsection{Comparing the predictive distributions}
We consider two primary metrics: \textit{agreement} and \textit{total variation}.
We define the agreement between the predictive distributions $\hat p$ of HMC and 
$p$ of another method as the fraction of the test data points for which the top-1 predictions of $\hat p$ and $p$ are the same:
\begin{equation*}
    \frac 1 n \sum_{i=1}^n I[\arg\max_j \hat p (y = j \vert x_i) = \arg\max_j p(y = j \vert x_i)],
\end{equation*}
where $I[\cdot]$ is the indicator function and $n$ is the number of test data points $x_i$.
We define the total variation metric between $\hat p$ and $p$ as the total variation distance between the predictive distributions averaged over the test data points:
\begin{equation*}
    \frac 1 n \sum_{i=1}^n \frac{1}{2}\sum_j \bigg| \hat p(y = j \vert x_i) - p(y = j \vert x_i)\bigg|.
\end{equation*}
The agreement (higher is better) captures how well a method is able to capture the top-1 predictions of HMC,
while the total variation (lower is better) compares the predictive probabilities for each of the classes.

In \autoref{tab:pred_comparison} we report the agreement and total variation metrics as well as the predictive accuracy on the CIFAR-10 and CIFAR-10-C test sets
for a deep ensemble of $50$ models and several SGLD variations: 
standard SGLD \citep{welling2011bayesian}, SGLD with momentum (SGHMC) \citep{chen2014stochastic}, 
SGLD with momentum and a cyclical learning rate schedule (SGHMC-CLR) \citep{zhang2019cyclical} and 
SGLD with momentum, cyclical learning rate schedule and a preconditioner (SGHMC-CLR-Prec) \citep{wenzel2020good}.
All methods were trained on CIFAR-10.
For more details, please see \autoref{sec:app_hypers}.

Overall, the absolute value of agreement achieved by all methods is fairly low
on CIFAR-10 and especially on CIFAR-10-C.
More advanced SGHMC-CLR and SGHMC-CLR-Prec methods provide a better fit of the HMC predictive distribution
while not necessarily improving the accuracy.
Notably, these methods are also \textit{less} robust to the data corruptions in CIFAR-10-C, again suggesting that 
higher fidelity representations of the predictive distribution can lead to decreased robustness to covariate shift, 
as we found in section~\ref{sec:image_classification}.

Deep ensembles, while not typically considered to be a Bayesian method, provide a reasonable
approximation to the HMC predictive distribution.
In particular, deep ensembles outperform both SGLD and SGHMC in terms of total variation on CIFAR-10 and in terms of agreement on CIFAR-10-C.
These results support the argument that deep ensembles, while not typically characterized as a Bayesian method, provide 
a \emph{higher fidelity} approximation to a Bayesian model average than methods that are conventionally accepted as Bayesian 
inference procedures in modern deep learning \citep{wilson2020bayesian}.

In \autoref{sec:app_robustness} we explore the performance of HMC, SGD, deep ensembles, SGLD and SGHMC-CLR-Prec under different corruptions individually. 
Interestingly, the behavior of SGLD and SGHMC-CLR-Prec appears more similar to that of deep ensembles than that of HMC.
So, while both SGMCMC and deep ensembles are very compelling practically, they provide relatively distinct predictive distributions from HMC.
Mean-field variational inference methods are particularly far from the HMC predictive distribution.
Thus, we should be very careful when making judgements about \textit{true} Bayesian neural networks based on the SGMCMC or MFVI performance.

\subsection{Predictive entropy and calibration curves}

\begin{figure}
    \centering
    \includegraphics[height=0.18\textwidth]{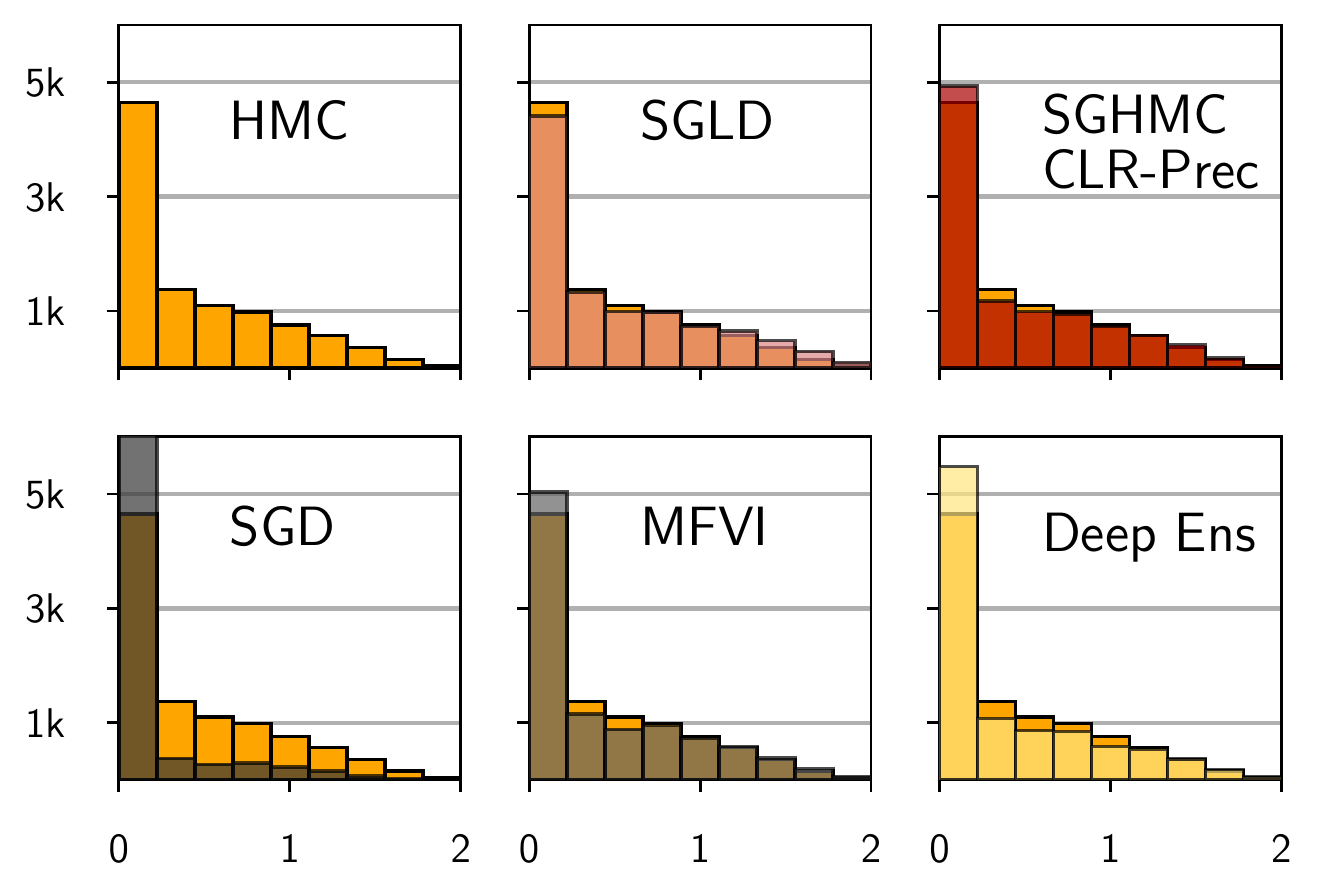}
    \includegraphics[height=0.18\textwidth]{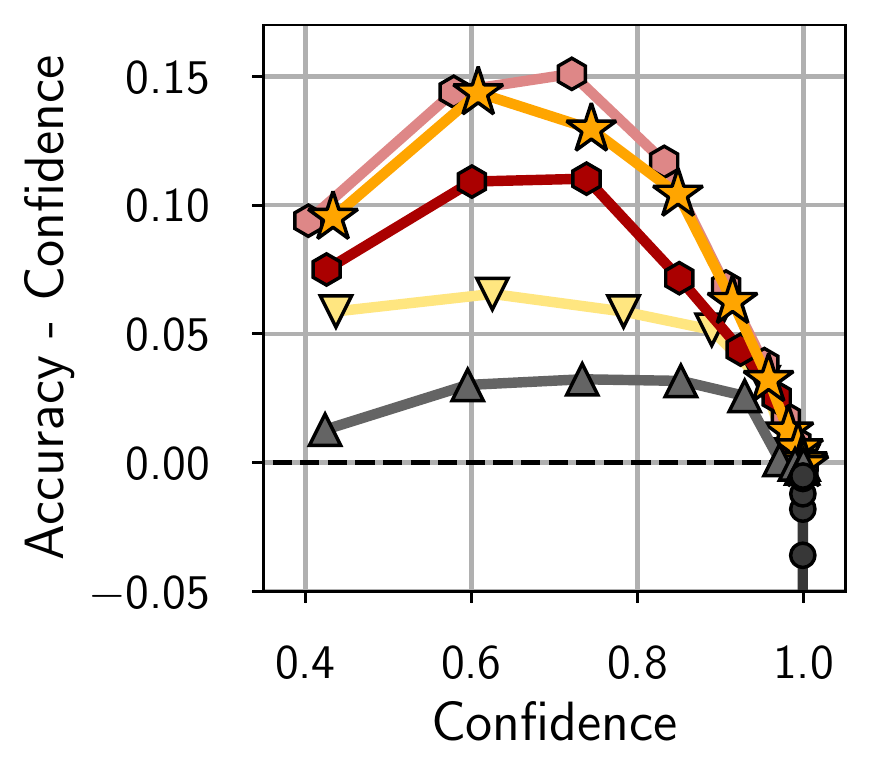}
    \vspace{-0.1cm}

    \includegraphics[width=0.35\textwidth]{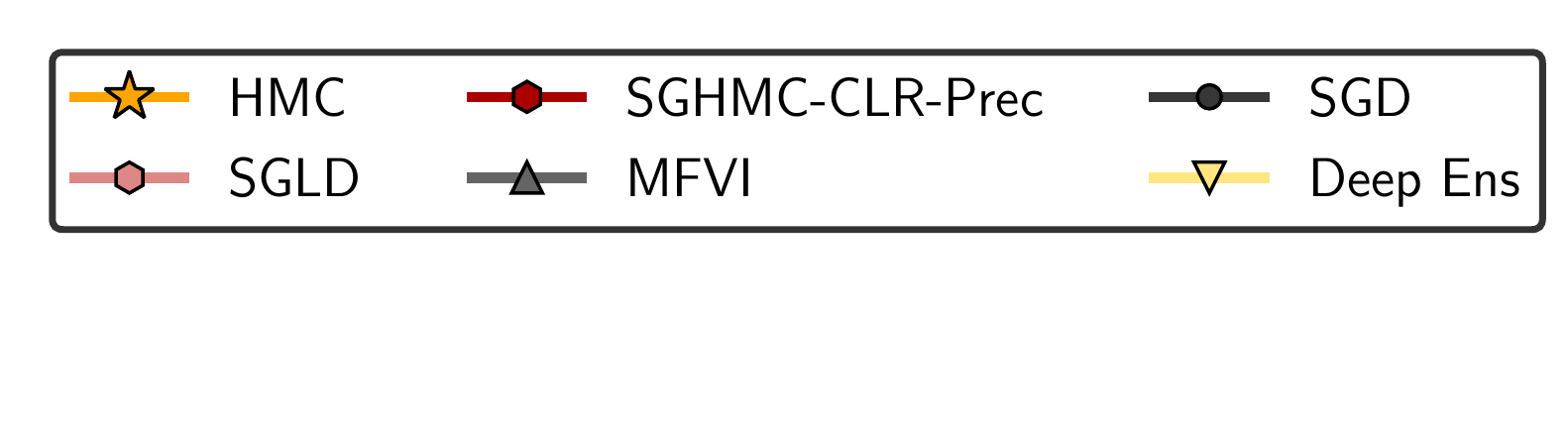}\vspace{-0.7cm}
    
    \caption{
    Distribution of predictive entropies (\textbf{left}) and calibration curve (\textbf{right}) of posterior predictive distributions for 
    HMC, SGD, deep ensembles, MFVI, SGLD and SGHMC-CLR-Prec for ResNet20-FRN on CIFAR-10.
    On the left, for all methods, except HMC we plot a pair of histograms: for HMC and for the corresponding method.
    SGD, Deep ensembles and MFVI provide more confident predictions than HMC.
    SGMCMC methods appear to fit the predictive distribution of HMC better: SGLD is slightly underconfident relative to HMC while SGHMC-CLR-Prec is slightly over-confident.
    }
    \label{fig:posterior_predictive}
\end{figure}

To provide an additional comparison of the predictive distributions between HMC and other methods, in \autoref{fig:posterior_predictive}
we visualize the distribution of predictive entropies and the calibration curves for HMC, SGD, deep ensembles, MFVI, SGLD and SGHMC-CLR-Prec on CIFAR-10 using ResNet-20-FRN.

All methods except fot SGD make conservative predictions: their confidences tend to underestimate their accuracies (\autoref{fig:posterior_predictive}, right);
SGD on the other hand is very over-confident \citep[in agreement with the results in ][]{guo2017calibration}.
Deep ensembles and MFVI provide the most calibrated predictions, while SGLD and SGHMC-CLR-Prec match the HMC entropy distribution and calibration curve closer.

\section{Discussion}

Despite the rapidly increasing popularity of approximate Bayesian inference in modern deep learning, 
little is known about the behaviour of truly Bayesian neural networks. 
To the best of our knowledge, our work provides the first realistic evaluation of Bayesian neural networks
with precise and exhaustive posterior sampling. We establish several properties of Bayesian neural networks,
including good generalization performance, lack of a cold posterior effect, and a surprising lack of robustness
to covariate shift.
We hope that our observations and the tools that we develop will facilitate fundamental progress in understanding
the behaviour of Bayesian neural networks.

\textbf{Is HMC Converging?}\quad
In general, it is not possible to ensure that an MCMC method has converged to sampling from the true posterior distribution: 
theoretically, there may always remain regions of the posterior that cannot be discovered by the method but that contain most of the posterior mass.
To maximize the performance of HMC, we choose the hyper-parameters that are the most likely to provide convergence: long trajectory lengths, and multiple long chains. 
In \autoref{sec:mixing}, we study the convergence of HMC using the available convergence diagnostics.
We find that while HMC does not mix perfectly in weight space, in the space of predictions we cannot find evidence of non-mixing.

\textbf{Should we use Bayesian neural networks?}\quad
Our results show that Bayesian neural networks can improve the performance over models trained with SGD in a variety of settings,
both in terms of uncertainty calibration and predictive accuracy.
On most of the problems we considered in this work, the best results were achieved by Bayesian neural networks.
We believe that our results provide motivation to use Bayesian neural networks with accurate posterior approximation in practical applications.

\textbf{Should we use HMC in practice?}\quad
For most realistic scenarios in Bayesian Deep Learning, HMC is an impractical method.
On the image classification benchmarks, HMC takes orders of magnitude more compute than any of the baselines that we considered.
We hope that our work will inspire the community to produce new accurate and scalable approximate inference methods for Bayesian deep learning.

\subsection{Challenging conventional wisdom}

A conventional wisdom has emerged that deep ensembles are a non-Bayesian alternative to variational methods, that standard priors for neural networks are poor, and that cold posteriors
are a problematic result for Bayesian deep learning. Our results highlight that one should take care in uncritically 
repeating such claims. In fact, 
deep ensembles appear to provide a higher fidelity representation of the Bayesian predictive distribution than widely accepted approaches to approximate
Bayesian inference. If anything, the takeaway from the relatively good performance of deep ensembles is that we would benefit from approximate inference 
being \emph{closer} to the Bayesian ideal! Moreover, the details over the priors in weight space can have a relatively minor effect on performance, and there
is no strong evidence that standard Gaussian priors are particularly bad. In fact, there are many reasons to believe these priors have useful properties
\citep{wilson2020bayesian}. Similarly, on close inspection, we found no evidence for a general cold posterior effect, which we identify as largely an artifact of 
data augmentation. Although we see here that tempering does not in fact seem to be required, as argued in \citet{wilson2020bayesian} tempering is also not necessarily unreasonable or even divergent from Bayesian principles. 

Even the results we found that are less favourable to Bayesian deep learning are contrary to the current orthodoxy. Indeed, 
higher fidelity Bayesian inference surprisingly appears to suffer more greatly from covariate shift, despite the popularity of approximate Bayesian inference procedures in this 
setting.

\subsection*{Acknowledgements}
The authors would like to thank many people at Google for many helpful discussions and much helpful feedback, especially Rodolphe Jenatton, Rif A. Saurous, Jasper Snoek, Pavel Sountsov, Florian Wenzel, and the entire TensorFlow Probability team. This  research  is  supported  by  an  Amazon  Research  Award,  NSF  I-DISRE  193471,  NIH  R01DA048764-01A1,  NSF IIS-1910266,  and NSF 1922658NRT-HDR: FUTURE Foundations,  Translation,  and Responsibility for Data Science.

\bibliography{ubdl}
\bibliographystyle{icml2021}

\appendix

\begin{table*}[]
\begin{center}
\begin{small}
\begin{sc}
\begin{tabular}{@{}clclll@{}}
\toprule
                        &                            &            & \multicolumn{3}{c}{Experiments}                           \\ 
													                \cline{4-6}\\[-0.2cm]
                        &                            &            & CIFAR-10          & CIFAR-100         & IMDB              \\
Method                  & Hyper-parameter            & Was Tuned  & Resnet-20-FRN     & Resnet-20-FRN     & CNN LSTM          \\
\midrule
\multirow{7}{*}{HMC}    & Prior Variance             & \checkmark & $\frac{1}{5}$     & $\frac{1}{5}$     & $\frac{1}{40}$    \\
                        & Step Size                  & \checkmark & $10^{-5}$         & $10^{-5}$         & $10^{-5}$ \\
                        & Num. Burnin Iterations     & \ding{55}  & 50                & 50                & 50                \\
                        & Num. Samples per Chain     & \ding{55}  & 240               & 40                & 400               \\
                        & Num. of Chains             & \ding{55}  & 3                 & 3                 & 3                 \\
                        & Total Samples              & \ding{55}  & 720               & 120               & 1200              \\
                        \cline{2-6}\\[-0.2cm]
                        & Total Epochs               &            & $5 \cdot 10^7$    & $8.5 \cdot 10^6$  & $3 \cdot 10^7$    \\
\midrule
\multirow{6}{*}{SGD}    & Weight Decay               & \checkmark & $10$              & $10$              & $3$    \\
                        & Initial Step Size          & \checkmark & $3 \cdot 10^{-7}$ & $1 \cdot 10^{-6}$ & $3 \cdot 10^{-7}$ \\
                        & Step Size Schedule         & \ding{55}  & cosine            & cosine            & cosine \\
                        & Batch Size                 & \checkmark & 80                & 80                & 80 \\
                        & Num. Epochs                & \ding{55}  & 500               & 500               & 500 \\
                        & Momentum                   & \ding{55}  & $0.9$             & $0.9$             & $0.9$ \\
                        \cline{2-6}\\[-0.2cm]
                        & Total Epochs               &            & $5\cdot 10^2$     & $5\cdot 10^2$      & $5\cdot 10^2$    \\
\midrule
Deep Ensembles          & Num. Models                & \ding{55}  & $50$              & $50$              & $50$    \\
                        \cline{2-6}\\[-0.2cm]
                        & Total Epochs               &            & $2.5 \cdot 10^4$  & $2.5 \cdot 10^4$  & $2.5 \cdot 10^4$    \\
\midrule
\multirow{10}{*}{SGLD}  & Prior Variance             & \checkmark & $\frac{1}{5}$     & $\frac{1}{5}$     & $\frac{1}{5}$   \\
                        & Step Size                  & \checkmark & $10^{-6}$ & $3 \cdot 10^{-6}$ & $1 \cdot 10^{-5}$ \\
                        & Step Size Schedule         & \checkmark & constant          & constant          & constant \\
                        & Batch Size                 & \checkmark & 80                & 80                & 80 \\
                        & Num. Epochs                & \ding{55}  & 10000             & 10000             & 10000 \\
                        & Num. Burnin Epochs         & \ding{55}  & 1000              & 1000              & 1000                \\
                        & Num. Samples per Chain     & \ding{55}  & 900               & 900               & 900               \\ 
                        & Num. of Chains             & \ding{55}  & 5                 & 5                 & 5                 \\
                        & Total Samples              & \ding{55}  & 4500              & 4500             & 4500              \\
                        \cline{2-6}\\[-0.2cm]
                        & Total Epochs               &            & $5 \cdot 10^4$    & $5 \cdot 10^4$   & $5 \cdot 10^4$    \\
\midrule
\multirow{10}{*}{MFVI}  & Prior Variance             & \ding{55}  & $\frac{1}{5}$     & $\frac{1}{5}$     & $\frac{1}{5}$   \\
						& Num. Epochs                & \ding{55}  & 300               & 300               & 300 \\
						& Optimizer                  & \checkmark & Adam              & Adam              & Adam \\
						& Initial Step Size          & \checkmark & $10^{-4}$         & $10^{-4}$         & $10^{-4}$ \\
						& Step Size Schedule         & \ding{55}  & cosine            & cosine            & cosine \\
                        & Batch Size                 & \checkmark & 80                & 80                & 80 \\
                        & VI mean init               & \ding{55}  & SGD solution      & SGD solution      & SGD solution \\
                        & VI variance init           & \checkmark & $10^{-2}$         & $10^{-2}$         & $10^{-2}$ \\
                        & Number of samples          & \ding{55}  & $50$              & $50$              & $50$ \\
                        \cline{2-6}\\[-0.2cm]
                        & Total Epochs               &            & $8 \cdot 10^2$    & $8 \cdot 10^2$    & $8 \cdot 10^2$    \\ 
\bottomrule
\end{tabular}
\end{sc}
\end{small}
\end{center}
\caption{\textbf{Hyper-parameters for CIFAR and IMDB.}
We report the hyper-parameters for each method our main evaluations on CIFAR and IMDB datasets in \autoref{sec:bnn_evaluation}.
For each method we report the total number of training epochs equivalent to the amount of compute spent.
We run HMC on a cluster of $512$ TPUs, and the baselines on a cluster of $8$ TPUs.
For each of the hyper-parameters we report whether it was tuned via cross-validation, or whether a value was selected without tuning.
}
\label{tab:hyperparameters}
\end{table*}

\begin{table*}[]
\begin{center}
\begin{small}
\begin{sc}
\begin{tabular}{@{}clclllll@{}}
\toprule
                        &                            &            & \multicolumn{5}{c}{Experiments}                           \\ 
													                \cline{4-8}\\[-0.2cm]
Method                  & Hyper-parameter            & Was Tuned  & Concrete          & Yacht             & Energy            & Boston            & Naval\\
\midrule
\multirow{7}{*}{HMC}    & Prior Variance             & \checkmark & $\frac{1}{10}$    & $\frac{1}{10}$    & $\frac{1}{10}$    & $\frac{1}{10}$    & $\frac{1}{40}$ \\
                        & Step Size                  & \checkmark & $10^{-5}$         &  $10^{-5}$        & $10^{-5}$         & $10^{-5}$         & $5 \cdot 10^{-7}$ \\
                        & Num. Burnin Iterations     & \ding{55}  & 10                & 10                & 10                & 10                & 10\\
                        & Num. Iterations            & \ding{55}  & 90                & 90                & 90                & 90                & 90               \\
                        & Num. of Chains             & \ding{55}  & 1                 & 1                 & 1                 & 1                 & 1 \\
\midrule
\multirow{6}{*}{SGD}    & Weight Decay               & \checkmark & $10$              & $10^{-1}$         & $10$              & $10^{-1}$         & $1$  \\
                        & Initial Step Size          & \checkmark & $3 \cdot 10^{-5}$ & $3 \cdot 10^{-6}$ & $3 \cdot 10^{-6}$ & $3 \cdot 10^{-6}$ & $10^{-6}$ \\
                        & Step Size Schedule         & \ding{55}  & cosine            & cosine            & cosine            & cosine            & cosine \\
                        & Batch Size                 & \ding{55}  & 927               & 277               & 691               & 455               & 10740 \\
                        & Num. Epochs                & \checkmark & $1000$            & $5000$            & $5000$            & $500$             & $1000$\\
                        & Momentum                   & \ding{55}  & $0.9$             & $0.9$             & $0.9$             & $0.9$             & $0.9$ \\
\midrule
\multirow{10}{*}{SGLD}  & Prior Variance             & \checkmark & $\frac{1}{10}$    & $\frac{1}{10}$    & $\frac{1}{10}$    & $\frac{1}{10}$    & $1$\\
                        & Step Size                  & \checkmark & $3\cdot 10^{-5}$  & $10^{-4}$         & $3\cdot 10^{-5}$  & $3\cdot 10^{-5}$  & $10^{-6}$\\
                        & Step Size Schedule         & \ding{55}  & constant          & constant          & constant          & constant          & constant \\
                        & Batch Size                 & \ding{55}  & 927               & 277               & 691               & 455               & 10740 \\
                        & Num. Epochs                & \ding{55}  & 10000             & 10000             & 10000             & 10000             & 10000 \\
                        & Num. Burnin Epochs         & \ding{55}  & 1000              & 1000              & 1000              & 1000              & 1000 \\
                        & Num. Samples per Chain     & \ding{55}  & 900               & 900               & 900               & 900               & 900 \\ 
                        & Num. of Chains             & \ding{55}  & 1                 & 1                 & 1                 & 1                 & 1 \\
\bottomrule
\end{tabular}
\end{sc}
\end{small}
\end{center}
\caption{\textbf{Hyper-parameters for UCI.}
We report the hyper-parameters for each method our main evaluations on UCI datasets in \autoref{sec:bnn_evaluation}.
For HMC, the number of iterations is the number of HMC iterations after the burn-in phase; the number of accepted samples is lower.
For each of the hyper-parameters we report whether it was tuned via cross-validation, or whether a value was selected without tuning.
}
\label{tab:uci_hypers}
\end{table*}

\begin{table*}[t]
    \begin{center}
    \begin{small}
    \begin{sc}
    \begin{tabular}{cccccc}
    \toprule
    Hyper-parameter& Was Tuned &
    SGLD &  
    SGHMC & 
    \begin{tabular}{c}SGHMC\\CLR\end{tabular} & 
    \begin{tabular}{c}SGHMC\\CLR-Prec\end{tabular} 
    \\
    \midrule
    Initial Step size       & \checkmark & $10^{-6}$  & $3 \cdot 10^{-7}$ & $3 \cdot 10^{-7}$ & $3 \cdot 10^{-5}$ \\
   	Step Size Schedule      & \ding{55}  & constant   & constant          & cyclical          & cyclical \\
   	Momentum                & \checkmark & 0.         & 0.9               & 0.95              & 0.95 \\
   	Preconditioner          & \ding{55}  & None       & None              & None              & RMSprop \\
   	Num. Samples per chain  & \ding{55}  & 900        & 900               & 180               & 180 \\
   	Num. of Chains          & \ding{55}  & 3          & 3                 & 3                 & 3   \\
    \bottomrule
    \end{tabular}
    \end{sc}
    \end{small}
    \end{center}
    \caption{\textbf{SGMCMC hyper-parameters on CIFAR-10.}
     We report the hyper-parameter values used by each of the SGMCMC methods in \autoref{sec:alternatives}.
     The remaining hyper-parameters are the same as the SGLD hyper-parameters reported in \autoref{tab:hyperparameters}.
     For each of the hyper-parameters we report whether it was tuned via cross-validation, or whether a value was selected without tuning.
    }
    \label{tab:sgmcmc_hypers}
\end{table*}

\FloatBarrier

\section*{Appendix Outline}

This appendix is organized as follows.
We present the Hamiltonian Monte Carlo algorithm that we implement in the paper in \autoref{alg:hmc}, \autoref{alg:leapfrog}.
In \autoref{sec:app_hypers} we provide the details on hyper-parameters used in our experiments.
In \autoref{sec:app_rhat} we provide a description of the $\hat R$ statistic used in \autoref{sec:r_hat}.
In \autoref{sec:app_marginals} we study the marginals of the posterior over parameters estimated by HMC.
In \autoref{sec:app_surface_visualizations} we provide additional posterior density surface visualizations.
In \autoref{sec:app_synthreg} we compare the BMA predictions using two independent HMC chains on a synthetic regression problem.
In \autoref{sec:app_robustness} we show that BNNs are not robust to distribution shift and discuss the reasons
for this behavior. 
In \autoref{sec:app_cold_posteriors} we provide a further discussion of the effect of posterior temperature.
Finally, in \autoref{sec:app_prediction_variations} we visualize the predictions of HMC samples from a single chain on several CIFAR-10 test inputs.

\section{Hyper-Parameters and Details}
\label{sec:app_hypers}

\textbf{CIFAR and IMDB.}\quad
In \autoref{tab:hyperparameters} we report the hyperparameters used by each of the methods in our main evaluation on CIFAR and IMDB datasets in \autoref{sec:bnn_evaluation}.
HMC was run on a cluster of 512 TPUs and the other baselines were run on a cluster of $8$ TPUs.
On CIFAR datasets the methods used a subset of $40960$ datapoints.
All methods were ran at posterior temperature $1$.
We tuned the hyper-parameters for all methods via cross-validation maximizing the accuracy on a validation set.
For the step-sizes we considered an exponential grid with a step of $\sqrt{10}$ with $5$-$7$ different values, 
where the boundaries were selected for each method so it would not diverge.
We considered weight decays $1, 5, 10, 20, 40, 100$ and the corresponding prior variances.
For batch sizes we considered values $80$, $200$, $400$ and $1000$; for all methods lower batch sizes resulted in the best performance.
For HMC we set the trajectory length according to the strategy described in \autoref{sec:trajectory}.
For SGLD, we experimented with using a cosine learning rate schedule decaying to a non-zero final step size, but it did not improve upon
a constant schedule.
For MFVI we experimented with the SGD and Adam optimizers;
we initialize the mean of the MFVI distribution with a pre-trained SGD solution, and the per-parameter variance with a value $\sigma^\text{VI}_\text{Init}$;
we tested values $10^{-2}, 10^{-1}, 10^{0}$ for $\sigma^\text{VI}_\text{Init}$.
For all HMC hyper-parameters, we provide ablations illustrating their effect in \autoref{sec:hmc}.
Producing a single sample with HMC on CIFAR datasets takes roughly one hour on our hardware, and on IMDB it takes $105$ seconds;
we can run up to three chains in parallel.

\textbf{Temperature scaling on IMDB.}\quad
For the experiments in \autoref{sec:cold_posteriors} we run a single HMC chain producing $40$ samples after $10$ burn-in epochs for each temperature.
We used step-sizes 
$5\cdot 10^{-5}$, $3\cdot 10^{-5}$, $10^{-5}$, $3\cdot 10^{-6}$, $10^{-6}$ and $3\cdot 10^{-7}$ for temperatures
$10$, $3$, $1$, $0.3$, $0.1$ and $0.03$ respectively, ensuring that the accept rates were close to $100\%$.
We used a prior variance of $1 / 50$ in all experiments; 
the lower prior variance compared to \autoref{tab:hyperparameters} was chosen to reduce the number of leapfrog iterations,
as we chose the trajectory length according to the strategy described in \autoref{sec:trajectory}.
We ran the experiments on $8$ NVIDIA Tesla V-100 GPUs, as we found that sampling at low temperatures requires \texttt{float64}
precision which is not supported on TPUs.

\textbf{UCI Datasets.}\quad
In \autoref{tab:uci_hypers} we report the hyperparameters used by each of the methods in our main evaluation on UCI datasets in \autoref{sec:bnn_evaluation}.
For each datasets we construct $20$ random splits with $90\%$ of the data in the train and $10\%$ of the data in the test split.
In the evaluation, we report the mean and standard deviation of the results across the splits.
We use another random split for cross-validation to tune the hyper-parameters.
For all datasets we use a fully-connected network with a single hidden layer with $50$ neurons and $2$ outputs
representing the predictive mean and variance for the given input.
We use a Gaussian likelihood to train each of the methods.
For the SGD and SGLD baselines, we did not use mini-batches: the gradients were computed over the entire dataset.
We run each experiment on a single NVIDIA Tesla V-100 GPU.

\textbf{SGMCMC Methods.}\quad
In \autoref{tab:sgmcmc_hypers} we report the hyper-parameters of the SGMCMC methods on the CIFAR-10 dataset used in the evaluation in \autoref{sec:alternatives}.
We considered momenta in the set of $\{0.9, 0.95, 0.99\}$ and step sizes in $\{10^{-4}, 3\cdot 10^{-5}, 10^{-5}, 3\cdot 10^{-6}, 10^{-6}, 3\cdot 10^{-7}, 10^{-7}\}$.
We selected the hyper-parameters with the best accuracy on the validation set.
SGLD does not allow a momentum.

\begin{figure}
\begin{algorithm}[H]
\begin{algorithmic}
\STATE \textbf{Input:} Trajectory length $\tau$, number of burn-in interations $N_{\mathrm{burnin}}$, 
initial parameters $w_\text{init}$, step size $\Delta$, 
number of samples $K$,
unnormalized posterior log-density function $f(w) = \log p(D \vert w) + \log p(w)$.

\STATE \textbf{Output:} Set $S$ of samples $w$ of the parameters.\\

 \STATE $w \leftarrow w_{\text{init}}$; ~~
        $N_{\mathrm{leapfrog}} \leftarrow \frac \tau \Delta$; \\
 \# Burn-in stage\\
 \FOR{$i \leftarrow 1 \ldots N_{\mathrm{burnin}}$} 
  \STATE$m \sim \mathcal N(0, I)$; \\
  \STATE$(w, m) \leftarrow \text{Leapfrog}(w, m, \Delta, N_{\mathrm{leapfrog}}, f)$;
 \ENDFOR
 
 \# Sampling\\
 \STATE $S \leftarrow \varnothing$;
 \FOR{$i \leftarrow 1 \ldots K$}
  \STATE $m \sim \mathcal N(0, I)$; \\
  \STATE $(w', m') \leftarrow \text{Leapfrog}(w, m, \Delta, N_{\mathrm{leapfrog}}, f)$;\\[0.3cm]
  \STATE \# Metropolis-Hastings correction\\
  \STATE $p_{\text{accept}} \leftarrow  \min\left\{1, \frac{f(w')}{f(w)}\cdot \exp{\left(\frac 1 2 \|m\|^2 - \|m'\|^2 \right)} \right\}$; \\
  \STATE $u \sim \text{Uniform}[0, 1]$;\\
  \IF {$u \le p_\text{accept}$}
    \STATE $w \leftarrow w'$;\\
  \ENDIF
 \STATE $S \leftarrow S \cup \{w\}$;
 \ENDFOR
 \caption{Hamiltonian Monte Carlo}
 \label{alg:hmc}
\end{algorithmic}
\end{algorithm}
\vspace{.5cm}

\begin{algorithm}[H]
\begin{algorithmic}
\STATE \textbf{Input:} Parameters $w_0$, initital momentum $m_0$, step size $\Delta$, 
number of leapfrog steps $N_{\mathrm{leapfrog}}$, 
posterior log-density function $f(w) = \log p(w \vert D)$.
\STATE \textbf{Output:} New parameters $w$; new momentum $m$.\\

 \STATE $w \leftarrow w_0$; ~~ $m \leftarrow m_0$;
 
 \FOR{$i \leftarrow 1 \ldots N_{\mathrm{leapfrog}}$}
  \STATE $m \leftarrow m + \frac \Delta  2 \cdot \nabla f(w)$;\\
  \STATE $w \leftarrow w + \Delta \cdot m$;\\
  \STATE $m \leftarrow m + \frac \Delta  2 \cdot \nabla f(w)$;
 \ENDFOR
\STATE  $\text{Leapfrog}(w_0, m_0, \Delta, N_{\mathrm{leapfrog}}, f) \leftarrow (w, m)$
\caption{Leapfrog integration}
\label{alg:leapfrog}
\end{algorithmic}
\end{algorithm}
\end{figure}

\section{Description of $\hat R$ Statistics}
\label{sec:app_rhat}

$\hat R$ \citep{gelman1992inference} is a popular MCMC convergence diagnostic. It is defined in terms of some scalar function $\psi(\theta)$ of the Markov chain iterates $\{\theta_{mn} | m\in\{1,\ldots,M\}, n\in \{1,\ldots,N\}\}$, where $\theta_{mn}$ denotes the state of the $m$th of $M$ chains at iteration $n$ of $N$. Letting $\psi_{mn}\triangleq \psi(\theta_{mn})$, $\hat R$ is defined as follows:
\begin{gather}
\bar \psi_{m \cdot} \triangleq
\frac{1}{N}\sum_n \psi_{mn};
\quad \bar \psi_{\cdot \cdot} \triangleq
\frac{1}{MN}\sum_{m,n} \psi_{mn};
\\
\frac{B}{N}\triangleq \frac{1}{M-1}\sum_m
(\bar\psi_{m\cdot} - \bar\psi_{\cdot\cdot})^2;
\\
W\triangleq \frac{1}{M(N-1)}\sum_{m,n}
(\psi_{mn}-\bar\psi_{m\cdot})^2;
\\
\hat\sigma^2_+ \triangleq \frac{N-1}{N}W + \frac{B}{N};
\\
\hat R\triangleq \frac{M+1}{M}\frac{\hat\sigma^2_+}{W}
- \frac{N-1}{MN}.
\end{gather}
If the chains were initialized from their stationary distribution, then $\hat \sigma^2_+$ would be an unbiased estimate of the stationary distribution's variance.
$W$ is an estimate of the average within-chain variance; if the chains are stuck in isolated regions, then $W$ should be smaller than $\hat\sigma^2_+$, and $\hat R$ will be clearly larger than 1. The $\frac{M+1}{M}$ and $\frac{N-1}{MN}$ terms are there to account for sampling variability---they vanish as $N$ gets large if $W$ approaches $\hat\sigma^2_+$.

Since $\hat R$ is defined in terms of a function of interest $\psi$, we can compute it for many such functions.
In \autoref{sec:r_hat} we evaluated it for each weight and each predicted softmax probability in the test set.

\section{Marginal distributions of the weights}
\label{sec:app_marginals}

\label{sec:marginals}

\begin{figure}[t]
    \centering

    \hspace{-.5cm}
    \subfigure[ResNet-20-FRN]{
    \includegraphics[height=0.15\textwidth]{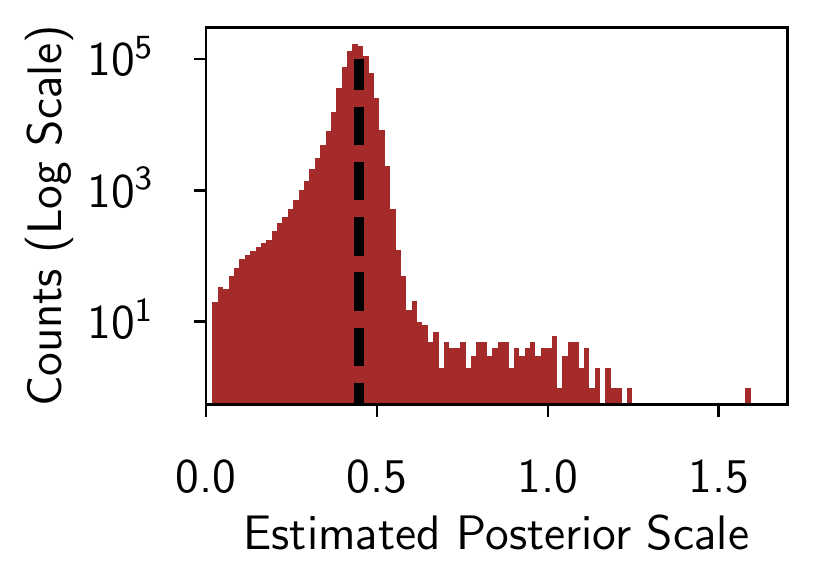}
	}
    \subfigure[CNN-LSTM]{
    \includegraphics[height=0.15\textwidth]{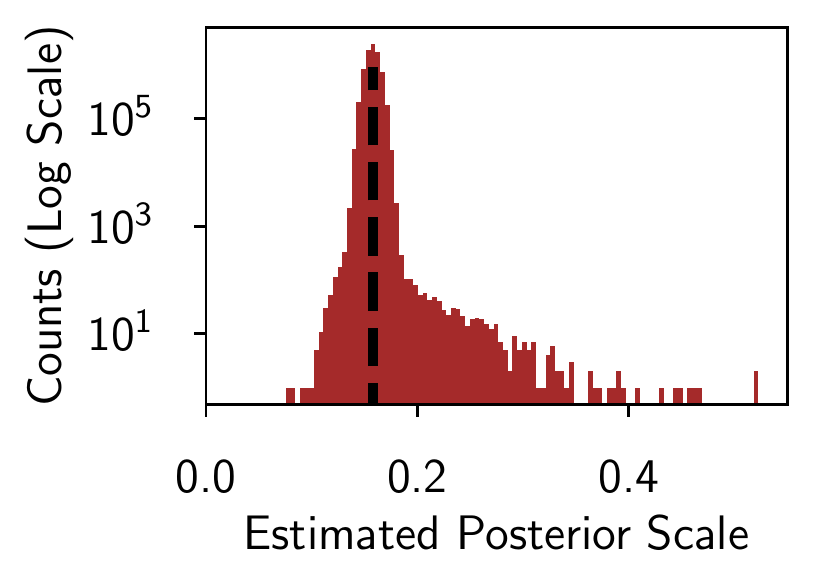}
	}
	\caption{
	    \textbf{Marginal distributions of the weights.}
	    Log-scale histograms of estimated marginal posterior standard deviations for ResNet-20-FRN on CIFAR-10 and CNN-LSTM on IMDB.
	    The histograms show how many parameters have empirical standard deviations that fall within a given bin.
	    For most of the parameters (notice that the plot is logarithmic) the posterior scale is very similar to that of the prior distribution. 
	}
	\label{fig:posterior_scales}
	\vspace{-0.2cm}
\end{figure}

In \autoref{sec:trajectory}, we argued for using a trajectory length $\tau=\frac{\pi\sigma_\mathrm{prior}}{2}$
based on the intuition that the posterior scale is
determined primarily by the prior scale.  
In \autoref{fig:posterior_scales} we examine this intuition.
For each parameter, we estimate the marginal standard deviation of that parameter under the distribution sampled by HMC.
Most of these marginal scales are close to the prior scale, and only a few are significantly larger (note logarithmic scale on y-axis), confirming that the posterior's scale is determined by the prior.

\begin{figure}[t]
    \centering
    \begin{tabular}{ccc}
    \hspace{-.5cm}
    \includegraphics[height=0.15\textwidth]{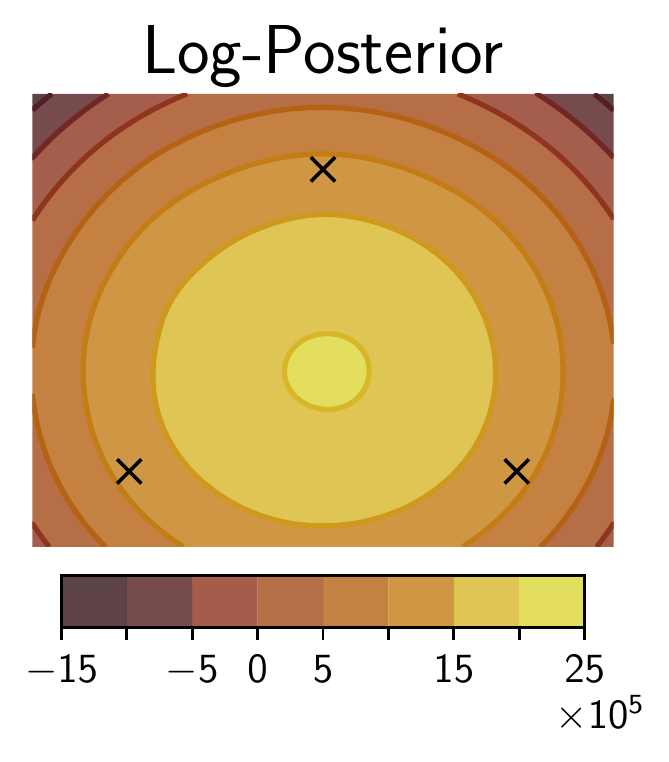}&
    \hspace{-.2cm}
    \includegraphics[height=0.15\textwidth]{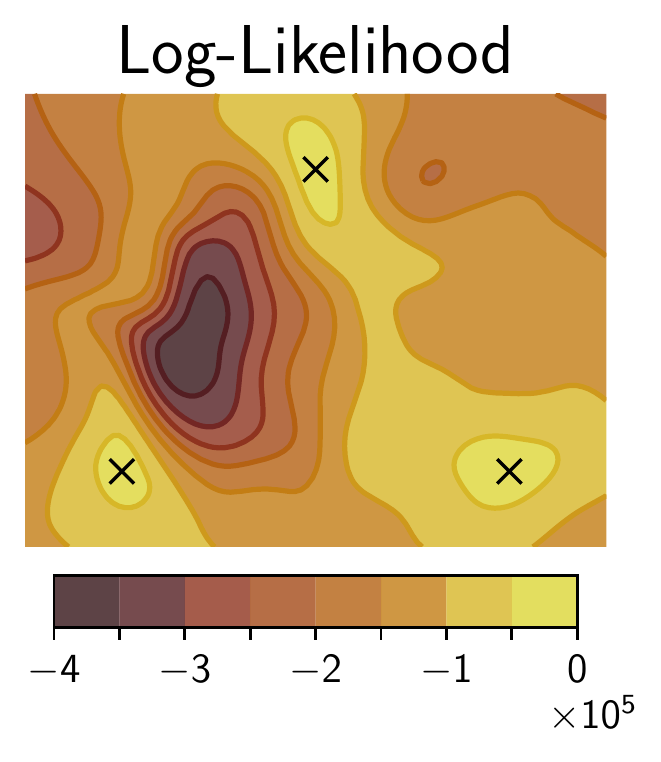}&
    \hspace{-.2cm}
    \includegraphics[height=0.15\textwidth]{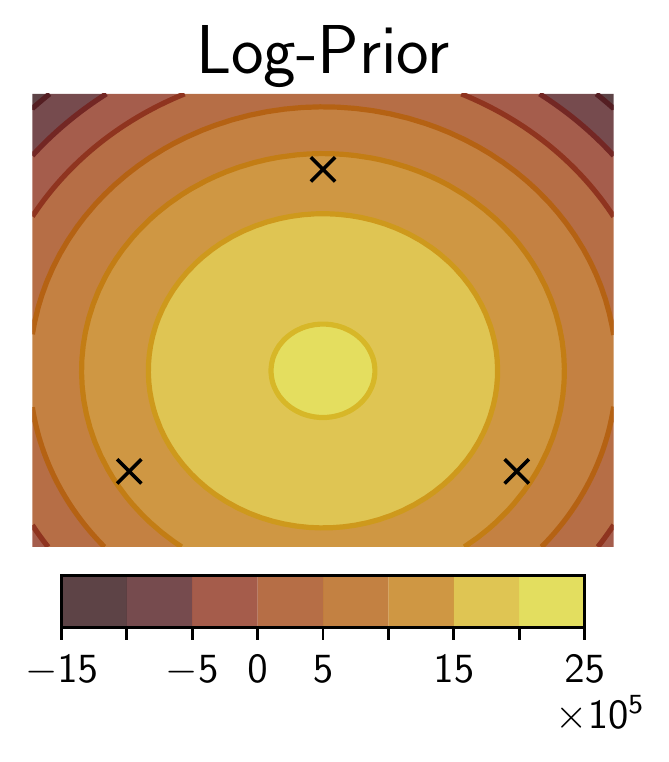}\\[-0.2cm]
    \multicolumn{3}{c}{\small (a) IMDB, same chain}\\
    \hspace{-.5cm}
    \includegraphics[height=0.15\textwidth]{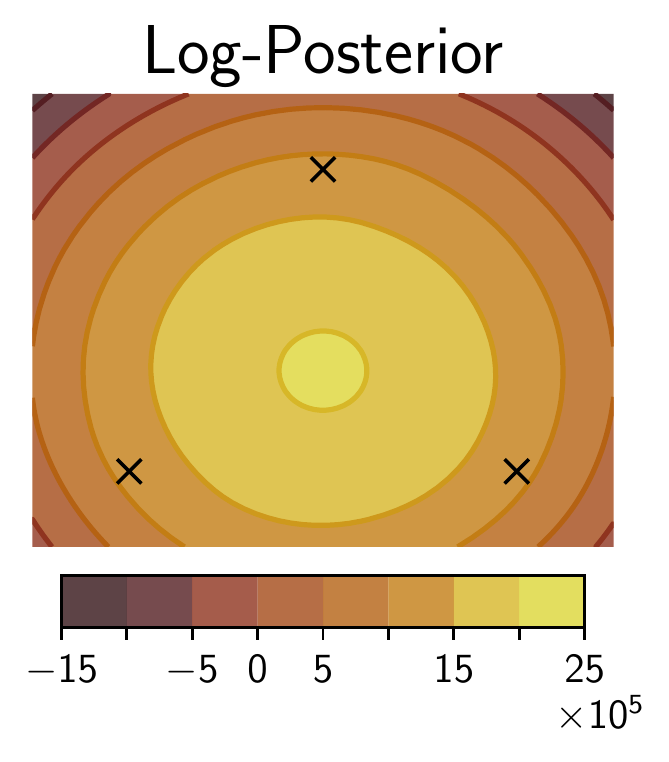}&
    \hspace{-.2cm}
    \includegraphics[height=0.15\textwidth]{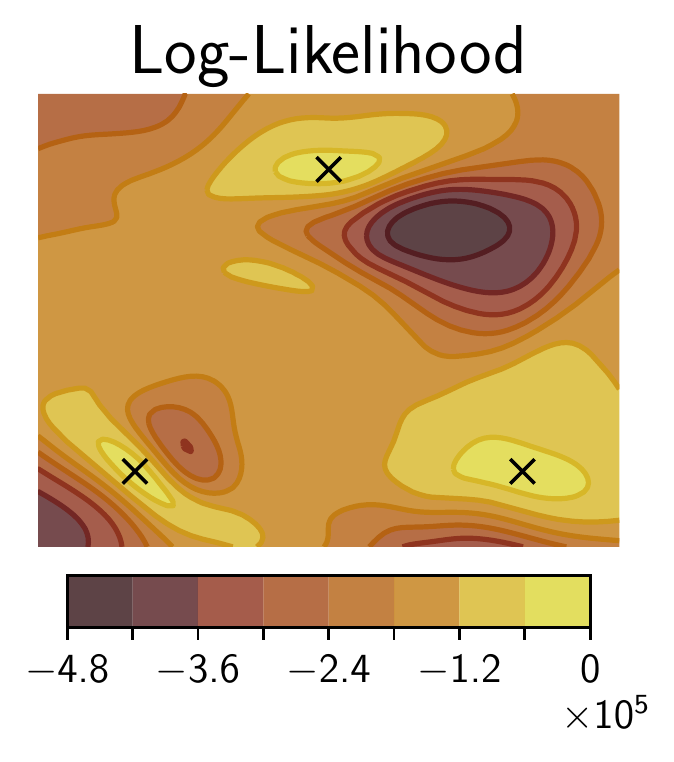}&
    \hspace{-.2cm}
    \includegraphics[height=0.15\textwidth]{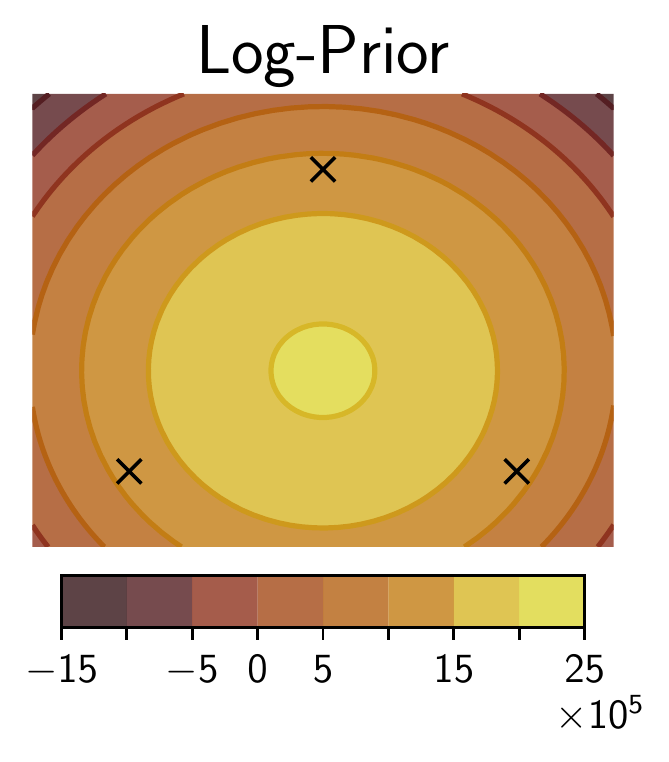}\\[-0.2cm]
    \multicolumn{3}{c}{\small (b) IMDB, independent chain}\\
    \hspace{-.5cm}
    \includegraphics[height=0.15\textwidth]{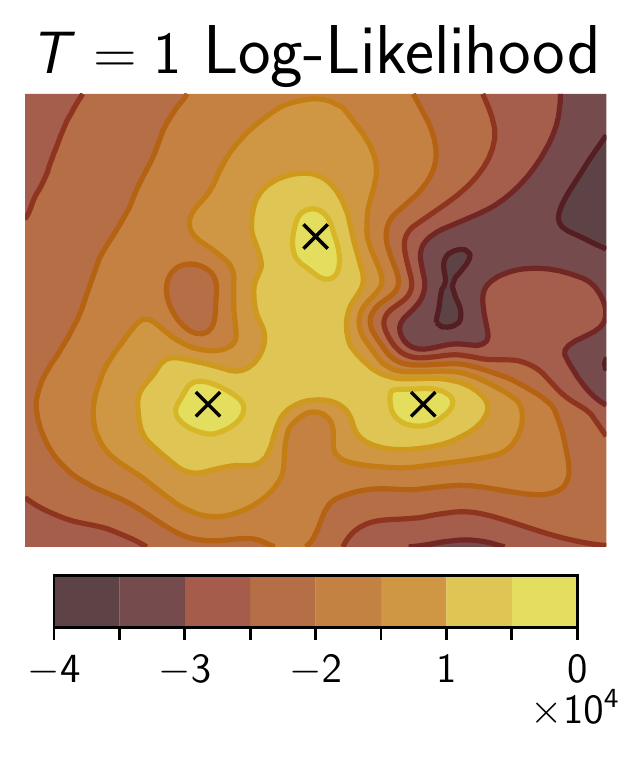} &
    \hspace{-.2cm}
    \includegraphics[height=0.15\textwidth]{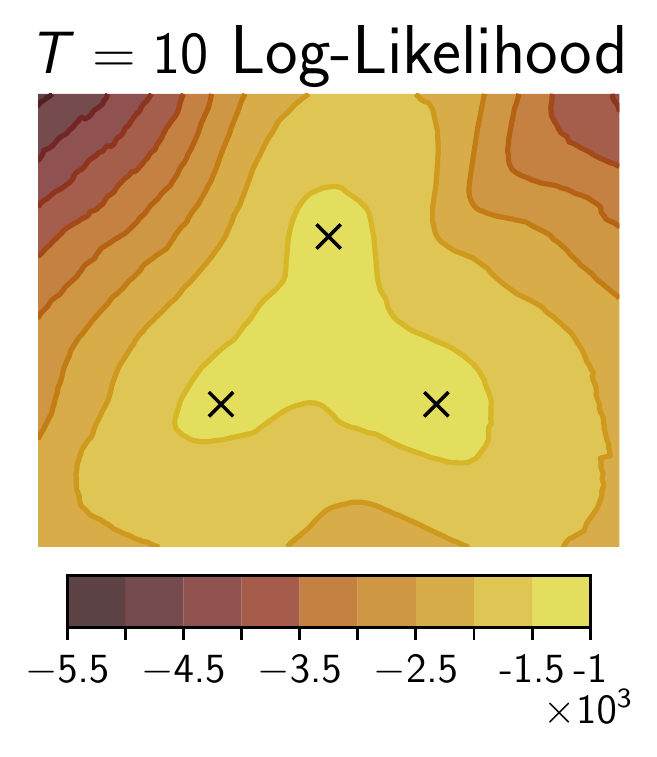} &
    \hspace{-.2cm}
    \includegraphics[height=0.15\textwidth]{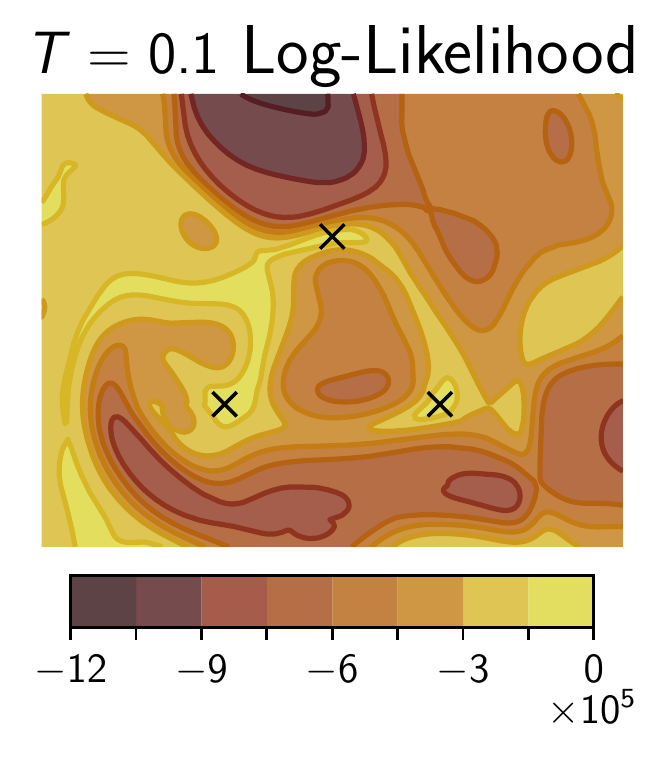}  \\
    \multicolumn{3}{c}{\small (c) IMDB, Log-Likelihood at different $T$}\\
    \end{tabular}
    
	\caption{
	    \textbf{Additional posterior density visualizations.}
        Visualizations of posterior log-density, log-likelihood and log-prior in two-dimensional subspaces
        of the parameter space spanned by three HMC samples on IMDB using CNN-LSTM.
        \textbf{(a):} samples from the same chain and \textbf{(b)}: independent chains;
        \textbf{(c):} Log-likelihood surfaces for samples from the same chain at posterior temperatures $T=1$, $10$ and $0.1$.
	}
	\label{fig:app_posterior_density}
	\vspace{-0.2cm}
\end{figure}

\section{Additional Posterior Visualizations}
\label{sec:app_surface_visualizations}

In \autoref{sec:surface_visualizations} we study two-dimensional cross-sections of posterior log-density, log-likelihood and log-prior surfaces.
We provide additional visualizations on the IMDB dataset in \autoref{fig:app_posterior_density}.

On IMDB, the posterior log-density is dominated by the prior, and the corresponding panels are virtually indistinguishable in \autoref{fig:app_posterior_density}.
For the CNN-LSTM on IMDB the number of parameters is much larger than the number of data points, and hence the scale of the prior density values is much larger than the scale of the likelihood.
Note that the likelihood still affects the posterior typical set, and the HMC samples land in the modes of the likelihood in the visualization.
In contrast, on ResNet-20, the number of parameters is smaller and the number of data points is larger, so the posterior is dominated by the likelihood in \autoref{fig:posterior_density}. 
The log-likelihood panels for both datasets show that HMC is able to navigate complex geometry: the samples fall in three isolated modes in our two-dimensional cross-sections.
On IMDB, the visualizations for samples from a single chain and for samples from three independent chains are qualitatively quite similar, hinting at better parameter-space mixing compared to CIFAR-10 
(see \autoref{sec:r_hat}). 

In \autoref{fig:app_posterior_density} (c), we visualize the likelihood cross-sections using our runs with varying posterior temperature on IMDB.
The visualizations show that, as expected, low temperature leads to a sharp likelihood, while the high-temperature likelihood appear soft.
In particular, the scale of the lowest likelihood values at $T=10$ is only $10^3$ while the scale at $T=0.1$ is $10^6$.

\textbf{How are the visualizations created?}\quad
To create the visualizations we pick the points in the parameter space corresponding to three HMC samples:
$w_1, w_2, w_3$. 
We construct a basis in the 2-dimensional affine subspace passing through these three points:
$u = w_2 - w_1$ and $v = w_3 - w_1$.
We then orthogonalize the basis: $\hat u = u / \|u\|$, $\hat v = (v - \hat u^T v) / \|v - \hat u^T v\|$.
We construct a 2-dimensional uniform grid in the basis $\hat u, \hat v$.
Each point in the grid corresponds to a vector of parameters of the network.
We evaluate the log-likelihood, log-prior and posterior log-density for each of the points in the grid, converting them to the corresponding network parameters.
Finally, we produce contour plots using the collected values.
The procedure is analogous to that used by \citet{garipov2018}\footnote{
See also the blogpost {\small \url{https://izmailovpavel.github.io/curves_blogpost/}}, Section "How to Visualize Loss Surfaces?".
}.

\section{HMC Predictive Distributions in Synthetic Regression}
\label{sec:app_synthreg}

\begin{figure}[h]
    \centering

    \hspace{-.5cm}
    \begin{tabular}{ccc}
    \hspace{-0.5cm}
    \includegraphics[height=0.17\textwidth]{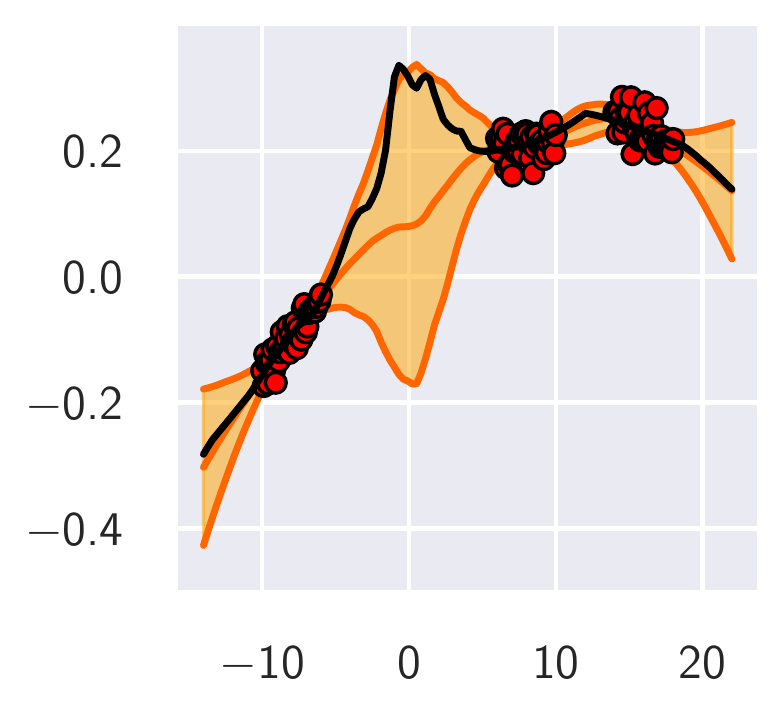}
	&
	\hspace{-0.5cm}
    \includegraphics[height=0.17\textwidth]{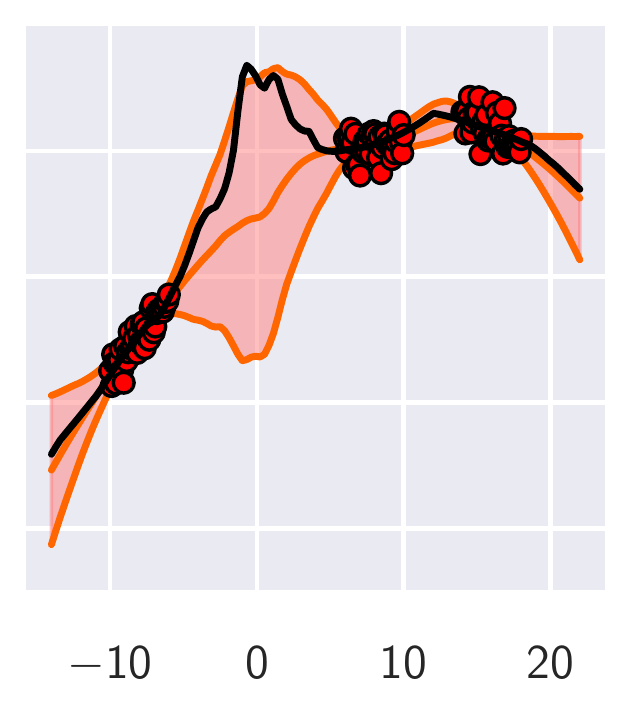}
	&
	\hspace{-0.5cm}
    \includegraphics[height=0.17\textwidth]{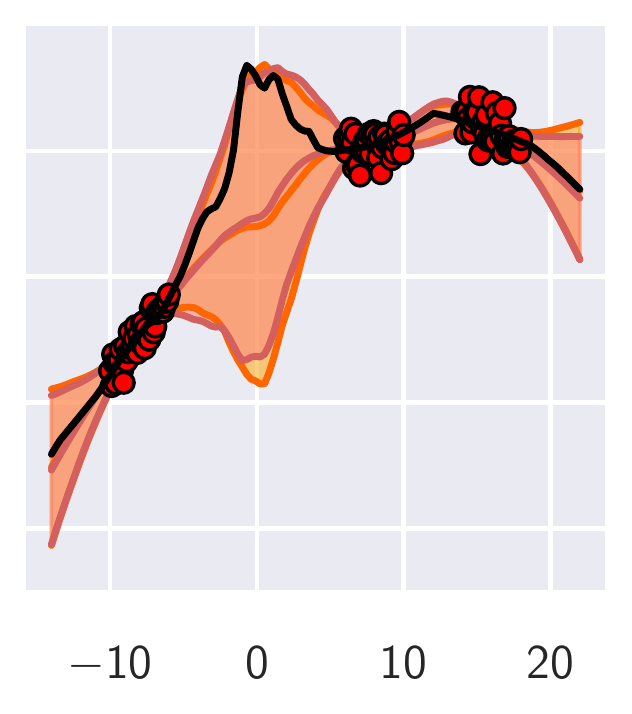}
    \\
    {\small (a) Chain 1}
    &
    {\small (b) Chain 2}
    &
    {\small (c) Overlaid}
	\end{tabular}
	\caption{
	    \textbf{HMC chains on synthetic regression.}
	    We visualize the predictive distributions for two independent HMC chains on a synthetic regression problem with a fully-connected network.
	    The data is shown with red circles, and the true data generating function is shown with a black line.
	    The shaded region shows $3$ standard deviations of the predictive distribution, and the predictive mean is shown
	    with a line of the same color. 
	    In panels \textbf{(a)}, \textbf{(b)} we show the predictive distributions for each of the two chains individually,
	    and in panel \textbf{(c)} we overlay them on top of each other.
	    The chains provide almost identical predictions, suggesting that HMC mixes well in the prediction space.
	}
	\label{fig:synthreg}
	\vspace{-0.2cm}
\end{figure}

We consider a one-dimensional synthetic regression problem. 
We follow the general setup of \citet{izmailov2019subspace} and \citet{wilson2020bayesian}.
We generate the training inputs as a uniform grid with $40$ points in each of the following intervals ($120$ datapoints in total):
$[-10, -6], [6, 10]$ and $[14, 18]$.
We construct the ground truth target values using a neural network with 3 hidden layers, each of dimension $100$, one output and
two inputs: following \citet{izmailov2019subspace}, for each datapoint $x$ we pass $x$ and $x^2$ as inputs to the network to enlarge 
the class of functions that the network can represent.
We draw the parameters of the network from a Gaussian distribution with mean $0$ and standard deviation $0.1$.
We show the sample function used to generate the target values as a black line in each of the panels in \autoref{fig:synthreg}.
We then add Gaussian noise with mean $0$ and standard deviation $0.02$ to each of the target values.
The final dataset used in the experiment is shown with red circles in \autoref{fig:synthreg}.

For inference, we use the same model architecture that was used to generate the data.
We sample the initialization parameters of the network from a Gaussian distribution with mean $0$ and standard deviation $0.005$.
We use a Gaussian distribution with mean zero and standard deviation $0.1$ as the prior over the parameters, same as the
distribution used to sample the parameters of the ground truth solution.
We use a Gaussian likelihood with standard deviation $0.02$, same as the noise distribution in the data.
We run two HMC chains from different random initializations.
Each chain uses a step-size of $10^{-5}$ and the trajectory length is set according to the strategy described in \autoref{sec:trajectory},
resulting in $15708$ leapfrog steps per HMC iteration.
We run each chain for $100$ HMC iterations and collect the predictions corresponding to all the accepted samples,
resulting in $89$ and $82$ samples for the first and second chain respectively.
We discard the first samples and only use the last $70$ samples from each chain.
For each input point we compute the mean and standard deviation of the predictions.

We report the results in \autoref{fig:synthreg}. 
In panels (a), (b) we show the predictive distributions for each of the chains, and in panel (c) we show them overlaid on top of  each other.
Both chains provide high uncertainty away from the data, and low uncertainty near the data as desired \citep{yao2019quality}.
Moreover, the true data-generating function lies in the $3\sigma$-region of the predictive distribution for each chain.
Finally, the predictive distributions for the two chains are almost identical.
This result suggests that on the synthetic problem under consideration HMC is able to mix in the space of predictions, 
and provides similar results independent of initialization and random seed.
We come to the same conclusion for more realistic problems in \autoref{sec:mixing}.

\begin{figure*}[t]
    \centering
\begin{tabular}{cc}
\begin{tabular}{ccc}
	\includegraphics[height=0.11\textwidth]{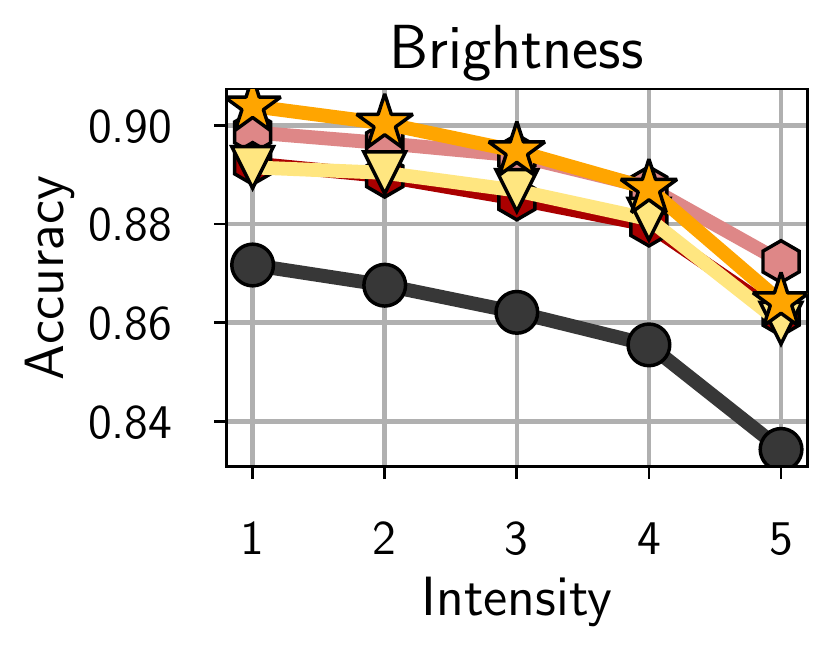} &
	\includegraphics[height=0.11\textwidth]{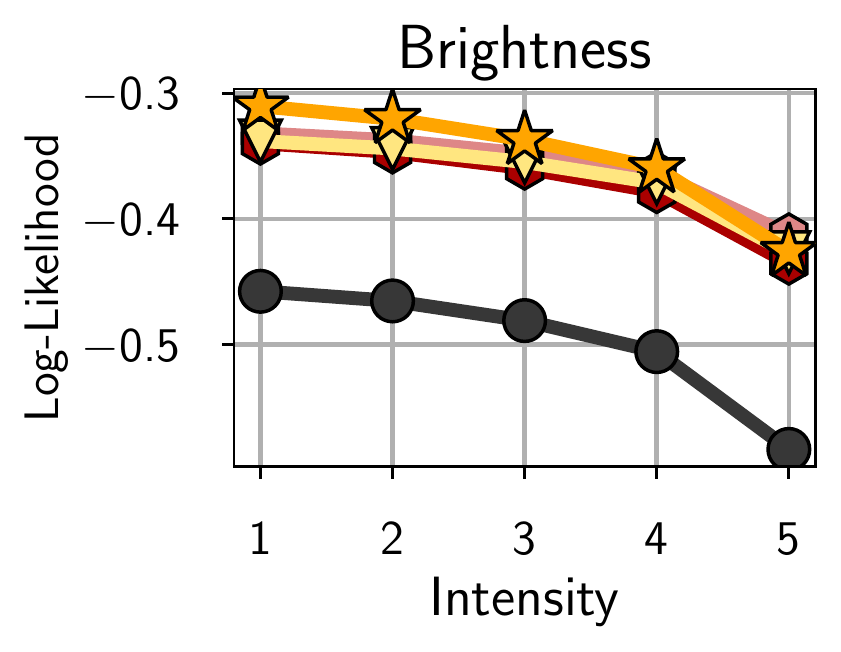} &
	\includegraphics[height=0.11\textwidth]{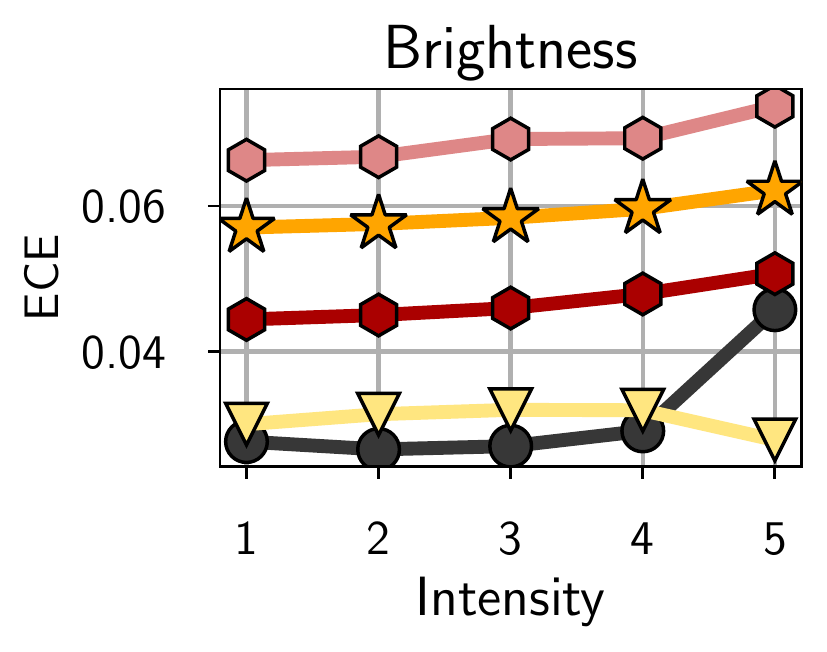}
	\\[-0.2cm]
	\includegraphics[height=0.11\textwidth]{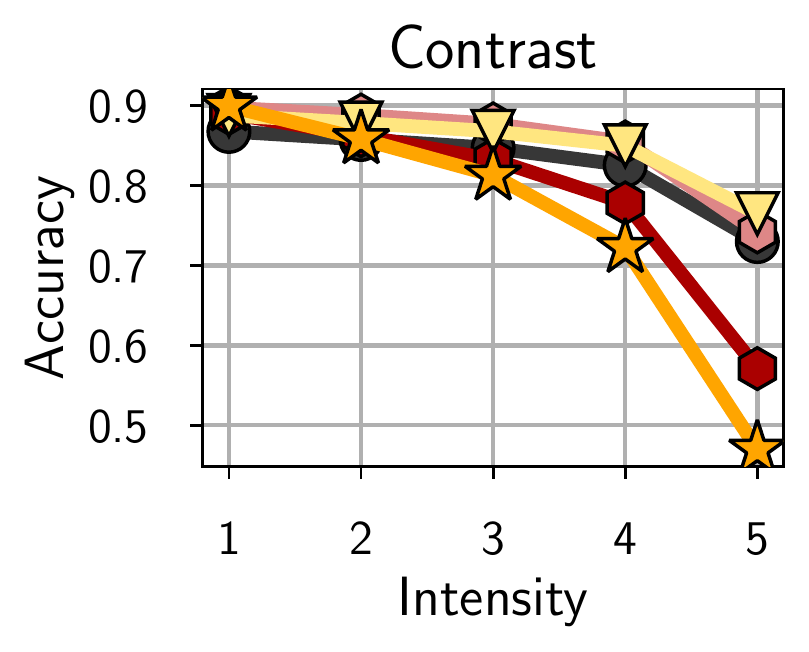} &
	\includegraphics[height=0.11\textwidth]{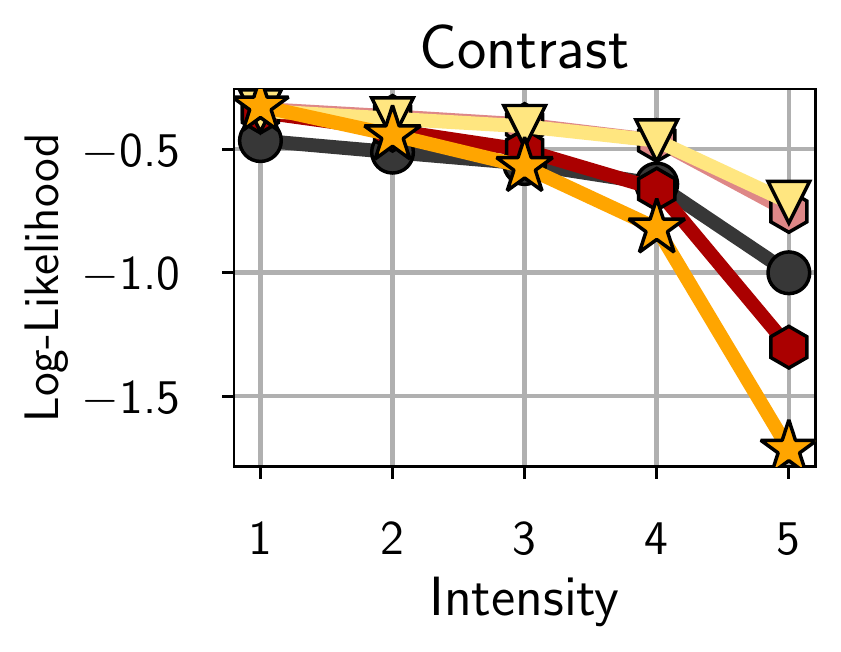} &
	\includegraphics[height=0.11\textwidth]{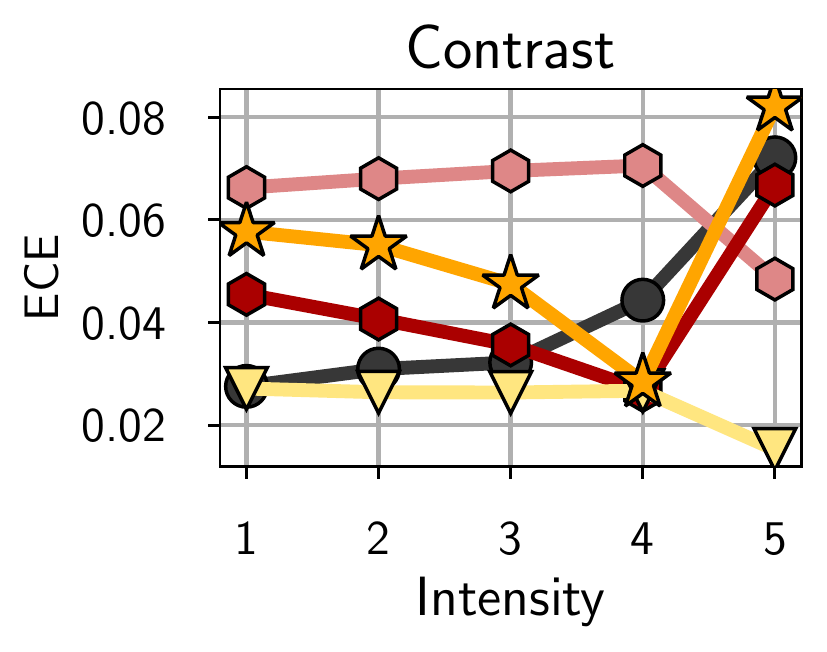}
	\\[-0.2cm]
	\includegraphics[height=0.11\textwidth]{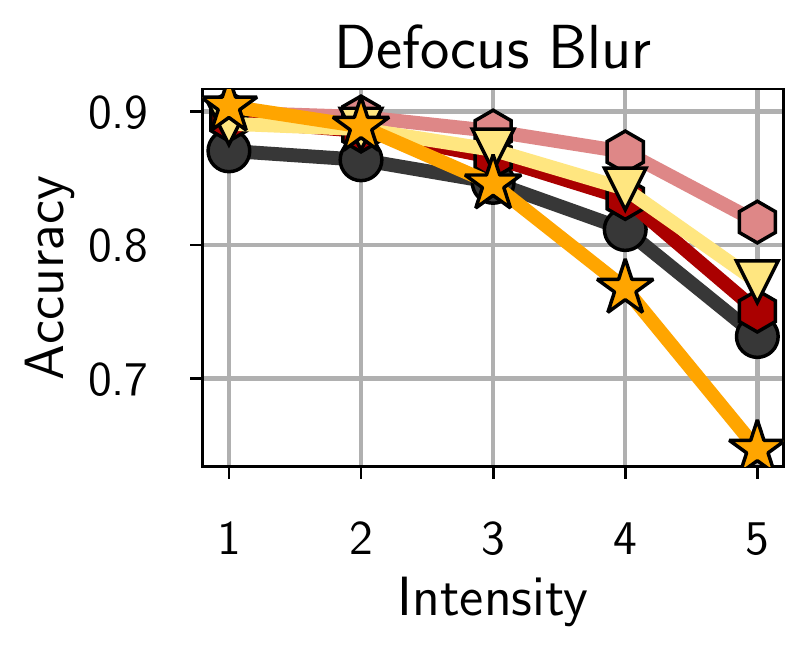} &
	\includegraphics[height=0.11\textwidth]{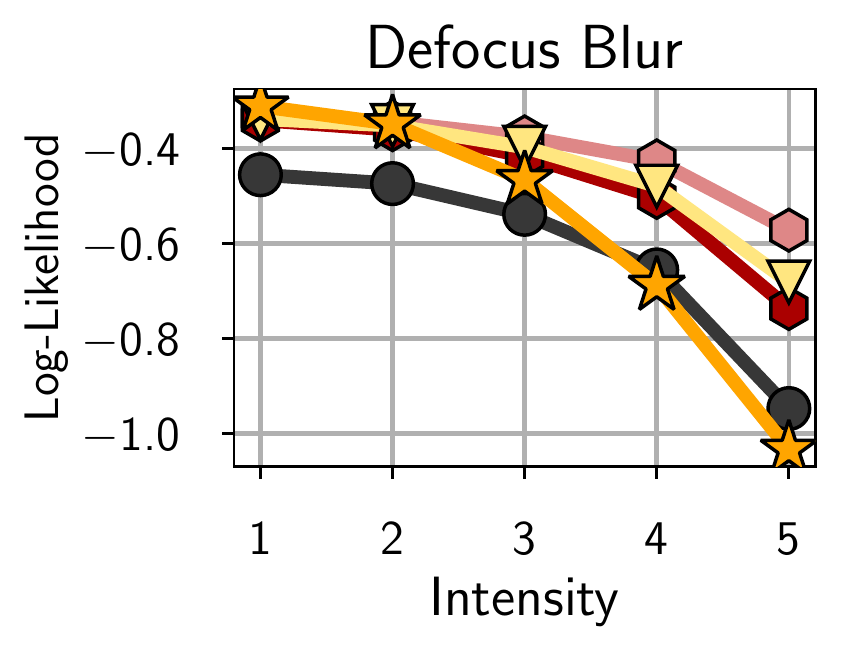} &
	\includegraphics[height=0.11\textwidth]{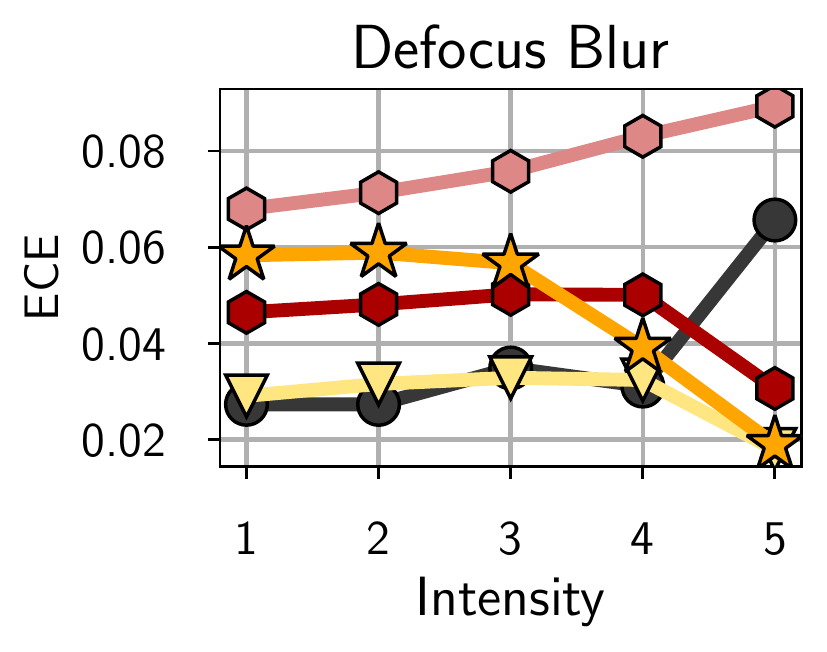}
	\\[-0.2cm]
	\includegraphics[height=0.11\textwidth]{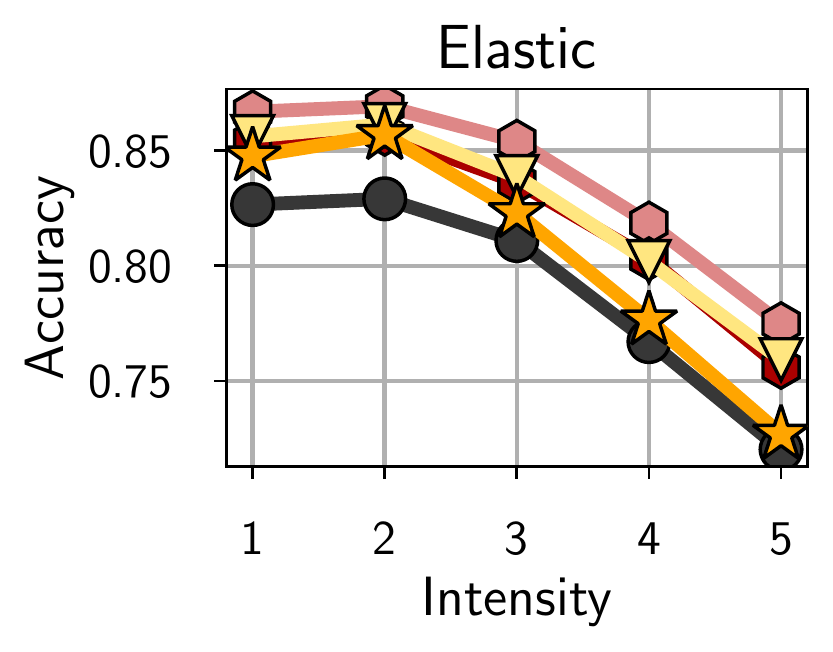} &
	\includegraphics[height=0.11\textwidth]{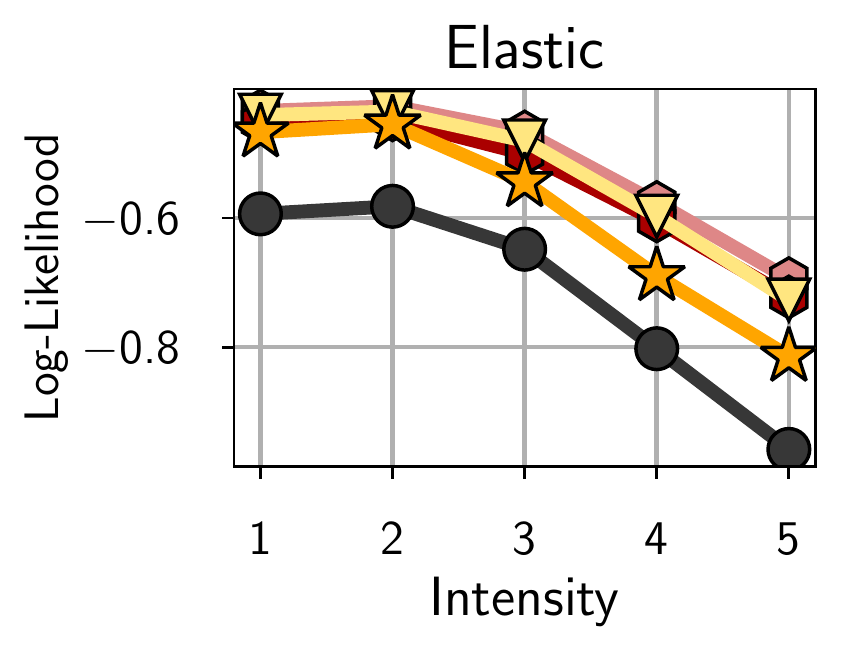} &
	\includegraphics[height=0.11\textwidth]{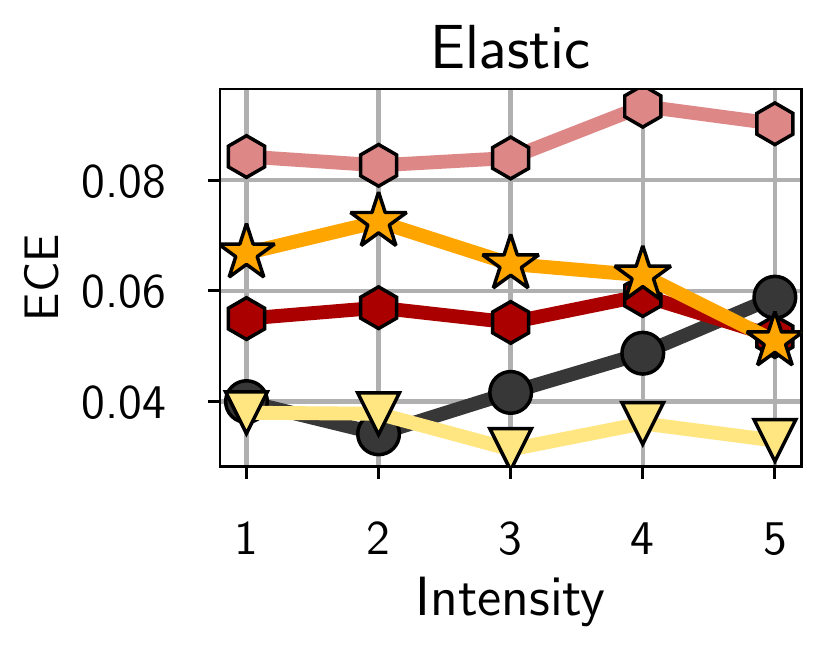}
	\\[-0.2cm]
	\includegraphics[height=0.11\textwidth]{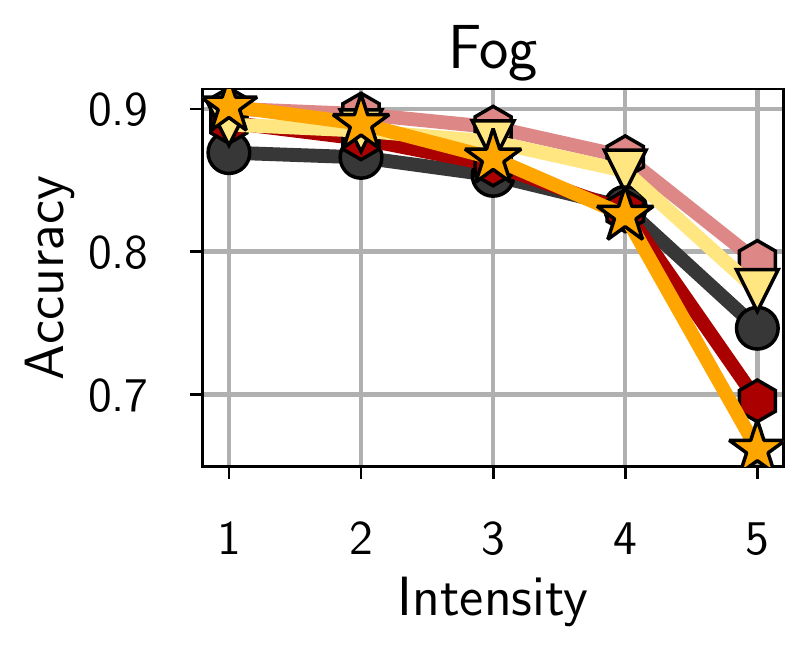} &
	\includegraphics[height=0.11\textwidth]{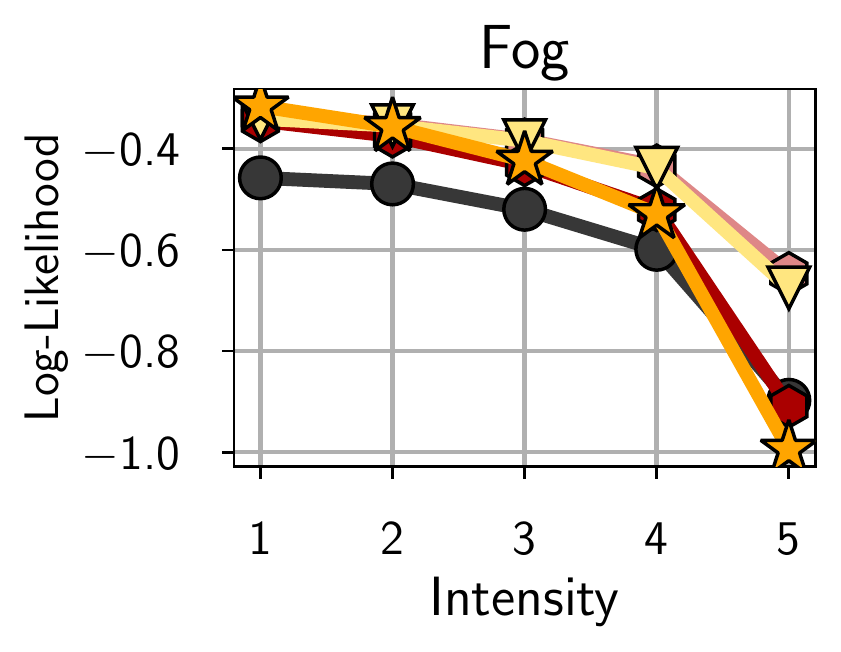} &
	\includegraphics[height=0.11\textwidth]{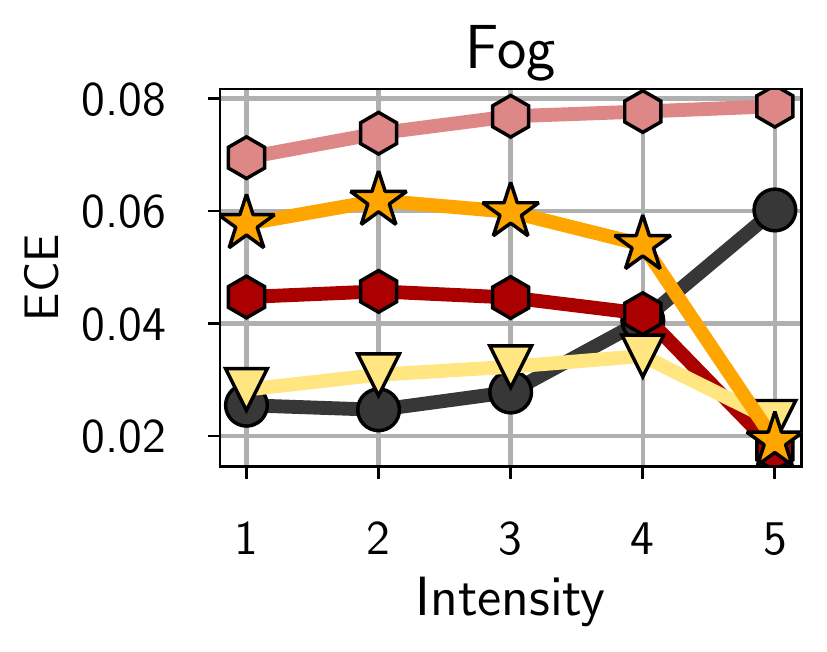}
	\\[-0.2cm]
	\includegraphics[height=0.11\textwidth]{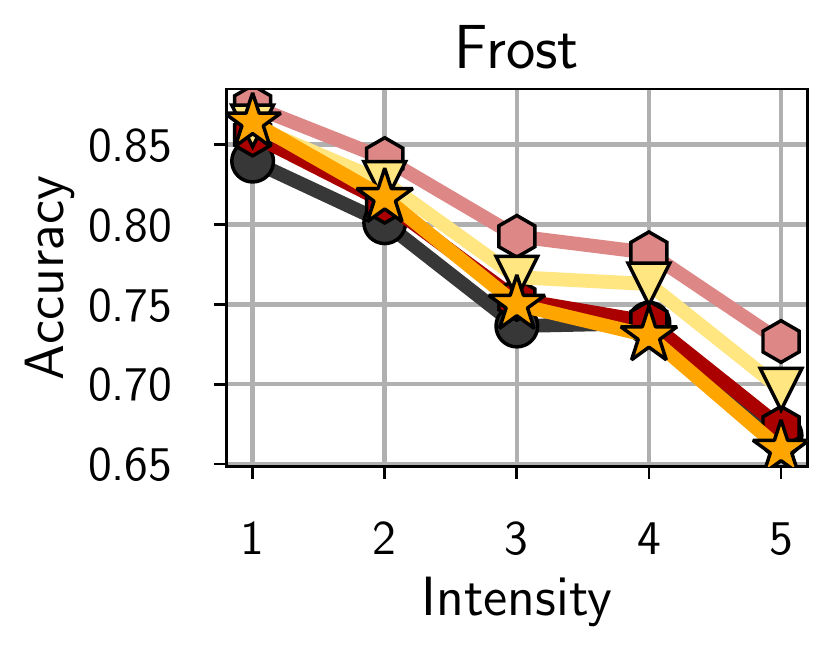} &
	\includegraphics[height=0.11\textwidth]{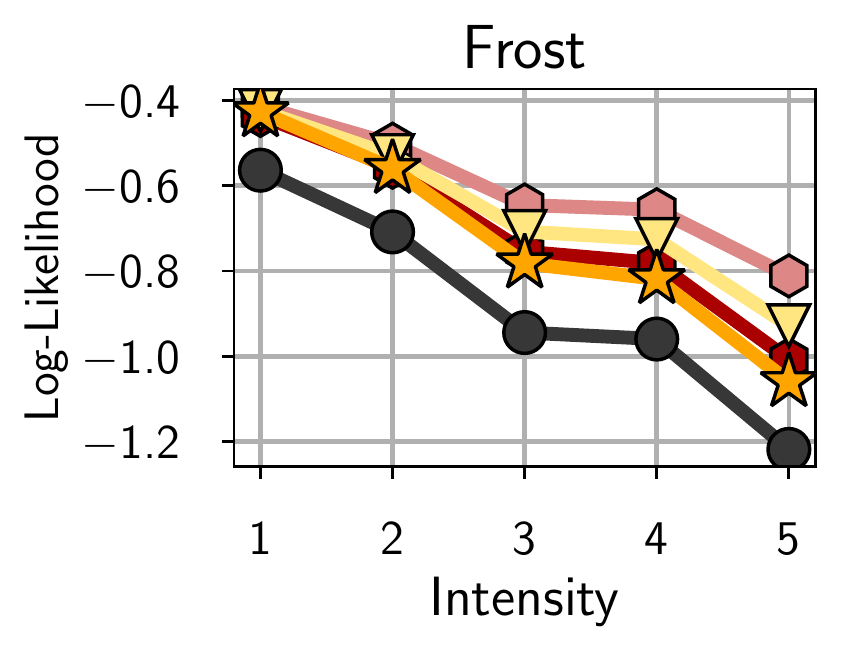} &
	\includegraphics[height=0.11\textwidth]{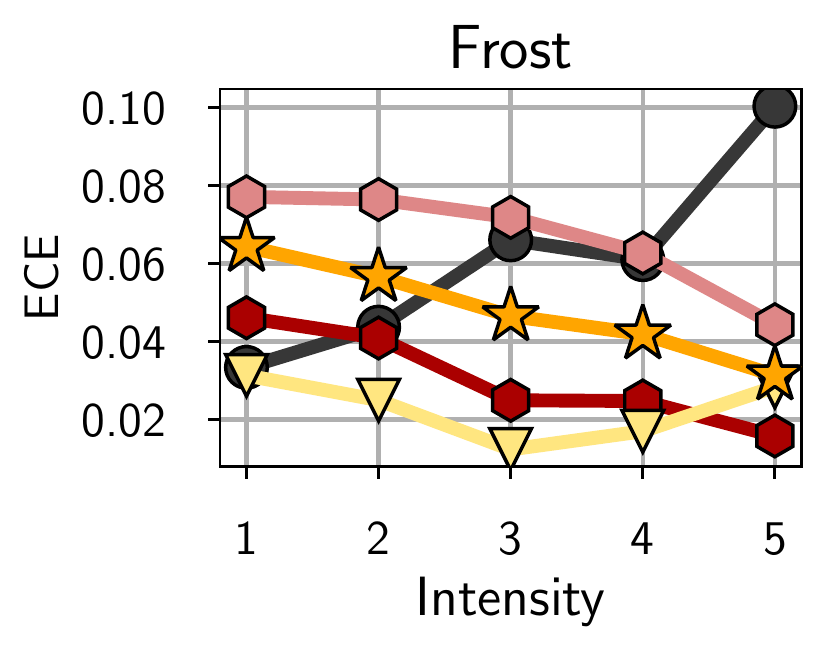}
	\\[-0.2cm]
	\includegraphics[height=0.11\textwidth]{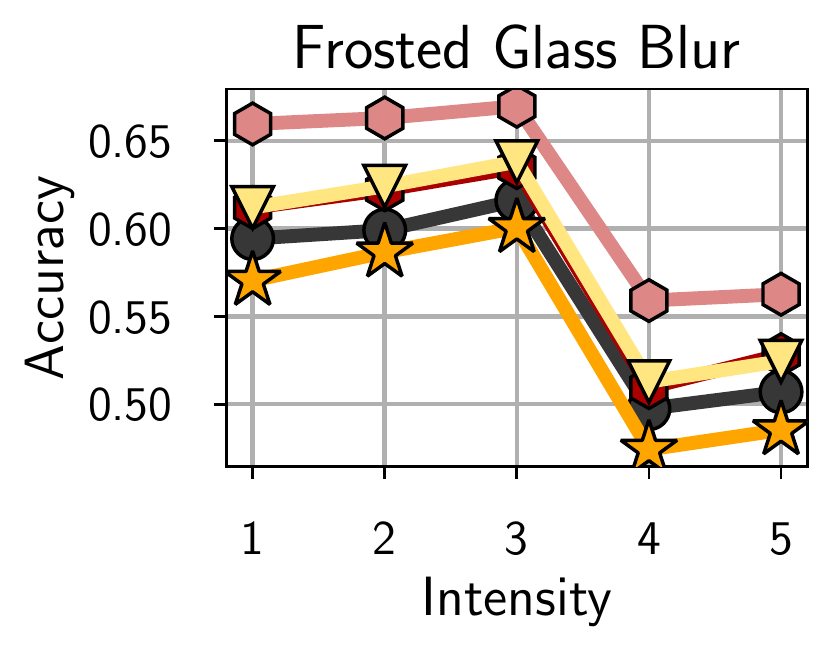} &
	\includegraphics[height=0.11\textwidth]{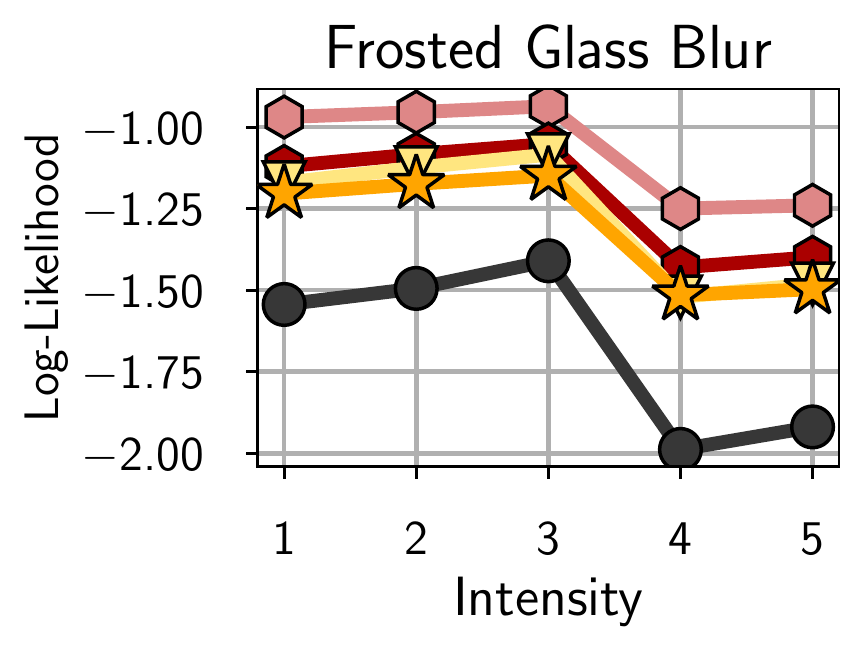} &
	\includegraphics[height=0.11\textwidth]{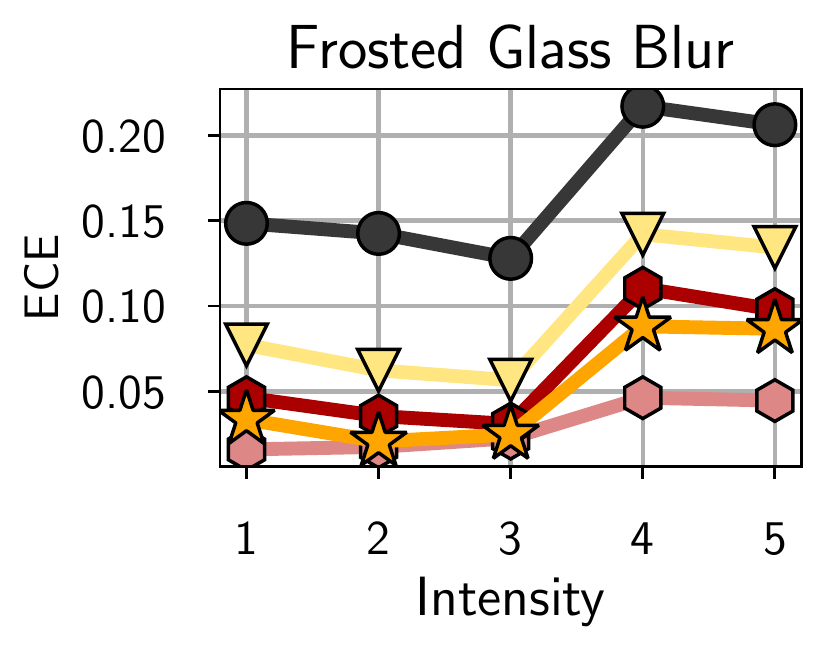}
	\\[-0.2cm]
	\includegraphics[height=0.11\textwidth]{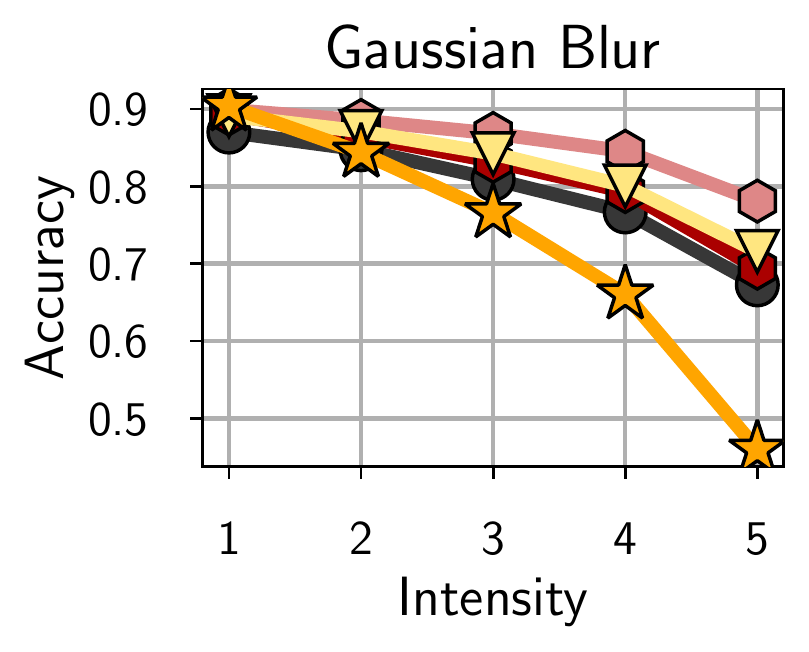} &
	\includegraphics[height=0.11\textwidth]{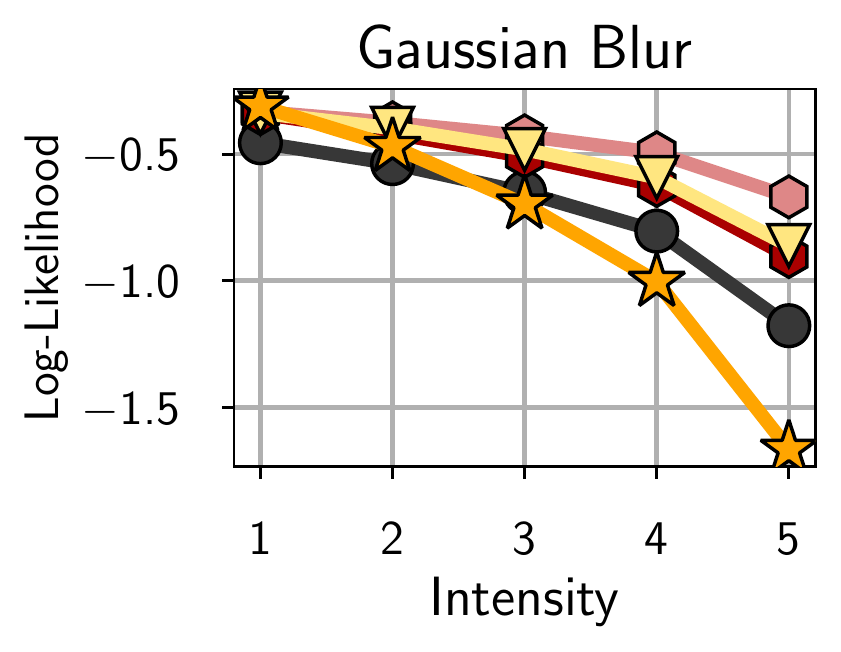} &
	\includegraphics[height=0.11\textwidth]{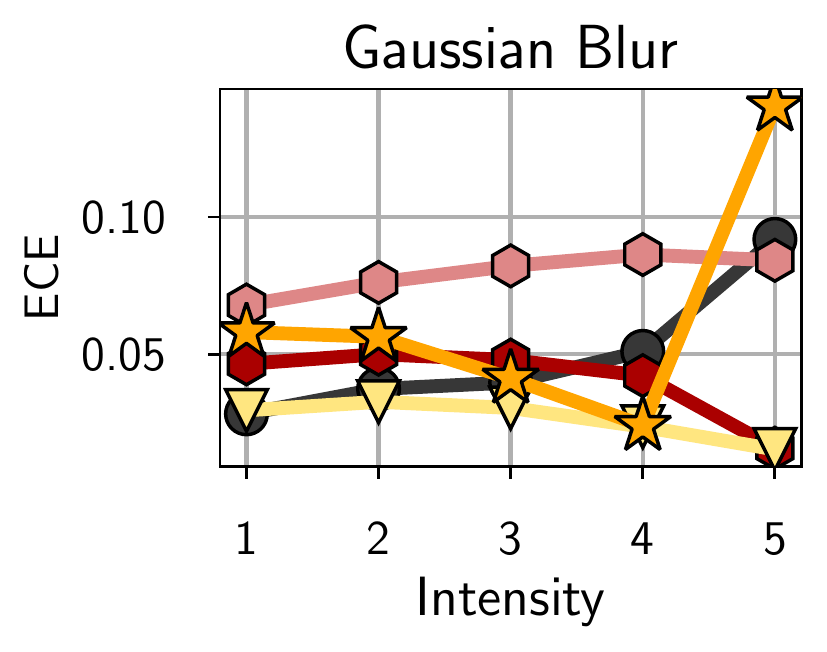}
	\\[-0.2cm]
\end{tabular} & \begin{tabular}{ccc}
	\includegraphics[height=0.11\textwidth]{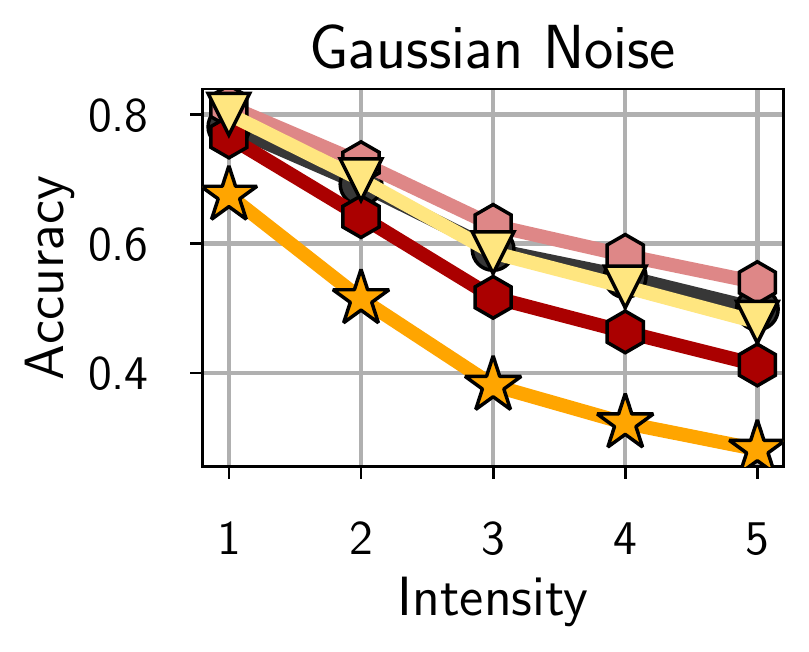} &
	\includegraphics[height=0.11\textwidth]{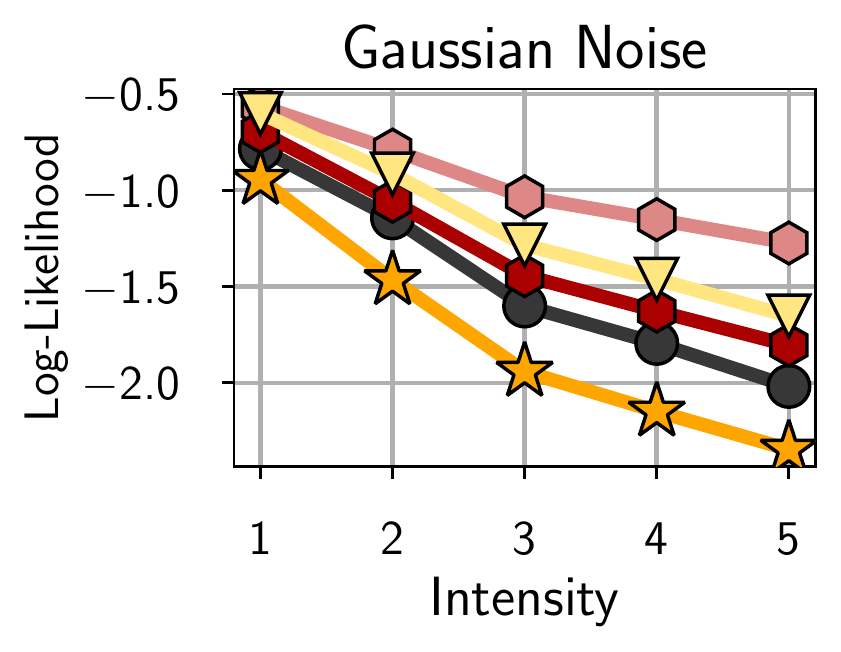} &
	\includegraphics[height=0.11\textwidth]{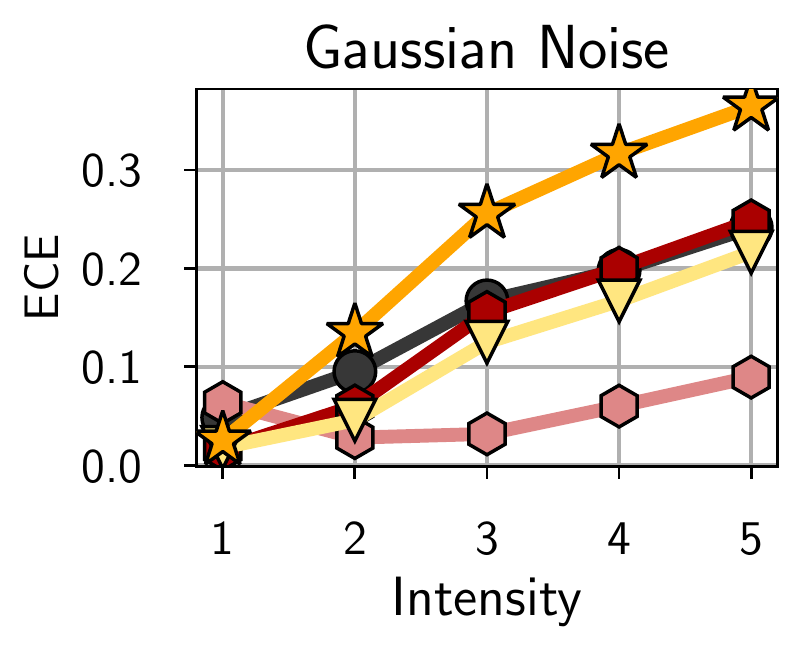}
	\\[-0.2cm]
	\includegraphics[height=0.11\textwidth]{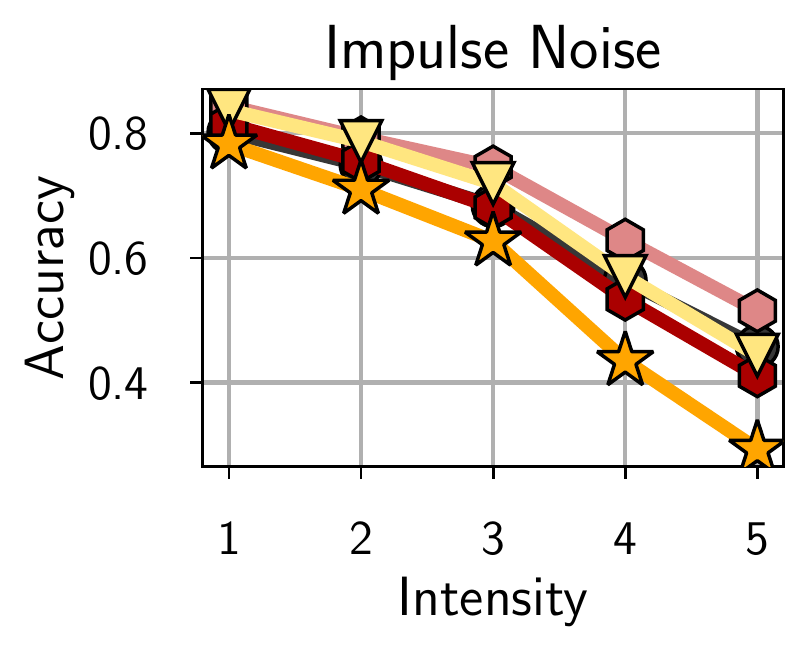} &
	\includegraphics[height=0.11\textwidth]{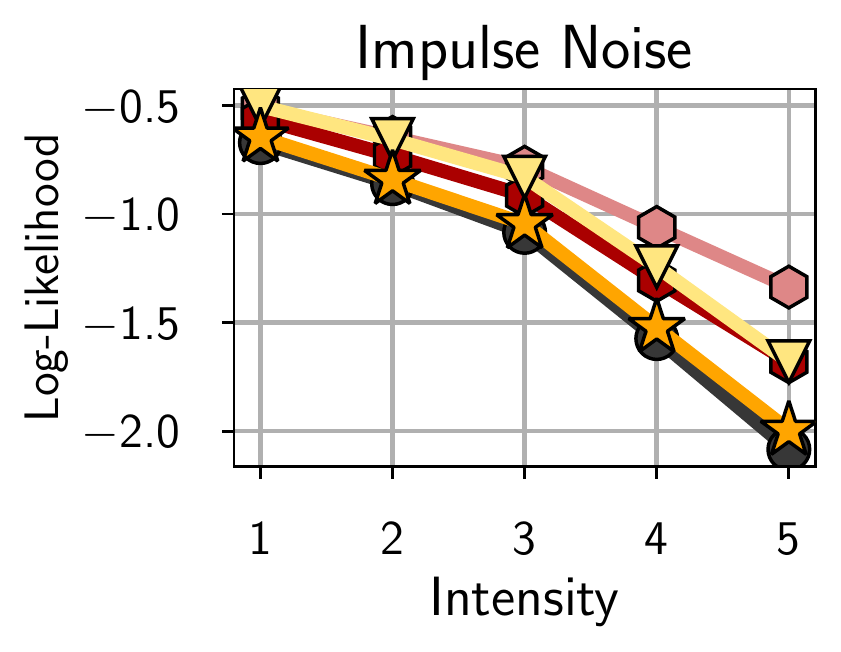} &
	\includegraphics[height=0.11\textwidth]{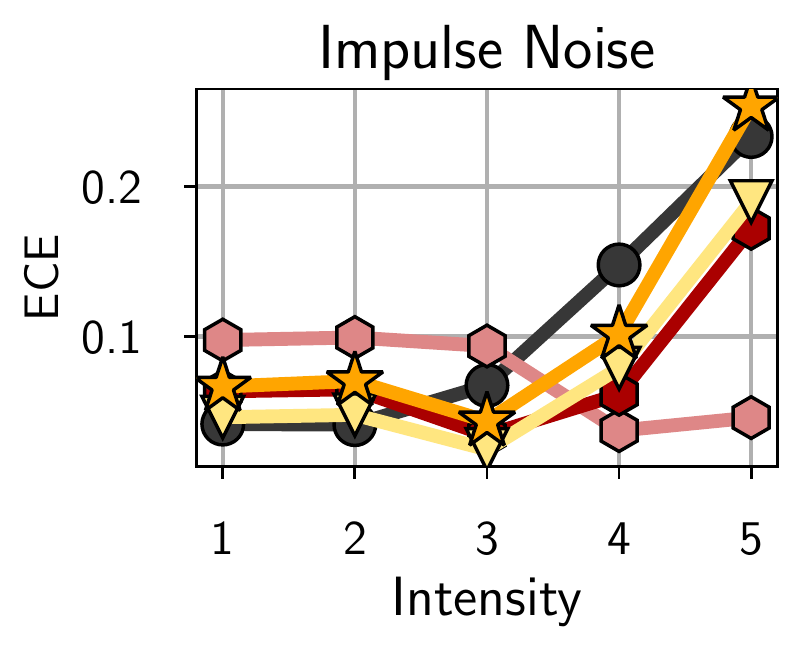}
	\\[-0.2cm]
	\includegraphics[height=0.11\textwidth]{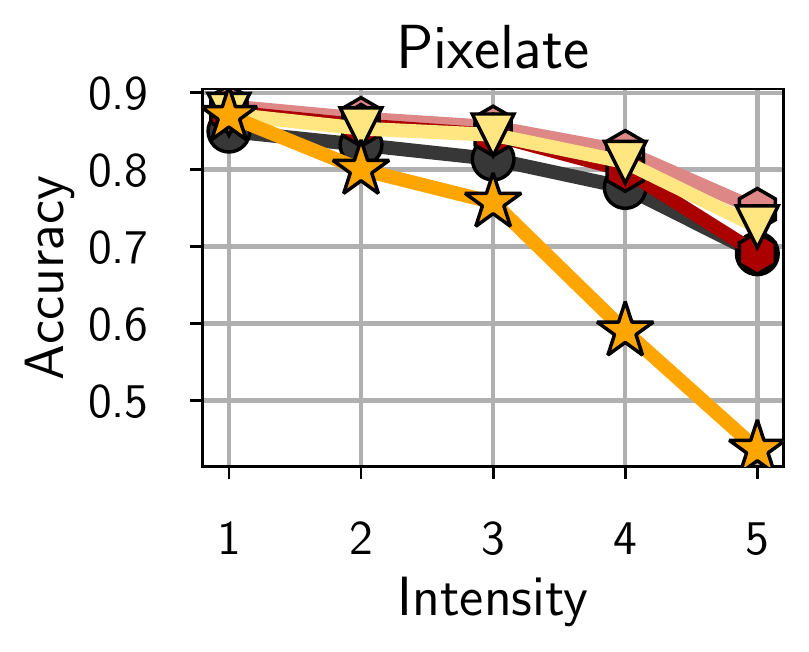} &
	\includegraphics[height=0.11\textwidth]{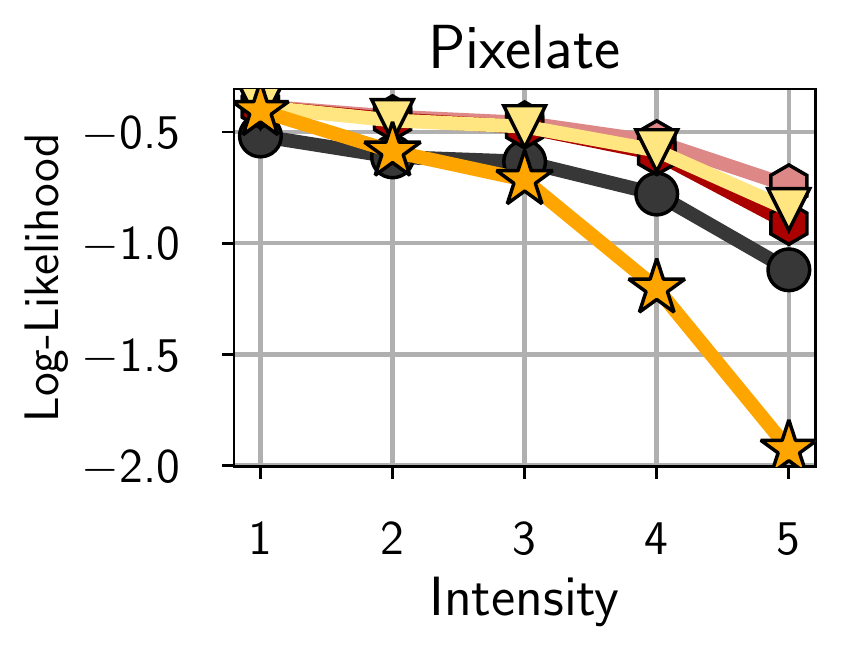} &
	\includegraphics[height=0.11\textwidth]{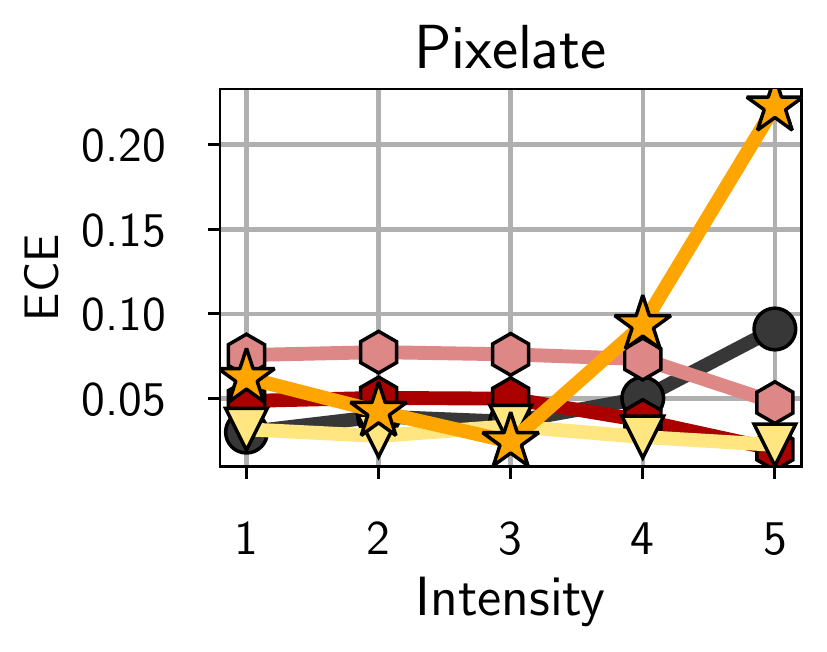}
	\\[-0.2cm]
	\includegraphics[height=0.11\textwidth]{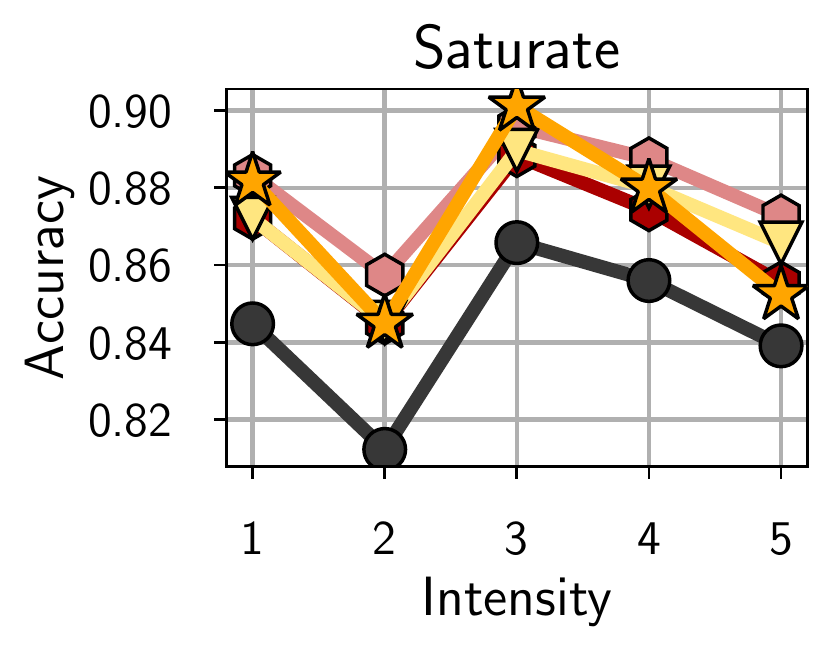} &
	\includegraphics[height=0.11\textwidth]{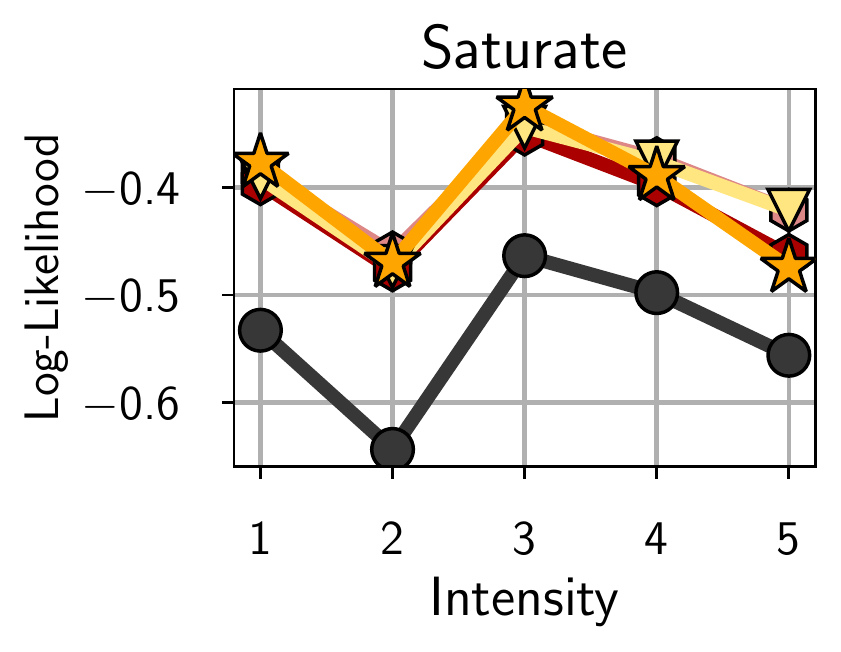} &
	\includegraphics[height=0.11\textwidth]{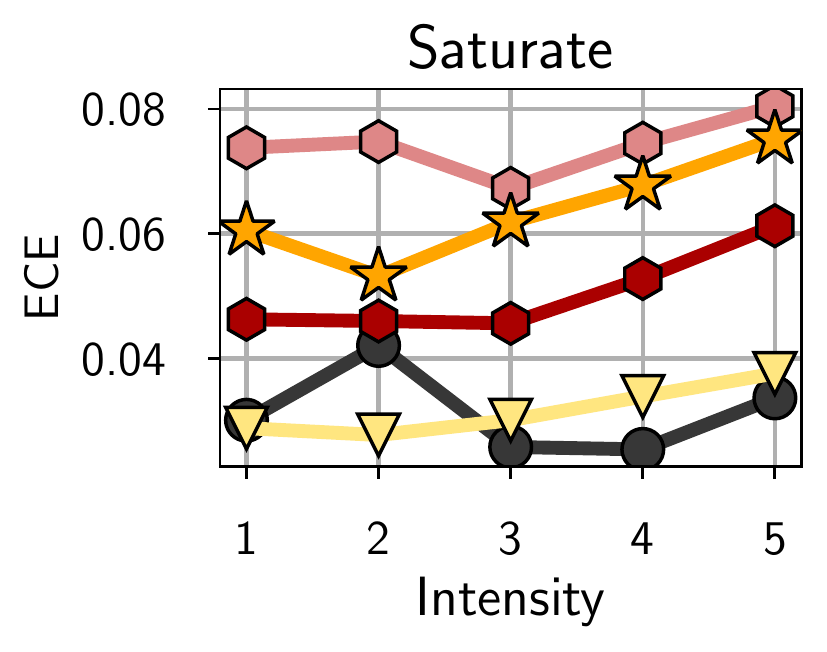}
	\\[-0.2cm]
	\includegraphics[height=0.11\textwidth]{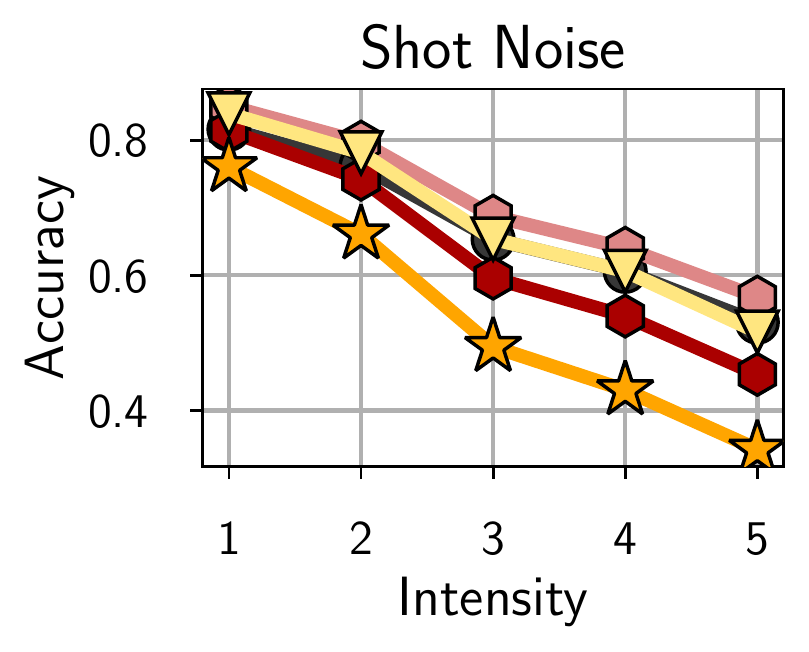} &
	\includegraphics[height=0.11\textwidth]{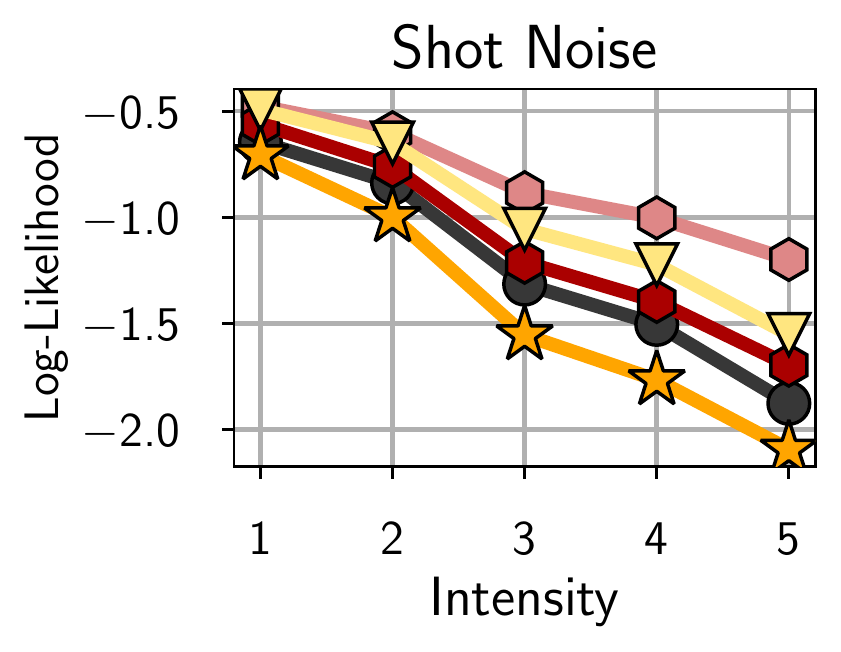} &
	\includegraphics[height=0.11\textwidth]{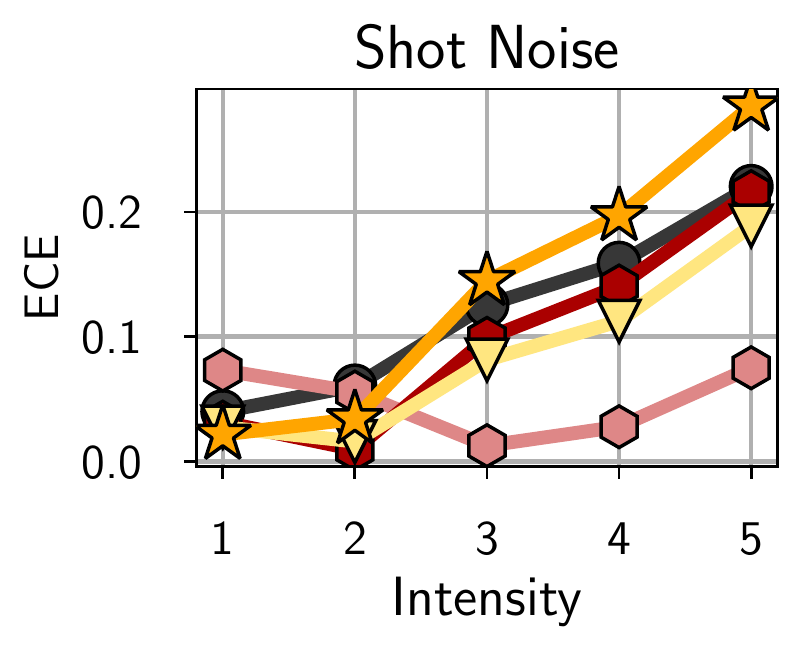}
	\\[-0.2cm]
	\includegraphics[height=0.11\textwidth]{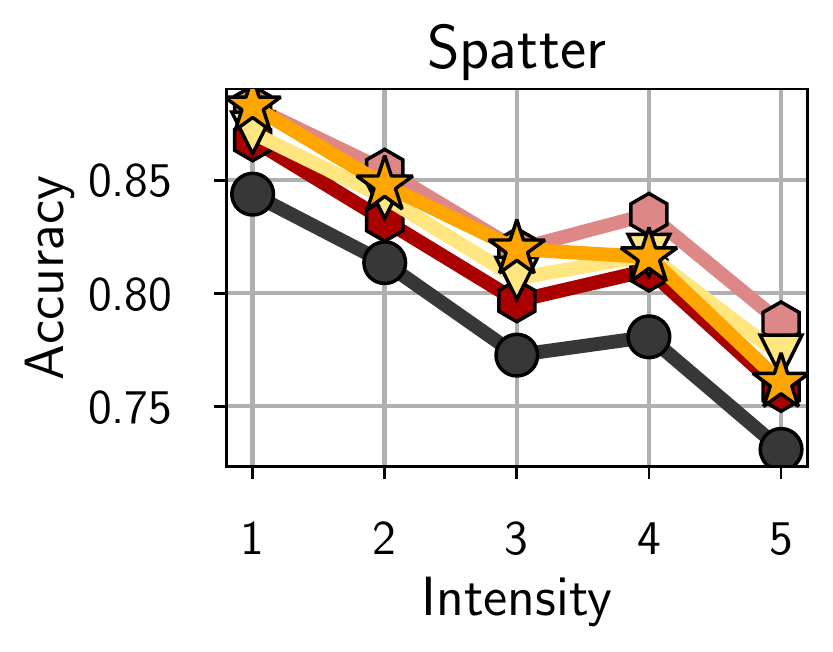} &
	\includegraphics[height=0.11\textwidth]{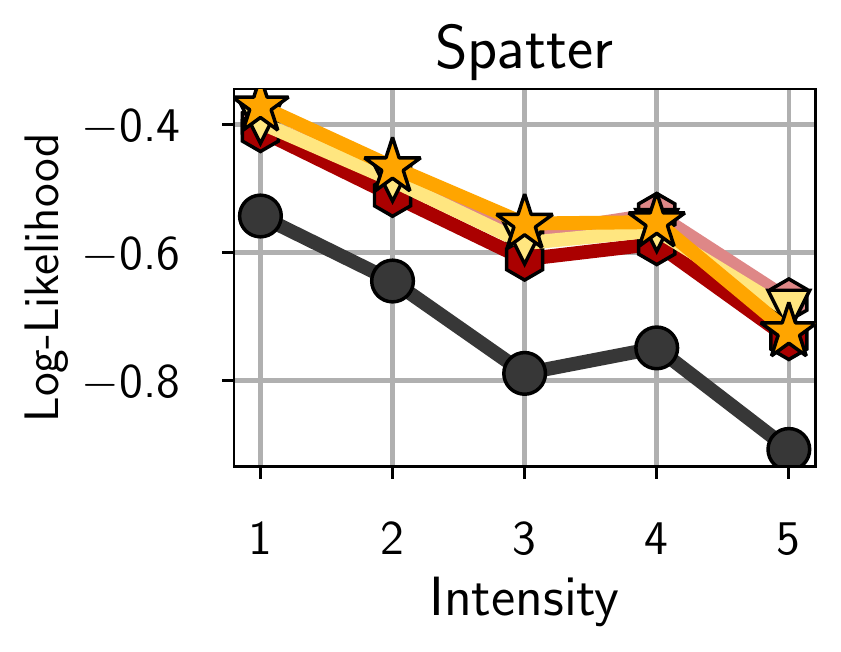} &
	\includegraphics[height=0.11\textwidth]{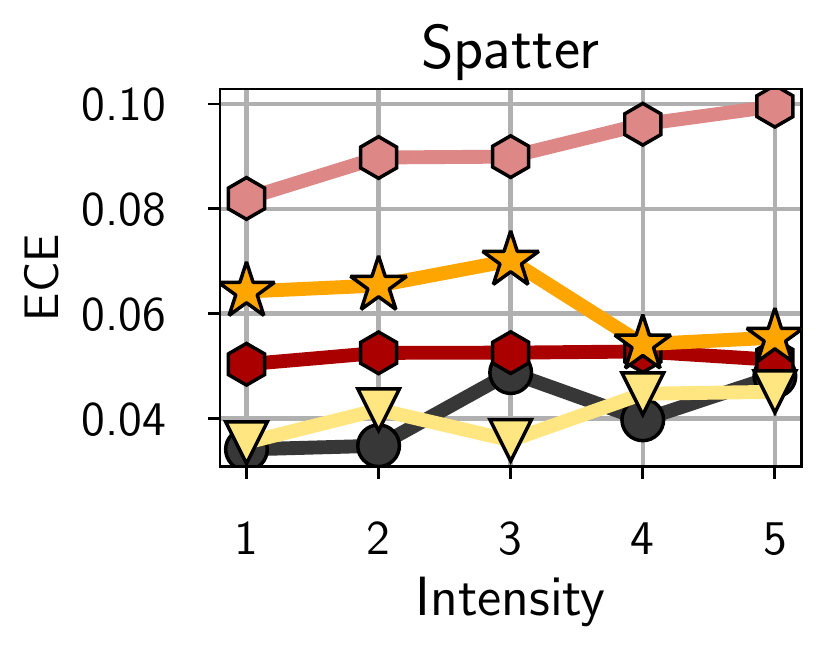}
	\\[-0.2cm]
	\includegraphics[height=0.11\textwidth]{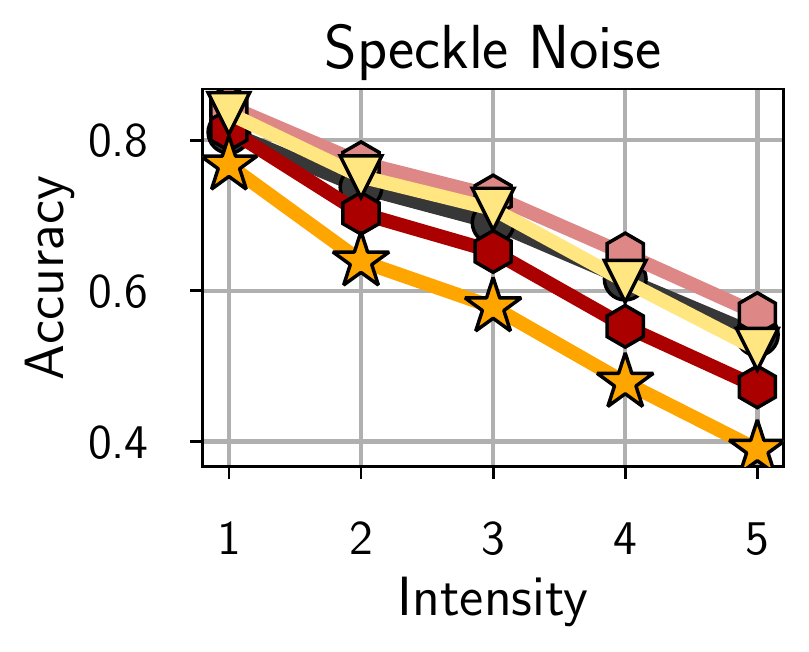} &
	\includegraphics[height=0.11\textwidth]{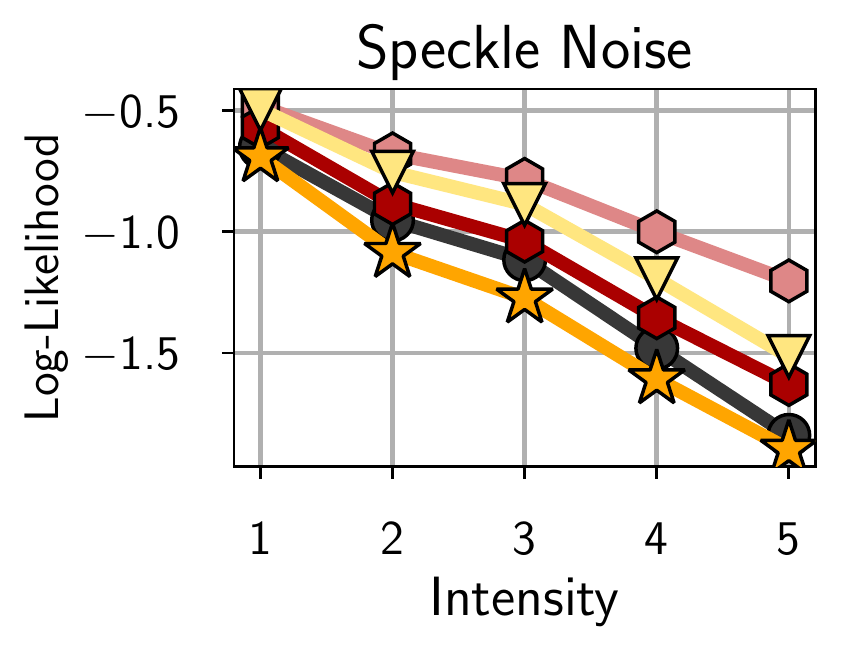} &
	\includegraphics[height=0.11\textwidth]{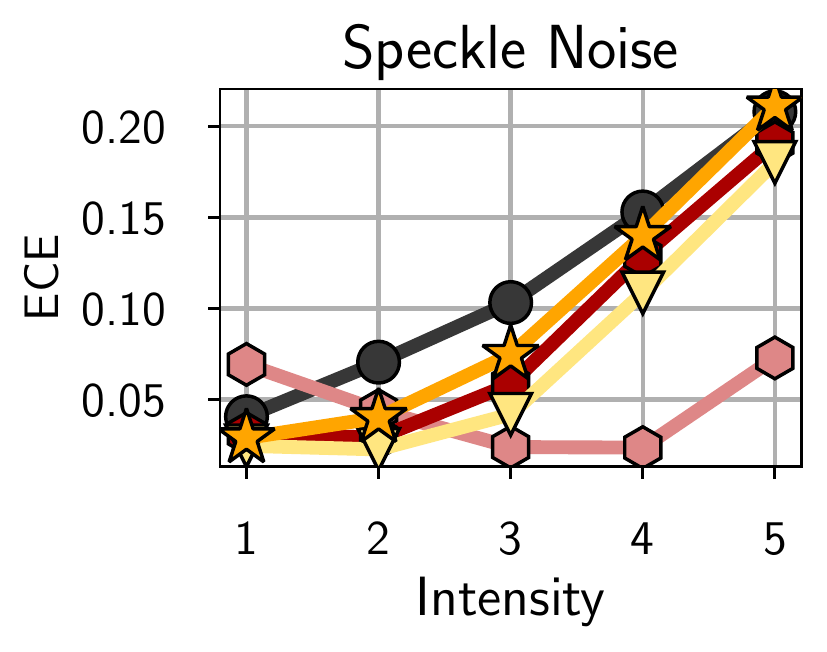}
	\\[-0.2cm]
	\includegraphics[height=0.11\textwidth]{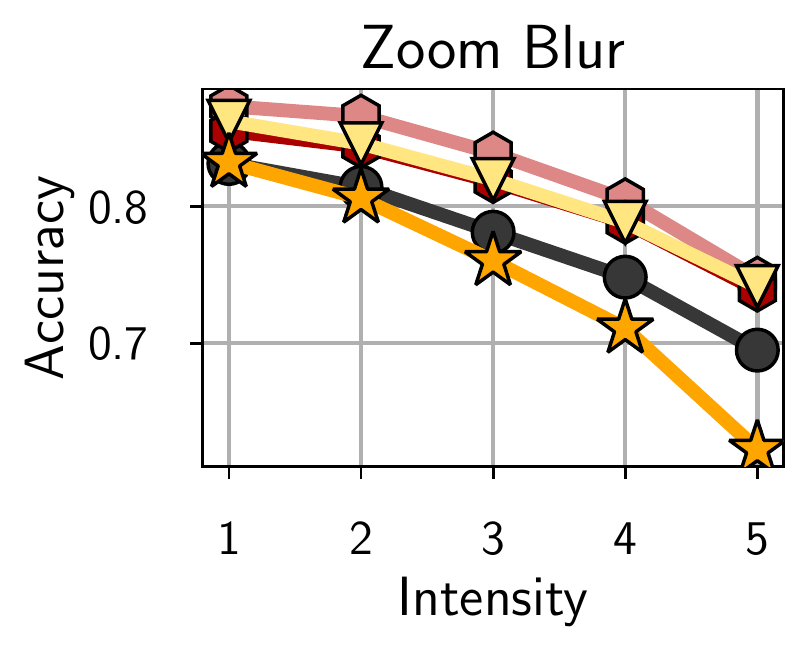} &
	\includegraphics[height=0.11\textwidth]{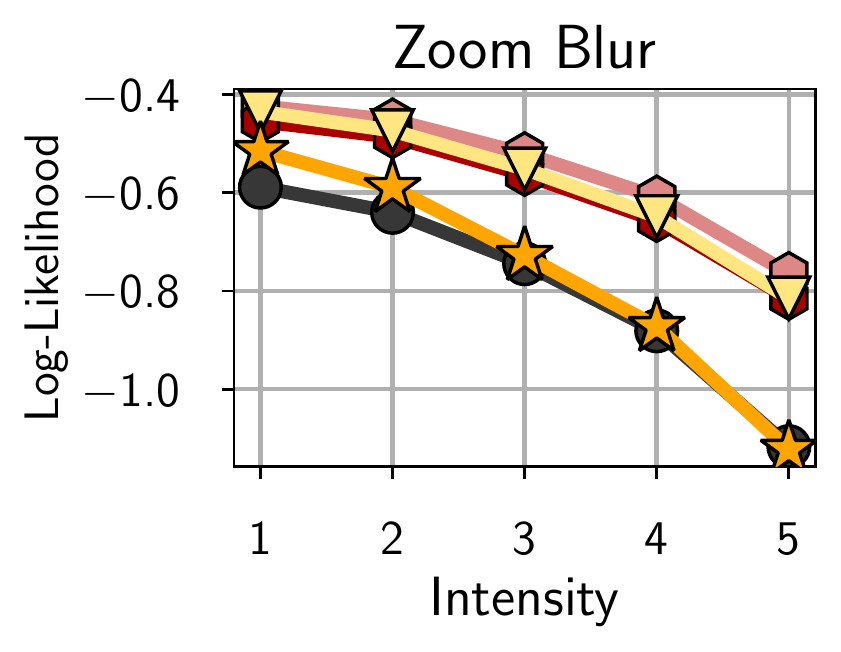} &
	\includegraphics[height=0.11\textwidth]{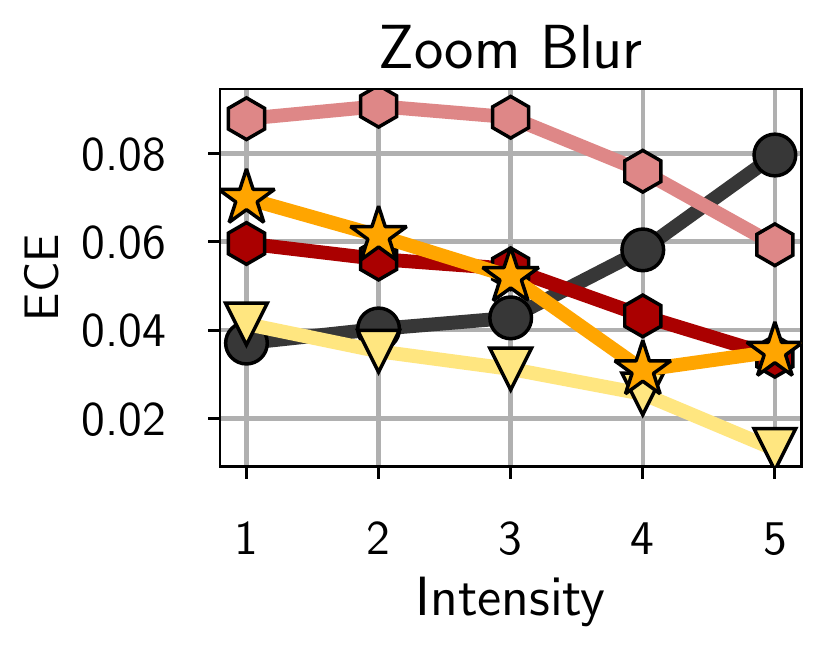}
	\\[-0.2cm]
\end{tabular}\\
\multicolumn{2}{c}{\includegraphics[height=0.07\textwidth]{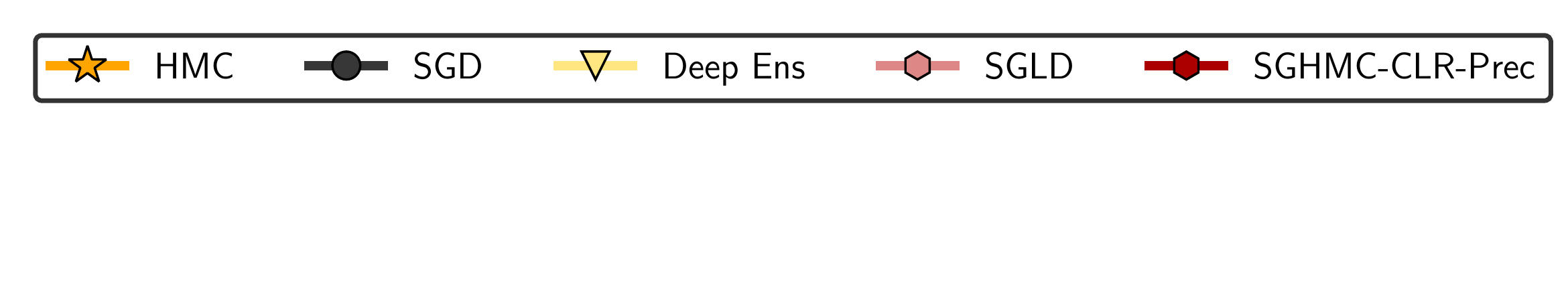}}
\\[-0.5cm]
\end{tabular}

	\caption{
	    \textbf{Performance under corruption.} We show accuracy, log-likelihood and ECE of HMC, SGD, Deep Ensembles, SGLD and SGHMC-CLR-Prec for all $16$ CIFAR-10-C corruptions as a function of corruption intensity.
	    HMC shows poor accuracy on most of the corruptions with a few exceptions.
	    SGLD provides the best robustness on average.
	}
	\label{fig:app_cifar10c_details}
\end{figure*}

\FloatBarrier

\section{BNNs are not Robust to Domain Shift}
\label{sec:app_robustness}

In \autoref{sec:image_classification}, \autoref{fig:cifar10c} we have seen that surprisingly BNNs via HMC underperform significantly on corrupted data from CIFAR-10-C
compared to SGLD, deep ensembles and even MFVI and SGD.
We provide detailed results in \autoref{fig:app_cifar10c_details}.
HMC shows surprisingly poor robustness in terms of accuracy and log-likelihood across the corruptions.
The ECE results are mixed.
In most cases, the HMC ensemble of $720$ models loses to a single SGD solution!

The poor performance of HMC on OOD data is surprising. 
Bayesian methods average the predictions over multiple models for the data, and faithfully represent uncertainty.
Hence, Bayesian deep learning methods are expected to be robust to noise in the data, and are often explicitly 
evaluated on CIFAR-10-C \citep[e.g.][]{wilson2020bayesian,dusenberry2020efficient}.
Our results suggest that the improvements achieved by Bayesian methods on corrupted data may be a sign of \textit{poor} posterior approximation.

\begin{figure}[t]
    \centering
    \includegraphics[height=0.2\textwidth]{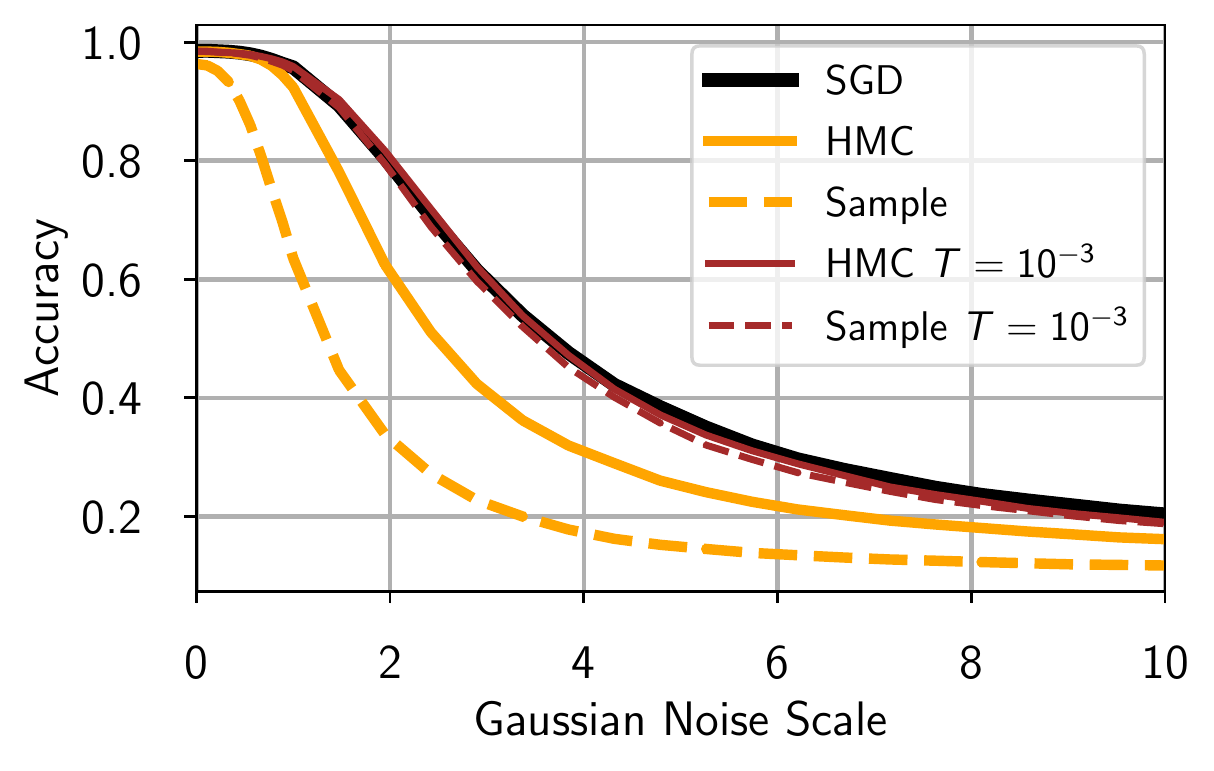}
    \caption{\textbf{Robustness on MNIST.}
    Performance of SGD, BMA ensembles and individual samples constructed by HMC at temperatures $T=1$ and $T=10^{-3}$ on
    the MNIST test set corrupted by Gaussian noise.
    We use a fully-connected network.
    Temperature $1$ HMC shows very poor robustness, while lowering the temperature allows us to close the gap to SGD.
    }
    \label{fig:app_mnist_robustness}
\end{figure}

To further understand the robustness results, we reproduce the same effect on a small fully-connected network with two hidden layers of width $256$ on MNIST.
We run HMC at temperatures $T=1$ and $T=10^{-3}$ and SGD and report the results for both the BMA ensembles and individual samples in \autoref{fig:app_mnist_robustness}.
For all methods, we train the models on the original MNIST training set, and evaluate on the test set with random Gaussian noise $\mathcal N(0, \sigma^2 I)$ of varying
scale $\sigma$.
We report the test accuracy as a function of $\sigma$.
We find that while the performance on the original test set is very close for all methods, the accuracy of HMC at $T=1$ drops much quicker compared to that of SGD as 
we increase the noise scale.

Notably, the individual sample performance of $T=1$ HMC is especially poor compared to SGD.
For example, at noise scale $\sigma=3$ the SGD accuracy is near $60\%$ while the HMC sample only achieves around $20\%$ accuracy!

HMC can be though of as sampling points at a certain sub-optimal level of the training loss, significantly lower than that of SGD solutions.
As a result, HMC samples are individually inferior to SGD solutions.
On the original test data ensembling the HMC samples leads to strong performance significantly outperforming SGD (see \autoref{sec:bnn_evaluation}).
However, as we apply noise to the test data, ensembling can no longer close the gap to the SGD solutions.
To provide evidence for this explanation, we run evaluate HMC at a very low temperature $T=10^{-3}$,
as low temperature posteriors concentrate on high-performance solutions similar to the ones found by SGD.
We find that at this temperature, HMC performs comparably with SGD, closing the gap in robustness
We have also experimented with varying the prior scale but were unable to close the gap in robustness at temperature $T=1$.

We hypothesize that using a lower temperature with HMC would also significantly improve robustness on CIFAR-10-C.
Verifying this hypothesis, and generally understanding the robustness of BNNs further is an exciting direction of future work.

\begin{table*}[h]
\begin{center}
\begin{small}
\begin{sc}
\begin{tabular}{lllll}
\toprule
             & Acc, $T=1$ & Acc, $T = 0.1$ & CE, $T = 1$ & CE, $T = 0.1$ \\
\midrule
BN + Aug     & 87.46     & 91.12         & 0.376      & 0.2818       \\
FRN + Aug    & 85.47     & 89.63         & 0.4337     & 0.317        \\
BN + No Aug  & 86.93     & 85.20         & 0.4006     & 0.4793       \\
FRN + No Aug & 84.27     & 80.84         & 0.4708     & 0.5739      \\
\bottomrule
\end{tabular}
\end{sc}
\end{small}
\end{center}
\caption{\textbf{Role of data augmentation in the cold posterior effect.}
Results of a single chain ensemble constructed with the SGHMC-CLR-Prec sampler of \citet{wenzel2020good} at
temperatures $T=1$ and $T=0.1$ for different combinations of batch normalization (BN) or filter response normalization (FRN) and data augmentation (Aug).
We use the ResNet-20 architecture on CIFAR-10.
Regardless of the normalization technique, the cold posteriors effect is present when data
augmentation is used, and not present otherwise.
}
\label{tab:app_cold_posteriors}
\end{table*}

\begin{figure*}[h]
    \centering
    \includegraphics[width=0.7\textwidth]{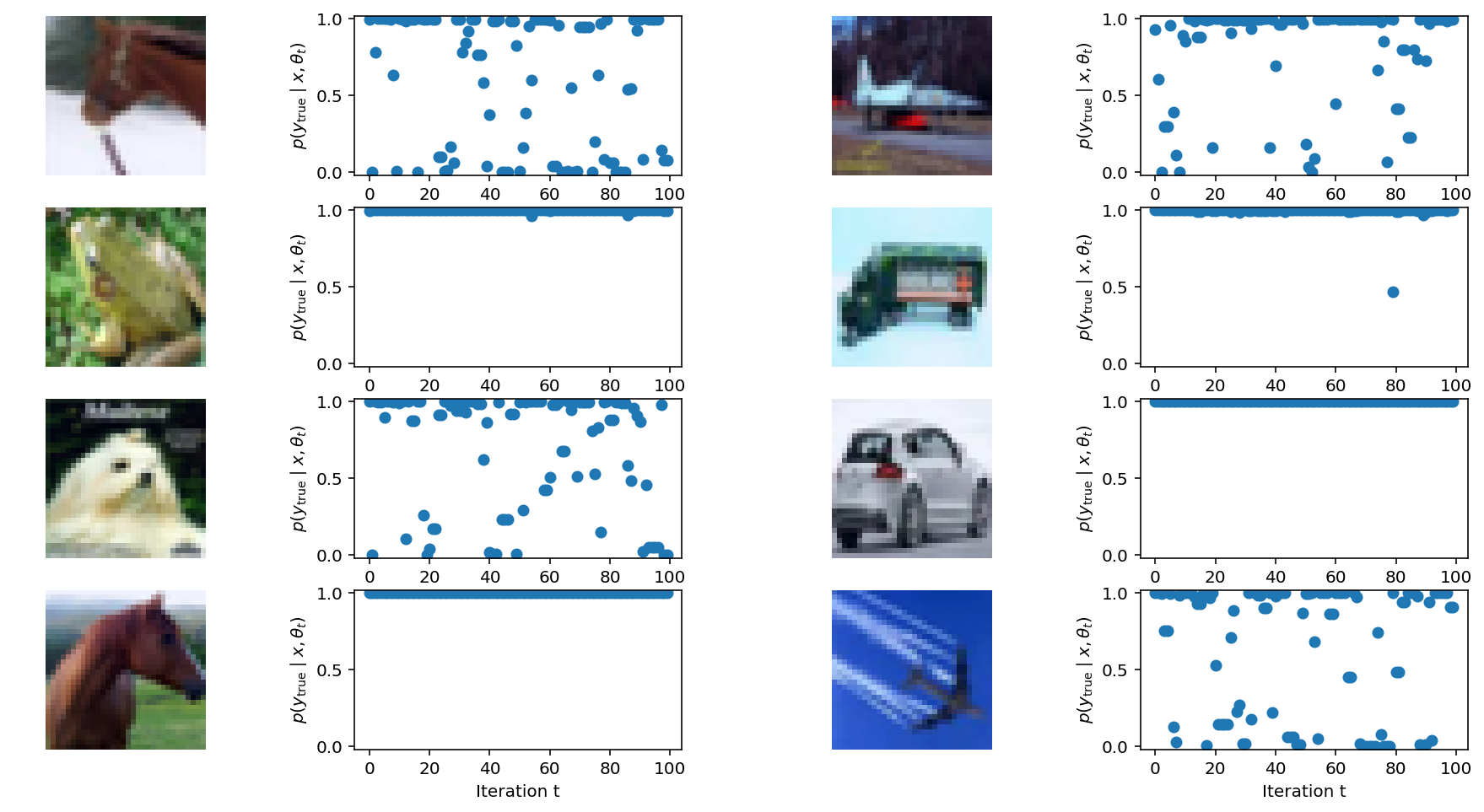}
	\caption{
	    \textbf{HMC samples are (over)confident classifiers.} Plots show the probability assigned by a series of HMC samples to the true label of a held-out CIFAR-10 image.
	    In many cases these probabilities are overconfident (i.e., assign the right answer probability near 0), but there are always \emph{some} samples that assign the true label high probability, so the Bayesian model average is both accurate and well calibrated. These samples were generated with a spherical Gaussian prior with variance $\frac{1}{5}$.
	}
	\label{fig:overconfident}
\end{figure*}

\section{Further discussion of cold posteriors}
\label{sec:app_cold_posteriors}

In \autoref{sec:cold_posteriors} we have seen that the cold posteriors are not needed to
achieve strong performance with BNNs.
We have even shown that cold (as well as warm) posteriors may hurt the performance.
On the other hand, in \autoref{sec:app_robustness} we have shown that lowering the temperature can improve
robustness under the distribution shift, at least for a small MLP on MNIST.
Here, we discuss the potential reasons for why the cold posteriors effect was observed in \citet{wenzel2020good}.

\subsection{What causes the difference with \citet{wenzel2020good}?}

There are several key differences between the experiments in our study and \citet{wenzel2020good}.

First of all, the predictive distributions of SGLD (a version of which was used in \citet{wenzel2020good})
are highly dependent on the hyper-parameters such as the batch size and learning rate, and are inherently biased:
SGLD with a non-vanishing step size samples from a perturbed version of the posterior, both because it omits a Metropolis-Hastings accept-reject step and because its updates include minibatch noise.
Both of these perturbations should tend to make the entropy of SGLD's stationary distribution increase with its step size; we might expect this to translate to approximations to the BMA that are overdispersed.

Furthermore, \citet{wenzel2020good} show in Figure 6 that with a high batch size they achieve good performance at $T=1$ for the CNN-LSTM.
Using the code provided by the autors\footnote{\small\url{https://github.com/google-research/google-research/tree/master/cold_posterior_bnn}}
with default hyper-parameters we achieved strong performance at $T=1$ for the CNN-LSTM (accuracy of $0.855$ and cross-entropy of $0.35$, compared to $0.81$ and $0.45$ reported in Figure 1 of \citet{wenzel2020good});
we were, however, able to reproduce the cold posteriors effect on CIFAR-10 using the same code. 

On CIFAR-10, the main difference between our setup and the configuration in \citet{wenzel2020good} is
the use of batch normalization and data augmentation. 
In the appendix K and Figure 28 of \citet{wenzel2020good}, the authors show that if both
the data augmentation and batch normalization are turned off, we no longer observe the cold posteriors effect.
In \autoref{tab:app_cold_posteriors} we confirm using the code provided by the authors that in fact it is
sufficient to turn off just the data augmentation to remove the cold posteriors effect.
It is thus likely that the results in \citet{wenzel2020good} are at least partly affected by the use of data augmentation.

\section{Visualizing prediction variations in HMC samples}
\label{sec:app_prediction_variations}

In \autoref{fig:overconfident} we visualize the predicted class probability of the true class for $100$ HMC samples on eight different input images.
While on some images the predicted class probability is always close to $1$, on other inputs it is close to $1$ for some of the samples (confidently correct),
close to $0$ for some of the samples (confidently wrong) and in between $0$ and $1$ for the remaining samples (unconfident).
So, some of the samples are individually over-confident, but the ensemble is well-calibrated as other samples assign high probabilities to the correct class.

\end{document}